\documentclass[a4paper,UKenglish,thm-restate,cleveref]{lipics-v2021}

\pdfoutput=1 %
\hideLIPIcs  %
\nolinenumbers

\crefname{claim}{Claim}{Claims}

\title{Learning Tree Pattern Transformations} %

\author{Daniel Neider}{TU Dortmund University, Germany\\Center for Trustworthy Data Science and Security, UA Ruhr, Germany}{daniel.neider@tu-dortmund.de}{https://orcid.org/0000-0001-9276-6342}{}
\author{Leif Sabellek}{CONTACT Research, Germany}{leif.sabellek@contact-software.com}{https://orcid.org/0000-0001-8051-5749}{}
\author{Johannes Schmidt}{Jönköping University, Sweden}{johannes.schmidt@ju.se}{https://orcid.org/0000-0001-8551-1624}{Partially supported by the Swedish Research Council (VR), grant 2022-03214.}
\author{Fabian Vehlken}{Ruhr University Bochum, Germany}{fabian.vehlken@rub.de}{https://orcid.org/0009-0002-1434-3672}{Supported by the Deutsche Forschungsgemeinschaft (DFG, German Research Foundation), grants 448468041 and 532727578.}
\author{Thomas Zeume}{Ruhr University Bochum, Germany}{thomas.zeume@rub.de}{https://orcid.org/0000-0002-5186-7507}{Supported by the Deutsche Forschungsgemeinschaft (DFG, German Research Foundation), grants 448468041 and 532727578.}

\authorrunning{D. Neider, L. Sabellek, J. Schmidt, F. Vehlken, and T. Zeume} %

\Copyright{Daniel Neider, Leif Sabellek, Johannes Schmidt, Fabian Vehlken, and Thomas Zeume} %



\acknowledgements{We are grateful to Florin Manea and Markus Schmid for insightful discussions.}%

\EventEditors{Sudeepa Roy and Ahmet Kara}
\EventNoEds{2}
\EventLongTitle{28th International Conference on Database Theory (ICDT 2025)}
\EventShortTitle{ICDT 2025}
\EventAcronym{ICDT}
\EventYear{2025}
\EventDate{March 25--28, 2025}
\EventLocation{Barcelona, Spain}
\EventLogo{}
\SeriesVolume{328}
\ArticleNo{21}

\newif\ifcomments
\newif\ifchanges
\commentsfalse\changesfalse
\commentstrue\changestrue %

\usepackage{xspace}
\usepackage{adjustbox}

\usepackage{algorithm}
\usepackage[noend]{algorithmic}

\algsetup{indent=2em}

\RequirePackage{adjustbox}
\newcommand{\toplabelled}[2]{%
\setlength{\tabcolsep}{1pt}
\begin{tabular}{r l}
    \adjustbox{valign=t}{#1:} &
    \adjustbox{valign=t}{#2}
\end{tabular}%
}

\RequirePackage{xspace}
\newcommand{\N}{\ensuremath{\mathbb{N}}}

\newcommand{\bigO}{\ensuremath{\mathcal{O}}}

\newcommand{\df}{\ensuremath{\mathrel{\smash{\stackrel{\scriptscriptstyle{
    \text{def}}}{=}}}} \;}
    
\newcommand  {\myclass} [1]  {\ensuremath{\textsf{\upshape #1}}}

\newcommand{\StaClass}[1]{\myclass{#1}\xspace}

\newcommand  {\algorithmicProblem} [1] {\normalfont{\textsc{#1}}\xspace}

\newcommand{\algorithmicProblemDescription}[4][10cm]{
    \vspace{2mm}
    \def\Name{#2}
    \def\Input{#3}
    \def\Question{#4}
      \setlength{\tabcolsep}{1mm}
      \begin{tabular}{rp{#1}r}%
      \textit{Problem:}&\algorithmicProblem{\Name} \\
     \textit{Input:}&\Input \\
     \textit{Question:}&\Question
     \end{tabular}%
    \vspace{2mm}
    }
\newcommand{\AlgorithmicProblemDescription}[4][10cm]{
    \vspace{2mm}
    \def\Name{#2}
    \def\Input{#3}
    \def\Output{#4}
      \setlength{\tabcolsep}{1mm}
      \begin{tabular}{rp{#1}r}%
      \textit{Problem:}&\algorithmicProblem{\Name} \\
     \textit{Input:}&\Input \\
     \textit{Find:}&\Output
     \end{tabular}%
    \vspace{2mm}
    }

\newcommand     {\PTIME}    {\myclass{PTime}}
\newcommand     {\NP}   {\StaClass{NP}}

\newcommand{\CQ}[1][]{\StaClass{CQ}}
\newcommand{\UCQ}[1][]{\StaClass{UCQ}}
\newcommand{\CQneg}[1][]{\StaClass{CQ\ensuremath{^{\mneg}}}}
\newcommand{\UCQneg}[1][]{\StaClass{UCQ\ensuremath{^{\mneg}}}}

\theoremstyle{plain}

\theoremstyle{definition}

\newtheorem*{question*}{Question}
\newtheorem*{openquestion*}{Open question}

\newenvironment{proofsketch}{\begin{proof}[Proof sketch.]}{\end{proof}}

\providecommand {\calC}      {{\mathcal C}\xspace}
\providecommand {\calD}      {{\mathcal D}\xspace}

\providecommand {\calI}      {{\mathcal I}\xspace}

\providecommand {\calN}      {{\mathcal N}\xspace}

\providecommand {\calS}      {{\mathcal S}\xspace}
\providecommand {\calT}      {{\mathcal T}\xspace}

\providecommand {\calX}      {{\mathcal X}\xspace}

\ifcomments
\newcommand{\commentbox}[1]{\noindent\framebox{\parbox{0.98\linewidth}{#1}}}

\newcommand{\acomment}[2]{\ \\ \fbox{\parbox{0.98\linewidth}{{\sc #1}: #2}}}
\newcommand{\mcomment}[2]{{\color{blue}(#1)}\footnote{#1: #2}} %
\else
\newcommand{\commentbox}[1]{}
\newcommand{\mcomment}[2]{}
\newcommand{\acomment}[2]{}
\fi

\ifchanges

\setul{}{0.2mm}
\setstcolor{red}

\else

\fi

\RequirePackage{tikz}
\tikzstyle{narrownode} = [draw, circle, inner sep=1pt]
\tikzstyle{narrowcoverednode} = [circle,fill=orange!40, inner sep=1pt]
\definecolor{solutionCol}{HTML}{f49016}
\tikzstyle{solutionNode} = [draw = solutionCol, very thick]
\tikzstyle{variableSubtree} = [draw, shape=regular polygon, regular polygon sides=5, inner sep=-3pt, rounded corners]

\usetikzlibrary{positioning}
\usetikzlibrary{arrows,backgrounds,calc}
\usetikzlibrary{decorations.shapes}
\pgfdeclarelayer{background}
\pgfsetlayers{background,main}
\newcommand{\convexpath}[2]{
[   
    create hullnodes/.code={
        \global\edef\namelist{#1}
        \foreach [count=\counter] \nodename in \namelist {
            \global\edef\numberofnodes{\counter}
            \node at (\nodename) [draw=none,name=hullnode\counter] {};
        }
        \node at (hullnode\numberofnodes) [name=hullnode0,draw=none] {};
        \pgfmathtruncatemacro\lastnumber{\numberofnodes+1}
        \node at (hullnode1) [name=hullnode\lastnumber,draw=none] {};
    },
    create hullnodes
]
($(hullnode1)!#2!-90:(hullnode0)$)
\foreach [
    evaluate=\currentnode as \previousnode using \currentnode-1,
    evaluate=\currentnode as \nextnode using \currentnode+1
    ] \currentnode in {1,...,\numberofnodes} {
  let
    \p1 = ($(hullnode\currentnode)!#2!-90:(hullnode\previousnode)$),
    \p2 = ($(hullnode\currentnode)!#2!90:(hullnode\nextnode)$),
    \p3 = ($(\p1) - (hullnode\currentnode)$),
    \n1 = {atan2(\y3,\x3)},
    \p4 = ($(\p2) - (hullnode\currentnode)$),
    \n2 = {atan2(\y4,\x4)},
    \n{delta} = {-Mod(\n1-\n2,360)}
  in 
    {-- (\p1) arc[start angle=\n1, delta angle=\n{delta}, radius=#2] -- (\p2)}
}
-- cycle
}

\tikzset{dotted pattern/.style args={#1 and #2}{
   decorate,
   fill,
   decoration={
    shape backgrounds,
    shape=circle,
    shape size=#1,
    shape sep={#2, between center}, 
    },
    draw=none
  },
  dotted pattern/.default={1pt and 1.5mm},
}
\RequirePackage{xifthen}
\RequirePackage{adjustbox}
\RequirePackage{tikz}
\usetikzlibrary{shapes.geometric}
\RequirePackage{tikz-qtree}
\tikzset{aligned/.style={baseline=(current bounding box.center)}}

\newcommand{\tree}[1]{%
    \begin{tikzpicture}[sibling distance=1em, level distance=2em, 
        every node/.style = {align=center}, aligned, remember picture]
        #1
    \end{tikzpicture}%
}

\newcommand{\ltree}[1]{%
    \begin{tikzpicture}[sibling distance=0.1em, level distance=2.5em, 
        every node/.style = {align=center}, aligned, remember picture]%
        #1%
    \end{tikzpicture}%
}

\newcommand{\lltree}[1]{%
    \begin{tikzpicture}[sibling distance=0.1em, level distance=3.5em, 
        every node/.style = {align=center}, aligned, remember picture]%
        #1%
    \end{tikzpicture}%
}

\newcommand{\ttree}[1]{%
    \begin{tikzpicture}[sibling distance=1em, level distance=3em, 
        every node/.style = {align=center},
        every leaf node/.style = {draw, regular polygon, regular polygon sides=3, anchor=center, inner sep=1pt, rounded corners}]
        #1
    \end{tikzpicture}%
}

\makeatletter
\DeclareRobustCommand{\rvdots}{%
  \vbox{
    \baselineskip4\p@\lineskiplimit\z@
    \kern-\p@
    \hbox{.}\hbox{.}\hbox{.}
  }}
\makeatother
 
\newcommand{\trafo}[3][2pt]{%
    \setlength{\tabcolsep}{#1}
    \begin{tabular}{c c c}
        \adjustbox{valign=m}{#2} & 
        \adjustbox{valign=m}{\Large$ \rightsquigarrow $} &
        \adjustbox{valign=m}{#3}
    \end{tabular}%
    
}

\newcommand{\ttrafo}[3][]{%
    \ifthenelse{\isempty{#1}}%
    {\ensuremath{#2 \rightsquigarrow #3}}%
    {\ensuremath{#2 \rightsquigarrow_{#1} #3}}%
}

\newcommand{\tsteptrafo}[4][]{%
    \ifthenelse{\isempty{#1}}%
    {\ensuremath{#3 \rightsquigarrow^{#2} #4}}%
    {\ensuremath{#3 \rightsquigarrow_{#1}^{#2} #4}}%
}

\newcommand{\body}{\ensuremath{\mathrm{body}}\xspace}
\newcommand{\head}{\ensuremath{\mathrm{head}}\xspace}
\newcommand{\map}{\ensuremath{\mathrm{map}}\xspace}

\newcommand{\subtree}{\ensuremath{\mathrm{subtree}}\xspace}

\newcommand{\Ell}{\ensuremath{\mathcal{L}}}
\newcommand{\lang}[1][\Ell]{\ensuremath{\Theta_{\Ell}}\xspace}
\newcommand{\context}{\operatorname{context}}

\newcommand{\SAT}{\algorithmicProblem{SAT}}

\newcommand*{\Iltis}{\textsc{Iltis}\xspace}
\newtheorem*{openproblem}{Open problem}
\newcommand{\proofstep}[1]{\vspace{-1.2em}\subparagraph*{#1}}

\begin{document}

\maketitle

\begin{abstract}
Explaining why and how a tree $t$ structurally differs from another tree $t^\star$ is a question that is encountered throughout computer science, including in understanding tree-structured data such as XML or JSON data. 
In this article, we explore how to learn explanations for structural differences between pairs of trees from sample data: suppose we are given a set $\{(t_1, t_1^\star),\dots, (t_n, t_n^\star)\}$ of pairs of labelled, ordered trees;  is there a small set of rules that explains the structural differences between all pairs $(t_i, t_i^\star)$? This raises two research questions:
(i) what is a good notion of ``rule'' in this context?; and (ii) how can sets of rules explaining a data set be learned algorithmically?

We explore these questions from the perspective of database theory by (1)
introducing a pattern-based specification language for tree transformations;
(2) exploring the computational complexity of variants of the above algorithmic
problem, e.g.\ showing \NP-hardness for very restricted variants; and (3)
discussing how to solve the problem for data from CS education research using
\SAT solvers.
\end{abstract}

\section{Introduction}
\label{section:introduction}
Explaining why and how a tree $t$ structurally differs from another tree $t^\star$ is a question that is encountered throughout computer science, including in understanding tree-structured data such as XML or JSON data, analysis of formal syntax trees, or code optimization. 

In this article, we explore how to learn explanations for structural differences between pairs of trees from sample data: suppose we are given a set $\{(t_1, t_1^\star),\dots, (t_n, t_n^\star)\}$ of pairs of labelled, ordered trees;  is there a small set $\Gamma$ of rules that explains the structural differences between all pairs $(t_i, t_i^\star)$? This raises two research questions:
\begin{romanenumerate}
 \item What is a good notion of ``rule'' in this context?
 \item How can sets of rules explaining a data set be learned algorithmically? 
\end{romanenumerate}
We explore these questions from the perspective of database theory. 

Our initial motivation for addressing these questions comes from CS education research on understanding common mistakes and misconceptions of students in introductory courses on \emph{Logic in Computer Science}. A common educational task for students in such courses is to model scenarios described in natural language by logical formulas. A teacher provides the description and a solution formula $\varphi$, and students try to model the description with a formula $\varphi^\star$. The student attempt is correct if it is equivalent to $\varphi$.
Empirically, most students try to write a formula that resembles the ``logical structure of the provided description'' very closely. If the teacher provides solution formulas with that aspect in mind, %
e.g. by providing a formula with an implication for ``if \dots, then \dots'' statements rather than other equivalent formulas,
a natural approach for understanding what kind of mistakes students make in modelling is to find explanations for structural differences between the syntax trees of the solution formula $\varphi$ and $\varphi^\star$ provided by students. 
Besides modelling, another type of exercise is having to transform a formula into some normal form. 
Here, individual transformation steps by students may incorrectly transform a formula $\varphi$ into a formula $\varphi^\star$. To identify the types of mistakes students make, it is also helpful to look for explanations for structural differences between such two formulas.

Within the educational support system \Iltis \cite{SchmellenkampVZ24}, a large set of sample data from students has been collected, e.g.\ $> 500$ pairs $(\varphi, \varphi^\star)$ of propositional formulas for many modelling exercises. In a preliminary analysis of this data with a custom approach, common types of mistakes in propositional logic have been identified by Schmellenkamp et al.\ \cite{SchmellenkampLZ23}. For scaling this analysis and to ensure that a broad set of explanations are taken into account, a more systematic approach is needed. This work takes a first step in this direction.

While our motivation comes from CS education research, we believe that the raised conceptual questions are relevant and interesting for tree-shaped data in general. We therefore state them in the language of database theory and address them with techniques from this field. 

\paragraph*{Contributions}

\begin{figure}
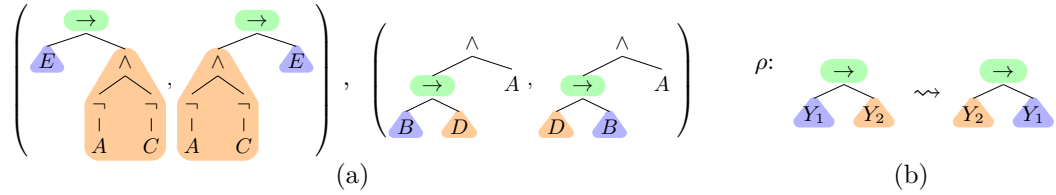

			\begin{minipage}[c][2cm][b]{0.65\textwidth}
			\centering
        \scalebox{0.8}{$\left(%
            \tree{%
                \Tree[.\node (root) {\ensuremath{\to}}; \node (left) {\ensuremath{E}}; [.\node (rroot) {\ensuremath{\land}}; [.\node (not) {\ensuremath{\lnot}}; \node (a) {\ensuremath{A}}; ] [.\node (not2) {\ensuremath{\lnot}}; \node (c) {\ensuremath{C}}; ] ] ]
                \coordinate (r_l) at ($(root.west) + (3pt,0.75pt)$);
                \coordinate (r_r) at ($(root.east) + (-3pt,0.75pt)$);
                \coordinate (y_t) at ($(left.north) + (0,-2pt)$);
                \coordinate (y_b_l) at ($(left.south west) + (2pt,4pt)$);
                \coordinate (y_b_r) at ($(left.south east) + (-2pt,4pt)$);
                \begin{pgfonlayer}{background}
                    \fill[blue,opacity=0.3] \convexpath{y_b_l,y_t,y_b_r}{2.5pt};%
                    \fill[orange,opacity=0.4] \convexpath{a,not,rroot,not2,c}{7pt};
                    \fill[green,opacity=0.3] \convexpath{r_l,r_r}{5pt};%
                \end{pgfonlayer}
            }, 
            \tree{%
                \Tree[.\node (root) {\ensuremath{\to}}; [.\node (rroot) {\ensuremath{\land}}; [.\node (not) {\ensuremath{\lnot}}; \node (a) {\ensuremath{A}}; ] [.\node (not2) {\ensuremath{\lnot}}; \node (c) {\ensuremath{C}}; ] ] \node (left) {\ensuremath{E}}; ]
                \coordinate (r_l) at ($(root.west) + (3pt,0.75pt)$);
                \coordinate (r_r) at ($(root.east) + (-3pt,0.75pt)$);
                \coordinate (y_t) at ($(left.north) + (0,-2pt)$);
                \coordinate (y_b_l) at ($(left.south west) + (2pt,4pt)$);
                \coordinate (y_b_r) at ($(left.south east) + (-2pt,4pt)$);
                \begin{pgfonlayer}{background}
                    \fill[blue,opacity=0.3] \convexpath{y_b_l,y_t,y_b_r}{2.5pt};%
                    \fill[orange,opacity=0.4] \convexpath{a,not,rroot,not2,c}{7pt};
                    \fill[green,opacity=0.3] \convexpath{r_l,r_r}{5pt};%
                \end{pgfonlayer}
            }
        \right)$} \text{, }
        \scalebox{0.8}{$\left(%
            \tree{%
                \Tree[
                    .\ensuremath{\land}
                    [.\node (root) {\ensuremath{\to}}; \node (left) {\ensuremath{B}}; \node (right) {\ensuremath{D}}; ]
                    \ensuremath{A}
                ]
                \coordinate (r_l) at ($(root.west) + (3pt,0.75pt)$);
                \coordinate (r_r) at ($(root.east) + (-3pt,0.75pt)$);
                \coordinate (y_t) at ($(left.north) + (0,-2pt)$);
                \coordinate (y_b_l) at ($(left.south west) + (2pt,4pt)$);
                \coordinate (y_b_r) at ($(left.south east) + (-2pt,4pt)$);
                \coordinate (z_t) at ($(right.north) + (0,-2pt)$);
                \coordinate (z_b_l) at ($(right.south west) + (2pt,4pt)$);
                \coordinate (z_b_r) at ($(right.south east) + (-2pt,4pt)$);
                \begin{pgfonlayer}{background}
                    \fill[blue,opacity=0.3] \convexpath{y_b_l,y_t,y_b_r}{2.5pt};%
                    \fill[orange,opacity=0.4] \convexpath{z_b_l,z_t,z_b_r}{2.5pt};
                    \fill[green,opacity=0.3] \convexpath{r_l,r_r}{5pt};%
                \end{pgfonlayer}
            }, 
            \tree{%
                \Tree[
                    .\ensuremath{\land}
                    [.\node (root) {\ensuremath{\to}}; \node (right) {\ensuremath{D}}; \node (left) {\ensuremath{B}}; ]
                    \ensuremath{A}
                ]
                \coordinate (r_l) at ($(root.west) + (3pt,0.75pt)$);
                \coordinate (r_r) at ($(root.east) + (-3pt,0.75pt)$);
                \coordinate (y_t) at ($(left.north) + (0,-2pt)$);
                \coordinate (y_b_l) at ($(left.south west) + (2pt,4pt)$);
                \coordinate (y_b_r) at ($(left.south east) + (-2pt,4pt)$);
                \coordinate (z_t) at ($(right.north) + (0,-2pt)$);
                \coordinate (z_b_l) at ($(right.south west) + (2pt,4pt)$);
                \coordinate (z_b_r) at ($(right.south east) + (-2pt,4pt)$);
                \begin{pgfonlayer}{background}
                    \fill[blue,opacity=0.3] \convexpath{y_b_l,y_t,y_b_r}{2.5pt};%
                    \fill[orange,opacity=0.4] \convexpath{z_b_l,z_t,z_b_r}{2.5pt};
                    \fill[green,opacity=0.3] \convexpath{r_l,r_r}{5pt};%
                \end{pgfonlayer}
            }
        \right)$}
        
        \vspace{0mm}
        (a)
        			\end{minipage}
        \hspace{5mm}
        	\begin{minipage}[c][2cm][b]{0.3\textwidth}
						\centering
         \toplabelled{$ \rho $}{%
                \scalebox{0.9}{%
                \trafo{%
                    \tree{%
                    \Tree[.\node (root) {\ensuremath{\to}}; \node (left) {\ensuremath{Y_1}}; \node (right) {\ensuremath{Y_2}}; ]
                    \coordinate (r_l) at ($(root.west) + (3pt,0.75pt)$);
                    \coordinate (r_r) at ($(root.east) + (-3pt,0.75pt)$);
                    \coordinate (y_t) at ($(left.north) + (0,-2pt)$);
                    \coordinate (y_b_l) at ($(left.south west) + (2pt,4pt)$);
                    \coordinate (y_b_r) at ($(left.south east) + (-2pt,4pt)$);
                    \coordinate (z_t) at ($(right.north) + (0,-2pt)$);
                    \coordinate (z_b_l) at ($(right.south west) + (2pt,4pt)$);
                    \coordinate (z_b_r) at ($(right.south east) + (-2pt,4pt)$);
                    \begin{pgfonlayer}{background}
                        \fill[blue,opacity=0.3] \convexpath{y_b_l,y_t,y_b_r}{2.5pt};%
                        \fill[orange,opacity=0.4] \convexpath{z_b_l,z_t,z_b_r}{2.5pt};
                        \fill[green,opacity=0.3] \convexpath{r_l,r_r}{5pt};%
                    \end{pgfonlayer}
                    }
                }{%
                    \tree{%
                        \Tree[.\node (root) {\ensuremath{\to}}; \node (right) {\ensuremath{Y_2}}; \node (left) {\ensuremath{Y_1}}; ]
                        \coordinate (r_l) at ($(root.west) + (3pt,0.75pt)$);
                        \coordinate (r_r) at ($(root.east) + (-3pt,0.75pt)$);
                        \coordinate (y_t) at ($(left.north) + (0,-2pt)$);
                        \coordinate (y_b_l) at ($(left.south west) + (2pt,4pt)$);
                        \coordinate (y_b_r) at ($(left.south east) + (-2pt,4pt)$);
                        \coordinate (z_t) at ($(right.north) + (0,-2pt)$);
                        \coordinate (z_b_l) at ($(right.south west) + (2pt,4pt)$);
                        \coordinate (z_b_r) at ($(right.south east) + (-2pt,4pt)$);
                        \begin{pgfonlayer}{background}
                            \fill[blue,opacity=0.3] \convexpath{y_b_l,y_t,y_b_r}{2.5pt};%
                            \fill[orange,opacity=0.4] \convexpath{z_b_l,z_t,z_b_r}{2.5pt};
                            \fill[green,opacity=0.3] \convexpath{r_l,r_r}{5pt};%
                        \end{pgfonlayer}
                    }
                }}}
                
                \vspace{4mm}
                (b)
        			\end{minipage}
    
        \caption{(a) Two pairs of syntax trees for the pairs $(E \to (\lnot A \land \lnot C), (\lnot A \land \lnot C) \to E) $ and $ ((B \to D) \wedge A, (D \to B) \wedge A) $ of propositional formulas. (b) A tree pattern transformation explaining the structural differences of both pairs. The transformation selects a $\rightarrow$-labelled node and swaps its subtrees $Y_1$ and $Y_2$.}
        \label{fig:intro-example-highlighted-matches}
\end{figure}

Towards a notion of rules for explaining structural differences between pairs of trees, we  introduce a tree pattern based transformation language for labelled, ordered  trees (see \cref{section:tree-transformations}). For us, a tree pattern essentially is a tree whose nodes are labelled with labels, node variables, and tree variables. Such a tree pattern can be matched into a tree, meaning that nodes of the pattern are injectively\footnote{While injective semantics are less common for tree patterns, they are more natural in the context of tree transformations, see \cref{section:tree-transformations}.} mapped to nodes of the tree respecting the tree structure and labels. A tree pattern transformation is of the form \ttrafo{\sigma}{\sigma^\star} for two tree patterns $\sigma$ and~$\sigma^\star$. It is applied to a tree $t$ by matching $\sigma$ into $t$ and constructing a tree $t^\star$ according to the shape of $\sigma^\star$ and according to how variables are assigned (see \cref{fig:intro-example-highlighted-matches} for an illustration).

We then study algorithmic properties of tree pattern transformations (see \cref{section:algorithmic-limitations}). We are particularly interested in variants of the following algorithmic problem:

\AlgorithmicProblemDescription[\textwidth-2cm]{LearningTreeTransformations}
    {
        Pairs $ (t_1, t_1^\star), \dots, (t_n, t_n^\star) $ of labelled, ordered  trees, and 
        $s, r \in \N$.
    }
    {
        A set $\Gamma$ of at most $r$ tree pattern transformations such that $t_i$ can be transformed into $t^\star_i$ with transformations from $\Gamma$ in at most $s$ steps, for all $i$.  
    }
It turns out that this algorithmic problem is already \NP-hard, even for very restricted inputs. For instance, we show that it is hard when fixing $s=1$ as well as when fixing $s = 3$ and $r = 2$. 
Due to the hardness, we discuss how to attack the problem by encoding it into \SAT and  employing a \SAT solver (see \cref{section:sat-solving}). We apply this approach exemplarily to educational data (see \cref{section:SAT-solving}).

\paragraph*{Related work}
Tree patterns have been studied extensively for tree-shaped data by the database theory community. Among other things, the computational complexity of algorithmic problems for tree patterns, such as containment and minimisation, has received much attention (see e.g.\ \cite{CzerwinskiMNP18}, also for an extensive list of references). Our notion of tree patterns slightly differs from the standard notion, as we require matches to be injective and use tree variables. We deviate to accommodate our motivation to learn structural differences, as discussed in the next section.%

Manipulation of tree-shaped data has been studied in the context of tree transductions \cite{LemayNG06, BojanczykD20}, especially in the context of XML \cite{Schwentick07, LemayMN10}. Edit distances and similar measures have also been studied \cite{Bille05, ZhangSS92}. We introduce another formalism for manipulating trees, tree pattern transformations, as it is unclear how to extract high-level explanations for structural differences between trees from existing formalisms.

Learning from examples has been studied before in the context of tree-shaped objects. Tree patterns representing queries on graphs are learned in~\cite{CW16}. The complexity of finding string patterns common to a set of strings is studied in~\cite{Angluin80}. Syntax trees of LTL and PSL formulas have been learned by employing \SAT solvers in \cite{NeiderG18} and \cite{RoyFN20}. %
Our setting conceptually differs from these standard approaches as we aim at learning tree pattern transformations explaining differences between pairs of trees (instead of learning trees only).
In the area of automata learning, learning from only positive examples has been studied \cite{AvellanedaP18, RoyGBNXT23}.
 In contrast to all these approaches, typically we are interested in learning multiple transformations, whereas in the above learning problems only a single ``object'' is learned (e.g.\ a string pattern, an LTL formula, an automaton, and so on).%

\paragraph*{Outline}
We introduce tree patterns and tree transformations based on these in \cref{section:tree-transformations}. In \cref{section:algorithmic-limitations}, we study algorithmic properties of the problem \algorithmicProblem{LearningTreeTransformations} and sketch a SAT-based approach to solve it in practice before presenting an extension of tree transformations in \cref{section:interval-variables}. 
Detailed proofs and additional explanations can be found in the appendix.

\section{Pattern-Based Tree Transformations}
\label{section:tree-transformations}
The goal of our tree transformation language is to explain structural differences between trees. Before formalizing our language, we discuss some of our design choices and fix notations.

\subparagraph*{Towards explaining structural differences between trees}

In many contexts, structural differences between trees $t$ and $t^\star$ are mostly local in the following sense: $t$ can be transformed into $t^\star$ by ``reshuffling'' a small subtree $t_v$ rooted at some node $v$ of $t$ and keeping the rest of $t$ as is. This basic idea is already illustrated in \cref{fig:intro-example-highlighted-matches}. Formally, a tree pattern transformation will be of the form $ \ttrafo{\sigma}{\sigma^\star} $ for two tree patterns $ \sigma $ and $ \sigma^\star $ called \emph{body} and \emph{head}, respectively. A transformation is applied to a tree $ t $ by matching its body into $ t $ and manipulating the ``covered'' subtree as indicated by the head. To this end, body and head may use explicit labels, node variables to ``copy'' labels of single nodes, and tree variables to ``copy'' subtrees. %

\begin{example}\label{ex:motivating-language}
    Consider the  tree pattern transformation  $\rho\colon %
            \scalebox{0.8}{
                \trafo{%
                \ltree{%
                    \Tree[.\ensuremath{x_1} \ensuremath{Y_1} \ensuremath{Y_2} ]
                }
            }{%
                \ltree{%
                    \Tree[.\ensuremath{x_1} \ensuremath{Y_2} \ensuremath{Y_1} ]
                }
            }}
        $
        with node variable $x_1$ and tree variables $Y_1$ and $Y_2$. The structural differences between the two pairs 
        \[
        \scalebox{0.75}{
        $
        \left(
            \tree{%
                \Tree[.\ensuremath{a} 
                        [.\ensuremath{b} \ensuremath{d} \ensuremath{e} ]
                        \ensuremath{c} ]
            }, 
            \tree{%
                \Tree[.\ensuremath{a} 
                        \ensuremath{c} 
                        [.\ensuremath{b} \ensuremath{d} \ensuremath{e} ]
                    ]
            }
        \right),
        \left(%
            \tree{%
                \Tree[.\ensuremath{b} 
                [.\ensuremath{e} 
                        \ensuremath{d} \ensuremath{g} ]
                        \ensuremath{c} 
                    ]
            }, 
            \tree{%
                \Tree[.\ensuremath{b} 
                [.\ensuremath{e} 
                        \ensuremath{g} 
                        \ensuremath{d}
                    ]
                        \ensuremath{c} 
                    ]
            }
        \right)
        $}
		\]
            of trees are explained by $\rho$ by applying it to the root in the first pair, and to the left child of the root in the second pair. The node variable $ x_1 $ then copies the labels $ a $ and $ e $, respectively, and the tree variables $ Y_1 $ and $ Y_2 $ copy suitable subtrees.     \qed
\end{example}

Our goal to use tree pattern transformations to explain structural differences between trees requires to exactly capture the structure of trees. This influences our choice of semantics for tree patterns as follows. We require that
\begin{enumerate}
    \item tree patterns can be matched anywhere in trees (and not just root nodes); and
    \item tree pattern nodes with $ k $ children must be matched to tree nodes with exactly $ k $ children.
\end{enumerate}%

\noindent
Requirement 1 allows for more flexible explanations as is already illustrated in \cref{ex:motivating-language}. Requirement 2, although non-standard, is desirable in our context because we use tree patterns as a building block for tree transformations.

\begin{example}\label{ex:motivate-degree}
    Consider the tree $t\colon \scalebox{0.8}{ \tree{%
                \Tree[.\ensuremath{a} 
                    [.\ensuremath{b} ] 
                    [.\ensuremath{c} \ensuremath{d} ] 
                ]}
            }$ and transformation $\rho\colon 
            \scalebox{0.8}{\trafo{%
                \ltree{%
                    \Tree[.\ensuremath{x_1} \ensuremath{x_2} ]
                }
            }{%
                \ltree{%
                    \Tree[.\ensuremath{x_2} ]
                }
						}}
$
    with node variables $ x_1 $ and $ x_2 $.
    Here, the tree pattern node labelled with $ x_1 $ has fewer children than the node labelled $ a $ in $ t $. If matching $ \rho $ to $ t $ by mapping $x_1$ to the $a$-labelled node and $x_2$ to the $b$-labelled node was allowed, the result would be a single node labelled $ b $. The $c$-labelled node and its $ d $-labelled child would be removed without a ``good explanation''.

    Now consider a slightly different transformation $ \rho'\colon
            \scalebox{0.8}{\trafo{%
                \ltree{%
                    \Tree[.\ensuremath{x_1} \ensuremath{x_2} \ensuremath{Y_1} \ensuremath{Y_2} ]
                }
            }{%
                \ltree{%
                    \Tree[.\ensuremath{x_1} \ensuremath{x_2} \ensuremath{Y_1} ]
                }
            }
            }$
    in which the body's root node has more children than the $a$-labelled node in $ t $. Matching the body of $\rho'$ into $ t $ would require the embedding to be non-injective, for example, the tree variables $ Y_1 $ and $ Y_2 $ could both capture the subtree rooted at $ c $. But then it is not clear what the result of applying $ \rho' $ to $ t $ would be, because the subtree rooted at $ c $ is supposed to be copied (due to $ Y_1 $) as well as removed (due to $ Y_2 $). To avoid this, we require embeddings to be injective.
    \qed
\end{example} 

We highlight that other design choices are possible, whose study we leave for future work.

\subparagraph*{Trees and tree patterns} A \emph{$\Sigma$-labelled (ordered) tree} $t$ is a prefix-closed subset $V \subseteq \N^*$ together with a labelling function $\ell\colon V \to \Sigma$. We also write trees $t$ as $(V, E, \ell)$ where $E$ is the set of edges induced by $V$, i.e.\ $(u, v) \in E$ if $v j = u$ for some $j \in \N$. The children of a node $v \in V$ are ordered according to the natural linear order on $\N$, i.e.\ for $u_1 \df v j_1$ and $u_2 \df v j_2$ we have $u_1 < u_2$ if $j_1 < j_2$. For a tree $t$ and a node $v$, we write $t_v$ for the subtree of $t$ rooted at~$v$. A \emph{context} $\operatorname{context}(t, h)$ is a tree $t = (V, E, \ell)$ with a \emph{hole} $h \in V$. A tree $t'$ can be \emph{plugged} into a context $\operatorname{context}(t, h)$ by replacing the node $h$ by $t'$ in the tree $t$. 

A \emph{tree pattern} $\sigma$ is a $\Sigma \uplus \calN \uplus \calT$-labelled tree, where $\calT$-labelled nodes must be leaves. The elements of $\Sigma$ are called \emph{labels}, the elements of $\calN$ are called \emph{node variables}, and the elements of $\calT$ are called \emph{tree variables}. A \emph{match} $\mu$ of a tree pattern $\sigma = (V_\sigma, E_\sigma, \ell_\sigma)$ in a tree $t = (V_t, E_t, \ell_t)$ is an injective mapping $\mu\colon V_\sigma \rightarrow V_t$ from nodes of $\sigma$ to nodes of $t$ such that 
\begin{enumerate}
    \item labels are mapped consistently, i.e.\ $\ell_\sigma(v) = \ell_t(\mu(v))$ for all $\Sigma$-labelled nodes in $V_\sigma$; 
    \item node variables are mapped consistently, i.e.\ if $u, v \in V_\sigma$ are labelled with the same node variable, then the nodes $\mu(u)$ and $\mu(v)$ must have the same label; 
    \item tree variables are mapped consistently, i.e.\ if $u, v \in V_\sigma$ are labelled with the same tree variable, then the subtrees of $t$ rooted at $\mu(u)$ and $\mu(v)$ are isomorphic; and
    \item \label{item:tree-pattern} the number and order of children is preserved, i.e.\ if $v \in V_\sigma$ has the children $u_1 < \ldots < u_k$ then $\mu(v)$ has exactly the children $\mu(u_1) < \ldots < \mu(u_k)$.
\end{enumerate}

Condition \ref{item:tree-pattern} is non-standard, but essential for an intuitive notion of tree pattern transformations (see discussion above).

\subparagraph*{Tree pattern transformations} The idea of tree pattern transformations is to select a subtree $t_r$ of a tree $t$ rooted at a node $\nu$ by matching a tree pattern $\sigma$, and then to construct a new tree $t^\star$ by plugging a re-combination $t^\star_\nu$ of $t_\nu$ according to a tree pattern $\sigma^\star$ into  the $\text{context}(t, \nu)$. The patterns $\sigma$ and $\sigma^\star$ may use the same node and tree variables, which allows for rearranging whole subtrees. For an example of a tree pattern transformation and its application we refer to \cref{fig:intro-example-highlighted-matches}.

A \emph{tree pattern transformation} $\rho\colon \ttrafo{\sigma}{\sigma^\star}$ 
consists of two tree patterns $ \sigma $ and $ \sigma^\star $ such that all variables occurring in $\sigma^\star$ also occur in $\sigma$. The pattern $\sigma$ is called \emph{body}, $\sigma^\star$ is called \emph{head}, and $\rho$ is called \emph{name} of the transformation. 

The semantics of tree pattern transformations is defined as follows.  A tree pattern transformation $\rho\colon \sigma \rightsquigarrow \sigma^\star$  \emph{transforms} a tree $t$ into a tree $t^\star$, written as $t^\star \in \rho(t)$ or $t \rightsquigarrow_\rho t^\star$, if there is a match $\mu_{\sigma, t}$ of $\sigma$ into $t$ with root $\nu$  and a match $\mu_{\sigma^\star, t^\star}$ of $\sigma^\star$ into $t^\star$ with root $\nu^\star$ such that 
\begin{enumerate}
 \item $t$ and $t^\star$ only differ on the subtrees $t_\nu$ and $t^\star_{\nu^\star}$, i.e.\ the contexts $\text{context}(t, \nu)$ and $\text{context}(t^\star, \nu^\star)$ are isomorphic (in particular, $\nu$ and $\nu^\star$ have the same positions in $t$ and $t^\star$);\label{def:contexts-of-applied-root-isomorphic}
 \item the subtree $t_\nu$ is rearranged into the subtree $t^\star_{\nu^\star}$, i.e.\ \label{def:transformation-properly-rearranges}
	\begin{romanenumerate}
		\item if nodes $u$ in $\sigma$ and $u^\star$ in $\sigma^\star$ are labelled with a node variable $x$, then  $\mu_{\sigma, t}(u)$ and $\mu_{\sigma^\star, t^\star}(u^\star)$ have the same label; and\label{def:transformation-properly-rearranges-node-variables}
		\item if nodes $u$ in $\sigma$ and $u^\star$ in $\sigma^\star$ are labelled with a tree variable $X$, then the subtrees of $t$ and $t^\star$ rooted at $\mu_{\sigma, t}(u)$ and $\mu_{\sigma^\star, t^\star}(u^\star)$ are isomorphic.\label{def:transformation-properly-rearranges-tree-variables}
	\end{romanenumerate}
\end{enumerate}

When $t \rightsquigarrow_\rho t^\star$, we also say that the tree pattern transformation $\rho$ \emph{explains} the pair $(t, t^\star)$, as $\rho$ can be seen as an explanation for the structural differences of $t$ and $t^\star$. Testing whether $\rho$ explains $(t, t^\star)$ is simple. 

\begin{proposition}\label{thm:propositionTransformationExplains}
    Given a tree pattern transformation $\rho\colon \ttrafo{\sigma}{\sigma^\star}$ and a pair $(t, t^\star)$ of trees, it can be tested in polynomial time whether $\rho$ can transform $t$ into $t^\star$, i.e.\ whether $t \rightsquigarrow_\rho t^\star$.
\end{proposition}

\begin{proofsketch}
	Try all pairs $(\nu, \nu^\star)$ of nodes of $t$ and $t^\star$ as roots for matches of $\sigma$ and~$\sigma^\star$.
\end{proofsketch}

For a set $\Gamma$ of tree pattern transformations and a pair $(t, t^\star)$ of trees, we also  write $t \rightsquigarrow_\Gamma^s t^\star$ if $t$ can be transformed into $t^\star$ with transformations from $\Gamma$ in (at most) $s$ steps. We also say that $\Gamma$ explains $(t, t^\star)$ in $s$ steps.
Note that we do not need an explicit wildcard label which is common for tree patterns, because it can be simulated by variables that only occur in the body of a transformation.

\section{Learning Tree Pattern Transformations}
\label{section:algorithmic-limitations}

In this section we study algorithmic properties of the learning problem for tree pattern transformations. Formally, we consider the following algorithmic problem:

\AlgorithmicProblemDescription[\textwidth-2.7cm]{LearningTreeTransformations}
    {
        Pairs $ (t_1, t_1^\star), \dots, (t_n, t_n^\star) $ of ordered, labelled trees, 
        and $s$, $r \in \N$.
    }
    {
        A set $\Gamma$ of at most $r$ tree pattern transformations that explains all pairs $(t_i, t_i^\star)$ in at most $s$ steps, i.e.\ a set $\Gamma$ such that $t_i \rightsquigarrow_\Gamma^s t^\star_i$, for all $ i $.
    }

    For a pair $ (t_i, t_i^\star) $, we sometimes call $ t_i $ source tree and $ t_i^\star $ target tree. Observe that the problem is trivial if $ r \ge n $, because in this case, one transformation $ \rho_i\colon \ttrafo{t_i}{t_i^\star} $ for each pair of trees can be chosen.

\subsection{Learning Tree Pattern Transformations Is Hard}
\label{section:learning-hard}

It is easy to see that \algorithmicProblem{LearningTreeTransformations} is \NP-hard in general.\footnote{Formally, hardness is for the decision version of \algorithmicProblem{LearningTreeTransformations}, i.e.\ ``Is there such a set $\Gamma$?''} For this reason, we take a closer look at fragments obtained by restricting $s$ and $r$. In practical applications, such as for the data from CS education research sketched in the introduction, $s$ and $r$ can be chosen to be very small. For instance, solving the above problem even when fixing the number of steps to $s=1$ would help to identify common types of mistakes. Unfortunately, the learning problem remains \NP-hard even for small numbers $s$ of steps and small numbers $r$ of transformations to learn.

\begin{theorem}\label{theorem:hardness}
	\algorithmicProblem{LearningTreeTransformations} is $\NP$-complete when 
        \begin{enumerate}
            \item fixing  $s \df 1$; or 
            \item fixing both $s \df 3$ and $r \df 2$. 
        \end{enumerate}
\end{theorem}

Membership in \NP is straightforward for these cases. We note that membership in \NP for the general problem, i.e.\ for arbitrary $s$ and $r$, is less clear because intermediate trees can become exponentially large. Since $s$ and $r$ are small in our applications, we leave the general question for future work. 

We prove the hardness of the two cases of \cref{theorem:hardness} in \cref{proposition:learning-hard-s-1,proposition:learning-hard-s-3-m-2}.

\begin{proposition}\label{proposition:learning-hard-s-1}
	\algorithmicProblem{LearningTreeTransformations} is \NP-hard, even when all examples are binary trees with labels from a binary alphabet and when $s \df 1$ is fixed. 
\end{proposition}

\begin{proofsketch}
	We start by proving hardness for a large label alphabet $\Sigma$. It can be adapted to binary alphabets by encoding single labelled nodes as trees whose sequence of leaves represents the original label. 

	We reduce from the \NP-complete problem \algorithmicProblem{VertexCover} where, given a graph $G$ with nodes $V = \{v_1, \ldots, v_n\}$ and edges $E = \{e_1, \ldots, e_m\}$ as well as a natural number $k$, one asks for a set of at most $k$ nodes from $V$ such that each edge from $E$ is incident to at least one of these nodes (cf. \cite{GareyJ1979}). Such a set of $k$ nodes is called a \emph{$k$-vertex cover}. 
	
 	\proofstep{Construction.} The idea for our reduction $f$ is to construct from a \algorithmicProblem{VertexCover} instance $(G, k)$, an instance of \algorithmicProblem{LearningTreeTransformations} with $s \df 1$, $r \df k$, and one example pair $ (t_{u,v}, t_{u,v}^\star) $ of trees for each  edge $ e = (u, v) $. The goal is that each $k$-vertex cover $S$ of $G$ corresponds to a solution set $\Gamma_S$ of transformations of size $k$ that allows to transform each $t_{u,v}$ into $t_{u,v}^\star$ in one step.

	Let $\ell$ be minimal such that $n \leq 2^\ell$. For each edge $ e = (u, v) $, construct the pair $ (t_{u,v}, t_{u,v}^\star) $ such that $t_{u,v}$ is the full binary tree of depth $\ell$ and $ t_{u,v}^\star $ is a single node. We associate the first $n$ leaves (from left to right) of $ t_{u,v} $ with the nodes $v_1, \ldots, v_n$ of the graph. In $t_{u,v}$, the leaves corresponding to $ u $ and $ v $ are labelled with $ \ell_{u, v} $ and all other nodes are labelled with $ b $. In $t_{u, v}^\star$, the single node is labelled with $ \ell_{u, v} $. This construction is demonstrated in \cref{ex:s-one-t-in-hardness}.

	The reduction $f$ is clearly computable in polynomial time.
	
	\proofstep{Correctness.} Suppose first that $(G, k)$ is a positive \algorithmicProblem{VertexCover} instance with vertex cover $S = \{ v_{c_1}, \dots, v_{c_k} \}$. We construct a solution set $\Gamma_S$ of transformations for $f(G, k)$. For each $v_{c_i}$, the set $\Gamma_S$ contains one transformation $\rho_i\colon \sigma \rightsquigarrow \sigma^\star_i$ where
	\begin{itemize}
	 \item $\sigma$ is the full binary tree of depth $\ell$, whose first $n$ leaves are labelled with node variables $x_1, \ldots, x_n$ and all other nodes are labelled with $b$;  and 
		\item $\sigma^\star_i$ consists of a single node labelled with the node variable $x_{c_i}$.
	\end{itemize}
	The set $\Gamma_S$ of transformations is of size $k$ and allows to transform each $t_{u, v}$ into  $t_{u, v}^\star$ in one step. The latter can be seen as follows. The leaves of $ t_{u, v} $ corresponding to $ u $ and $ v $ are labelled with $ \ell_{u, v} $, while all other nodes have label $ b $. Since $ S $ is a vertex cover of $ G $, we have that $ u \in S $ or $ v \in S $, i.e.\ there exists $i$ such that $u = v_{c_i}$ or $v = v_{c_i}$. Hence, applying the transformation $ \rho_i$ to $ t_{u, v} $ yields a single node labelled $\ell_{u, v}$, that is, $t_{u, v}^\star $.

    Suppose now that $f(G, k)$ is a positive \algorithmicProblem{LearningTreeTransformations} instance with solution set $\Gamma = \{\rho_1, \ldots, \rho_k\}$ of $k$ transformations. We  construct a $k$-vertex cover $S_\Gamma$ from $\Gamma$. To this end we show that 
    if a set $\Gamma$ of size $k$ solves $f(G, k)$, then its transformations must be of a certain form. From transformations of this form we then extract a $ k $-vertex cover of~$ G $. %

    We observe that due to the structural differences between source and target trees in all pairs $(t, t^\star)$, a transformation $ \rho \colon \ttrafo{\sigma}{\sigma^\star}$ can transform $t$ into $t^\star$ only, if
    \begin{alphaenumerate}
        \item\label{i:sigma_covers_t} the root of $\sigma$ is matched to the root of $t$ (because the whole tree $t$ must be replaced by the single node tree $t^\star$); and
        \item $ \sigma^\star $ is a single node (because $t^\star$ only has one node) and one of the following cases applies
        \begin{romanenumerate}
            \item\label{i:constant_label_head} $\sigma^\star = \ell_{u,v}$ for some edge $(u,v)$;
            \item\label{i:node_variable_head} $\sigma^\star = x$ for some node variable $x$ which occurs in $\sigma$ at the end of a path of length $\ell$; or
            \item\label{i:tree_variable_head} $\sigma^\star = Y$ for some tree variable $Y$ which occurs in $\sigma$ at the end of a path of length $\ell$.
        \end{romanenumerate}
    \end{alphaenumerate}
    The condition on $\sigma$ in Cases (b)(ii) and (b)(iii) comes from the shape of our examples: all second components $t^\star$ of pairs are single nodes labelled with some $\ell_{u,v}$. Nodes with such a label occur only at leaves of trees $t$ in first components.  We note that in these cases, the leftmost path to $x$ or $Y$ in $\sigma$ uniquely determines a leaf to which $x$ is matched when $\sigma$ is matched to the full binary tree of depth $\ell$; denote the node of $G$ corresponding to this leaf by $v_{\sigma, x}$ or $v_{\sigma, Y}$, respectively.

		The $k$-vertex cover $S_\Gamma$ of $G$ is constructed from $\Gamma$ as follows. For each $ \rho \colon \ttrafo{\sigma}{\sigma^\star} \in \Gamma $ that explains some example pair $(t, t^\star)$, we include exactly one node in $S_\Gamma$, depending on $\sigma^\star$:
		\begin{itemize}
		 \item if $\sigma^\star = \ell_{u,v}$ for some edge $(u,v)$, then include $u$ in $S_\Gamma$;
		 	\item if $\sigma^\star = x$ for some node variable, then include $v_{\sigma, x}$ in $S_\Gamma$; and
			\item if $\sigma^\star = Y$ for some tree variable $Y$, then include $v_{\sigma, Y}$  in $S_\Gamma$.
		\end{itemize}
		To see that $S_\Gamma$ is indeed a vertex cover of $ G $, consider an arbitrary edge $e = (u, v) \in E$. Suppose that $ \rho\colon\ttrafo{\sigma}{\sigma^\star} \in \Gamma$ transforms $t_{u,v}$ into $t_{u,v}^\star$ (such a $\rho$ exists as $\Gamma$ solves $f(G, k)$). For each of the three cases (i)--(iii) of what $\sigma^\star$ can look like, a suitable node is included in $S_\Gamma$. %

		This concludes the hardness proof for \algorithmicProblem{LearningTreeTransformations} with $s=1$.%
\end{proofsketch}

The following example illustrates the construction from the previous proof.

\begin{example}\label{ex:s-one-t-in-hardness}
    Consider the positive \algorithmicProblem{VertexCover} instance with 
    \begin{center}
        \toplabelled{$ G $}
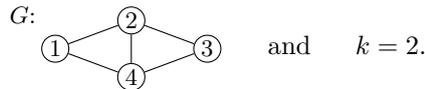
{%
            \tikz[scale=0.5, baseline=1mm, yscale=0.75]{
                \node[narrownode] (u1) at (0,1) {$ 1 $};
                \node[narrownode] (u2) at (2,2) {$ 2 $};
                \node[narrownode] (u3) at (4,1) {$ 3 $};
                \node[narrownode] (u4) at (2,0) {$ 4 $};
                \path (u1) edge (u2);
                \path (u1) edge (u4);
                \path (u2) edge (u3);
                \path (u2) edge (u4);
                \path (u3) edge (u4);
            }
        }
         $\quad$and $\quad$ $ k = 2 $.
    \end{center}

    We construct a \algorithmicProblem{LearningTreeTransformations} instance with $s=1$, $r=2$, and the following pairs $ (t_{1,2}, t_{1,2}^\star), (t_{1,4}, t_{1,4}^\star) (t_{2,3}, t_{2,3}^\star), (t_{2,4}, t_{2,4}^\star), (t_{3,4}, t_{3,4}^\star) $  of trees, one for each edge:

    \vspace{0.4cm}
    \noindent
    \scalebox{0.69}
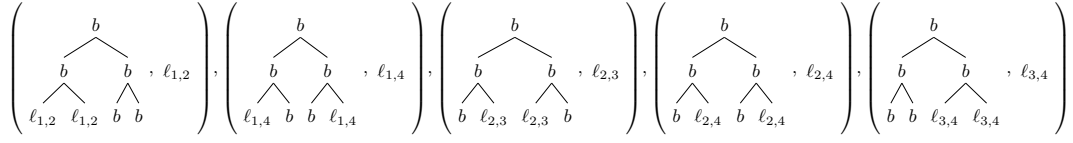
{
    \[
    \left(%
        \ltree{%
            \Tree[.\ensuremath{b} 
                        [.\ensuremath{b} \ensuremath{\ell_{1,2}} \ensuremath{\ell_{1,2}} ]
                        [.\ensuremath{b} \ensuremath{b} \ensuremath{b} ]
                    ]
        }, 
        \ltree{%
            \Tree[.\ensuremath{\ell_{1,2}} ]
        }
    \right)\!,
    \left(%
        \ltree{%
            \Tree[.\ensuremath{b} 
                    [.\ensuremath{b} \ensuremath{\ell_{1,4}} \ensuremath{b} ]
                    [.\ensuremath{b} \ensuremath{b} \ensuremath{\ell_{1,4}} ]
                ]
        }, 
        \ltree{%
            \Tree[.\ensuremath{\ell_{1,4}} ]
        }
    \right)\!,
    \left(%
        \ltree{%
            \Tree[.\ensuremath{b} 
                    [.\ensuremath{b} \ensuremath{b} \ensuremath{\ell_{2,3}} ]
                    [.\ensuremath{b} \ensuremath{\ell_{2,3}} \ensuremath{b} ]
                ]
        }, 
        \ltree{%
            \Tree[.\ensuremath{\ell_{2,3}} ]
        }
    \right)\!,
    \left(%
        \ltree{%
            \Tree[.\ensuremath{b} 
                    [.\ensuremath{b} \ensuremath{b} \ensuremath{\ell_{2,4}} ]
                    [.\ensuremath{b} \ensuremath{b} \ensuremath{\ell_{2,4}} ]
                ]
        }, 
        \ltree{%
            \Tree[.\ensuremath{\ell_{2,4}} ]
        }
    \right)\!,
    \left(%
        \ltree{%
            \Tree[.\ensuremath{b} 
                    [.\ensuremath{b} \ensuremath{b} \ensuremath{b} ]
                    [.\ensuremath{b} \ensuremath{\ell_{3,4}} \ensuremath{\ell_{3,4}} ]
                ]
        }, 
        \ltree{%
            \Tree[.\ensuremath{\ell_{3,4}} ]
        }
    \right)
    \]}
    \vspace{0.4cm}

    The 2-vertex cover $ \{ 2,4 \} $ of $G$ corresponds to the following two tree pattern transformations that explain all pairs of trees in one step:
    \begin{center}
        \toplabelled{$ \rho_{2} $}{%
            \trafo{%
                \ltree{%
                    \Tree[.\ensuremath{b} 
                    [.\ensuremath{b} \ensuremath{x_1} \ensuremath{x_2} ]
                    [.\ensuremath{b} \ensuremath{x_3} \ensuremath{x_4} ]
                    ]
                }
            }{%
                $ x_2 $
            }
        }
        $\quad$and$\quad$
        \toplabelled{$ \rho_4 $}{%
            \trafo{%
                \ltree{%
                    \Tree[.\ensuremath{b} 
                    [.\ensuremath{b} \ensuremath{x_1} \ensuremath{x_2} ]
                    [.\ensuremath{b} \ensuremath{x_3} \ensuremath{x_4} ]
                    ]
                }
            }{%
                \ensuremath{x_4}
            }
        }
    \end{center}

    The transformation $ \rho_2 $ explains $ (t_{1,2}, t_{1,2}^\star), (t_{2,3}, t_{2,3}^\star) $ and $ (t_{2,4}, t_{2,4}^\star) $ in one step, while $
    \rho_4 $ explains $ (t_{1,4}, t_{1,4}^\star), (t_{2,4}, t_{2,4}^\star) $ and $ (t_{3,4}, t_{3,4}^\star) $, also in one step. So the set $
    \Gamma = \{ \rho_2, \rho_4 \} $ explains all pairs of examples in one step. \qed
\end{example}

\begin{proposition}\label{proposition:learning-hard-s-3-m-2}
	\algorithmicProblem{LearningTreeTransformations} is \NP-hard, even when all examples are binary trees and when fixing both $s \df 3$ and $r \df 2$. 
\end{proposition}
	
 \newcommand{\vtree}[3]{%
	\scalebox{0.7}{
        \tikz[every node/.style = {align=center}, baseline=-2.5mm]{
            \node (A) at (0,0){#1};
            \node (B) at (0.2,-0.3){#2};
            \node (C) at (-0.2,-0.3){#3};
           
        }
        }
    }
    
\begin{proofsketch}
	We reduce from the $\NP$-hard problem \algorithmicProblem{3SAT} where, given a propositional formula $\varphi$ in conjunctive normal form with clauses $\calC \df \{C_1, \dots, C_m\}$ with at most three literals and variables $P \df \{p_1, \ldots, p_n\}$, one asks for a satisfying assignment for the variables. We assume, without loss of generality, that each clause has exactly three literals, each variable occurs at most once per clause, and that $n$ is sufficiently large.
	
	\proofstep{Construction.} The idea for our reduction $f$ is to construct from a \algorithmicProblem{3SAT}  instance $\varphi$, a \algorithmicProblem{LearningTreeTransformations} instance with $s \df 3$, $r \df 2$ and a set $\calT = \calT_\calC \uplus \calT_\calS$ of examples. The set $\calT_\calC$ contains one example pair $(t_C, t_C^\star)$ for each clause $C$ of $\varphi$; the set $\calT_\calS$ contains few additional example pairs that ensure that solution transformations have a certain structure. Intuitively, each satisfying assignment $\alpha$ of $\varphi$ will correspond to a solution set $\Gamma_\alpha$ of tree pattern transformations containing one tree transformation $\rho_\alpha$ that \emph{almost} explains all pairs in $\calT_\calC$ in one step and one additional  transformation which helps to \emph{fully} explain these pairs in at most two further steps.

	We explain the example pairs in $\calT$ next, and start with the pairs in $\calT_\calC$. For each clause $C \in \calC$, the trees $t_C$ and $t_C^\star$ in the example pair $(t_C, t_C^\star)$ have the following shape
        \begin{align*}%
            \left(%
                \scalebox{0.9}{\toplabelled{$ t_{C} $}{%
                    \lltree{%
                        \tikzset{level 4+/.style={sibling distance=28pt}}
                        \Tree[.\node(root) {\ensuremath{d}}; 
                            [.\node (dottedStart) {\ensuremath{d}};
                                \edge[draw=none]{};
                                [.\node (dottedEnd) {\ensuremath{d}};
                                    [.\node (000) {\ensuremath{d}};
                                        \ensuremath{a}
                                        \edge[draw=none]{};
                                        \node (pn) {\ensuremath{t_{C,p_n}}};
                                    ]
                                    \edge[draw=none]{};
                                    \node (pn-1) {\,\ensuremath{t_{C,p_{n-1}}}};
                                ]
                                \edge[draw=none]{};
                                \node (p2) {\ensuremath{t_{C,p_2}}};
                            ] 
                            \edge[draw=none]{};
                            \node (p1) {\ensuremath{t_{C,p_1}}};
                        ]
                        \coordinate (dottop) at ($(dottedStart.south west) + (1pt,-2pt)$);
                        \coordinate (dotbot) at ($(dottedEnd.north east) + (-1pt,-1pt)$);
                        \coordinate (rootbr) at ($(root.south east) + (-5.7pt,0pt)$);
                        \coordinate (dotrtop) at ($(p2.south west) + (-3pt,-2pt)$);
                        \coordinate (dotrbot) at ($(pn-1.north east) + (-11pt,6pt)$);
                        \coordinate (0br) at ($(dottedStart.south east) + (-5.7pt,0pt)$);
                        \coordinate (00br) at ($(dottedEnd.south east) + (-5.7pt,0pt)$);
                        \coordinate (000br) at ($(000.south east) + (-5.7pt,0pt)$);
                        \coordinate (p1t) at ($(p1.north) + (0pt,5pt)$);
                        \coordinate (p1tl) at ($(p1.north west) + (0pt,0pt)$);
                        \coordinate (p1tr) at ($(p1.north east) + (0pt,0pt)$);
                        \coordinate (p1bl) at ($(p1.south west) + (0pt,0pt)$);
                        \coordinate (p1br) at ($(p1.south east) + (0pt,0pt)$);
                        \coordinate (p2t) at ($(p2.north) + (0pt,5pt)$);
                        \coordinate (p2tl) at ($(p2.north west) + (0pt,0pt)$);
                        \coordinate (p2tr) at ($(p2.north east) + (0pt,0pt)$);
                        \coordinate (p2bl) at ($(p2.south west) + (0pt,0pt)$);
                        \coordinate (p2br) at ($(p2.south east) + (0pt,0pt)$);
                        \coordinate (pn-1t) at ($(pn-1.north) + (0pt,5pt)$);
                        \coordinate (pn-1tl) at ($(pn-1.north west) + (3pt,0pt)$);
                        \coordinate (pn-1tr) at ($(pn-1.north east) + (-3pt,0pt)$);
                        \coordinate (pn-1bl) at ($(pn-1.south west) + (3pt,0pt)$);
                        \coordinate (pn-1br) at ($(pn-1.south east) + (-3pt,0pt)$);
                        \coordinate (pnt) at ($(pn.north) + (0pt,5pt)$);
                        \coordinate (pntl) at ($(pn.north west) + (0pt,0pt)$);
                        \coordinate (pntr) at ($(pn.north east) + (0pt,0pt)$);
                        \coordinate (pnbl) at ($(pn.south west) + (0pt,0pt)$);
                        \coordinate (pnbr) at ($(pn.south east) + (0pt,0pt)$);
                        \path[dotted pattern] (dottop) -- (dotbot);
                        \path[draw, rounded corners=1.5pt] (p1t) -- (p1tl) -- (p1bl) -- (p1br) -- (p1tr) -- (p1t);
                        \path[draw] (rootbr) -- (p1t);
                        \path[draw, rounded corners=1pt] (p2t) -- (p2tl) -- (p2bl) -- (p2br) -- (p2tr) -- (p2t);
                        \path[draw] (0br) -- (p2t);
                        \path[draw, rounded corners=1pt] (pn-1t) -- (pn-1tl) -- (pn-1bl) -- (pn-1br) -- (pn-1tr) -- (pn-1t);
                        \path[draw] (00br) -- (pn-1t);
                        \path[draw, rounded corners=1pt] (pnt) -- (pntl) -- (pnbl) -- (pnbr) -- (pntr) -- (pnt);
                        \path[draw] (000br) -- (pnt);
                    }%
                }}, 
                \scalebox{0.9}{\toplabelled{$ t_{C}^\star $}{%
                    \lltree{%
                        \tikzset{level 4+/.style={sibling distance=17pt}}
                        \Tree[.\node(root) {\ensuremath{e}}; 
                            [.\node (dottedStart) {\ensuremath{e}};
                                \edge[draw=none]{};
                                [.\node (dottedEnd) {\ensuremath{e}};
                                    [.\node (000) {\ensuremath{e}};
                                        \edge[draw=none]{};
                                        \node (pn) {\ensuremath{t^\star_{C,p_n}}};
                                        \edge[draw=none]{};
                                        \node (pn-1) {\,\ensuremath{t^\star_{C,p_{n-1}}}};
                                    ]
                                    \edge[draw=none]{};
                                    \node (pn-2) {\,\ensuremath{t^\star_{C,p_{n-2}}}};
                                ]
                                \edge[draw=none]{};
                                \node (p2) {\ensuremath{t^\star_{C,p_2}}};
                            ] 
                            \edge[draw=none]{};
                            \node (p1) {\ensuremath{t^\star_{C,p_1}}};
                        ]
                        \coordinate (dottop) at ($(dottedStart.south west) + (1pt,-2pt)$);
                        \coordinate (dotbot) at ($(dottedEnd.north east) + (-1pt,-1pt)$);
                        \coordinate (rootbr) at ($(root.south east) + (-5.7pt,0pt)$);
                        \coordinate (dotrtop) at ($(p2.south west) + (-3pt,-2pt)$);
                        \coordinate (dotrbot) at ($(pn-2.north east) + (-12pt,4pt)$);
                        \coordinate (0br) at ($(dottedStart.south east) + (-5.7pt,0pt)$);
                        \coordinate (00br) at ($(dottedEnd.south east) + (-5.7pt,0pt)$);
                        \coordinate (000br) at ($(000.south east) + (-5.7pt,0pt)$);
                        \coordinate (000bl) at ($(000.south west) + (5.7pt,0pt)$);
                        \coordinate (p1t) at ($(p1.north) + (0pt,3pt)$);
                        \coordinate (p1tl) at ($(p1.north west) + (0pt,-2pt)$);
                        \coordinate (p1tr) at ($(p1.north east) + (0pt,-2pt)$);
                        \coordinate (p1bl) at ($(p1.south west) + (0pt,0pt)$);
                        \coordinate (p1br) at ($(p1.south east) + (0pt,0pt)$);
                        \coordinate (p2t) at ($(p2.north) + (0pt,3pt)$);
                        \coordinate (p2tl) at ($(p2.north west) + (0pt,-2pt)$);
                        \coordinate (p2tr) at ($(p2.north east) + (0pt,-2pt)$);
                        \coordinate (p2bl) at ($(p2.south west) + (0pt,0pt)$);
                        \coordinate (p2br) at ($(p2.south east) + (0pt,0pt)$);
                        \coordinate (pn-2t) at ($(pn-2.north) + (0pt,3pt)$);
                        \coordinate (pn-2tl) at ($(pn-2.north west) + (3pt,-2pt)$);
                        \coordinate (pn-2tr) at ($(pn-2.north east) + (-3pt,-2pt)$);
                        \coordinate (pn-2bl) at ($(pn-2.south west) + (3pt,0pt)$);
                        \coordinate (pn-2br) at ($(pn-2.south east) + (-3pt,0pt)$);
                        \coordinate (pn-1t) at ($(pn-1.north) + (0pt,3pt)$);
                        \coordinate (pn-1tl) at ($(pn-1.north west) + (3pt,-2pt)$);
                        \coordinate (pn-1tr) at ($(pn-1.north east) + (-3pt,-2pt)$);
                        \coordinate (pn-1bl) at ($(pn-1.south west) + (3pt,0pt)$);
                        \coordinate (pn-1br) at ($(pn-1.south east) + (-3pt,0pt)$);
                        \coordinate (pnt) at ($(pn.north) + (0pt,3pt)$);
                        \coordinate (pntl) at ($(pn.north west) + (0pt,-2pt)$);
                        \coordinate (pntr) at ($(pn.north east) + (0pt,-2pt)$);
                        \coordinate (pnbl) at ($(pn.south west) + (0pt,0pt)$);
                        \coordinate (pnbr) at ($(pn.south east) + (0pt,0pt)$);
                        \path[dotted pattern] (dottop) -- (dotbot);
                        \path[draw, rounded corners=1.5pt] (p1t) -- (p1tl) -- (p1bl) -- (p1br) -- (p1tr) -- (p1t);
                        \path[draw] (rootbr) -- (p1t);
                        \path[draw, rounded corners=1pt] (p2t) -- (p2tl) -- (p2bl) -- (p2br) -- (p2tr) -- (p2t);
                        \path[draw] (0br) -- (p2t);
                        \path[draw, rounded corners=1pt] (pn-2t) -- (pn-2tl) -- (pn-2bl) -- (pn-2br) -- (pn-2tr) -- (pn-2t);
                        \path[draw] (00br) -- (pn-2t);
                        \path[draw, rounded corners=1pt] (pn-1t) -- (pn-1tl) -- (pn-1bl) -- (pn-1br) -- (pn-1tr) -- (pn-1t);
                        \path[draw] (000br) -- (pn-1t);
                        \path[draw, rounded corners=1pt] (pnt) -- (pntl) -- (pnbl) -- (pnbr) -- (pntr) -- (pnt);
                        \path[draw] (000bl) -- (pnt);
                    }%
                }%
                }
            \right)
        \end{align*}
	
	where the subtrees $t_{C, p_i}$ and $t^\star_{C, p_i}$ are called \emph{variable subtrees}. The trees $t_{C, p_i}$ and $t^\star_{C, p_i}$ depend on whether and how the proposition variable $p_i$ occurs in $C$:
	
	\begin{itemize}
	 \item if $p_i$ occurs positively in $C$, then	 
	 $t_{C, p_i} \df $\ltree{%
        \Tree[.\ensuremath{d} \ensuremath{a_{C, p_i}} \ensuremath{b_{C, p_i}} ]
    } $\quad$and$\quad$ $t^\star_{C, p_i}\df$\ltree{%
        \Tree[.\ensuremath{e} \ensuremath{a_{C, p_i}} \ensuremath{b_{C, p_i}} ]
    }
    
    \item if $p_i$ occurs negatively in $C$, then $t_{C, p_i} \df $\ltree{%
        \Tree[.\ensuremath{d} \ensuremath{b_{C, p_i}} \ensuremath{a_{C, p_i}} ]
    } $\quad$and$\quad$ $t^\star_{C, p_i}\df$\ltree{%
        \Tree[.\ensuremath{e} \ensuremath{a_{C, i}} \ensuremath{b_{C, p_i}} ]
    }
    \item if $p_i$ does not occur in $C$, then $t_{C, p_i} \df $\ltree{%
        \Tree[.\ensuremath{d} \ensuremath{a_{C, p_i}} \ensuremath{a_{C, p_i}} ]
    }$\quad$and$\quad$  $t^\star_{C, p_i}\df$\ltree{%
                                \Tree[.\ensuremath{e} \ensuremath{a_{C, p_i}} \ensuremath{a_{C, p_i}} ]
                           }
	\end{itemize}

        Intuitively, the pairs in $\calT_\calC$ can \emph{almost} be explained by one tree pattern transformation $\rho\colon \ttrafo{\sigma}{\sigma^\star}$ with $\sigma$ of the same shape as the trees $t_C$ and $\sigma^\star$ of the same shape as $t^\star_C$. Such a transformation indicates in which of the $t_{C, p_i}$ the children need to be swapped, which essentially encodes a satisfying assignment.
		
	The set $T_\calS$ contains the  additional examples
\[
    \left(%
        \ltree{%
            \Tree[.\ensuremath{h_1} \ensuremath{h_2} \ensuremath{h_3} ]%
        }, 
        \ltree{%
            \Tree[.\ensuremath{h_1} \ensuremath{h_3} \ensuremath{h_2} ]%
        }
    \right) \text{ and }
    \left(%
        \ltree{%
            \Tree[.\ensuremath{h_4} \ensuremath{h_5} \ensuremath{h_6} ]
        }, 
        \ltree{%
            \Tree[.\ensuremath{h_4} \ensuremath{h_6} \ensuremath{h_5} ]
        }
    \right)\text{.}
\]
These two examples can be explained by a tree pattern transformation that swaps the leaves; we will see that such a transformation is also required for solutions.

The construction is demonstrated in \cref{ex:s-three-t-two}. The reduction $f$ is clearly computable in polynomial time.

\proofstep{Correctness.}  Suppose $\varphi$ is a satisfiable $3$-CNF formula with satisfying assignment~$\alpha$. We construct a solution set $\Gamma_\alpha$ of transformations for $f(\varphi)$, containing exactly two transformations that explain all examples in at most three steps. The first tree pattern transformation $\rho_\alpha\colon \sigma_\alpha \rightsquigarrow \sigma^\star_\alpha$ in $\Gamma_\alpha$ encodes the satisfying assignment $\alpha$: it switches the leaves of the variable subtree for variable $p_i$ if and only if $\alpha(p_i) = 0$. Its body and head are of the same form as the trees $t_C$ and $t^\star_C$ with  $t_{C, p_i}$ and $t^\star_{C, p_i}$ replaced by
subtrees $\sigma_{C, p_i}$ and $\sigma^\star_{C, p_i}$ %
which depend on the value $\alpha(p_i)$ as follows:
\begin{itemize}
 \item if $\alpha(p_i) = 0$, then $\sigma_{C, p_i} \df $\ltree{%
        \Tree[.\ensuremath{d} \ensuremath{x_i} \ensuremath{y_i} ]
    } $\quad$ and $\quad$ $\sigma^\star_{C, p_i} \df $\ltree{%
        \Tree[.\ensuremath{e} \ensuremath{y_i} \ensuremath{x_i} ]
    }
 \item if $\alpha(p_i) = 1$, then $\sigma_{C, p_i} \df $\ltree{%
        \Tree[.\ensuremath{d} \ensuremath{x_i} \ensuremath{y_i} ]
    } $\quad$ and $\quad$ $\sigma^\star_{C, p_i} \df $\ltree{%
        \Tree[.\ensuremath{e} \ensuremath{x_i} \ensuremath{y_i} ]
    }
\end{itemize}

\noindent
The second transformation 
\begin{align*}
    \rho_\mathrm{swap}\colon
    \ttrafo{%
        \tree{%
            \Tree[.\ensuremath{x} \ensuremath{y} \ensuremath{z} ]
        }
    }{%
        \tree{%
            \Tree[.\ensuremath{x} \ensuremath{z} \ensuremath{y} ]
        }
    }
\end{align*}
swaps two siblings.

It is easy to check that $\Gamma_S \df \{\rho_\alpha, \rho_\text{swap}\}$ explains all pairs in $\calT$.%

 Suppose now that $f(\varphi)$ is a positive instance of \algorithmicProblem{LearningTreeTransformations} with solution set $\Gamma$ containing two  tree pattern transformations that explain all examples in at most three steps. The idea for obtaining a satisfiable assignment $\alpha$ for $\varphi$ is to first show that the transformations in $\Gamma$ are of a very specific form: one of them has to be a transformation like $ \rho_{\text{swap}} $ that switches two children, and the structure of the other is induced by the shape of the example pairs.
 We then extract $\alpha$ from the latter transformation. We defer the technical proof and several more details to \cref{section:appendix-hardness}.
\end{proofsketch}

The following example illustrates the construction from the previous proof.
\begin{example}\label{ex:s-three-t-two}
    Consider the positive \algorithmicProblem{3SAT} instance
    \[ \varphi = (p_1 \lor \lnot p_2 \lor p_3) \land (p_2 \lor p_3 \lor p_4) \land (\lnot p_1 \lor \lnot p_3 \lor \lnot p_4).\] 
    Following the construction from the proof of \cref{proposition:learning-hard-s-3-m-2}, we construct the pair $(t_{C_1}, t^\star_{C_1})$ for the first clause $C_1 = p_1 \lor \lnot p_2 \lor p_3$:%
    \begin{align*}
        \left(%
            \toplabelled{$ t_{C_1} $}{%
                \ltree{%
                    \Tree[.\ensuremath{d} 
                        [.\ensuremath{d}
                            [.\ensuremath{d}
                                [.\ensuremath{d} 
                                    \ensuremath{a}
                                    [.\ensuremath{d} \ensuremath{a_4^1} \ensuremath{a_4^1} ]
                                ]
                                [.\ensuremath{d} \ensuremath{a_3^1} \ensuremath{b_3^1} ]
                            ]
                            [.\ensuremath{d} \ensuremath{b_2^1} \ensuremath{a_2^1} ]
                        ] 
                        [.\ensuremath{d} \ensuremath{a_1^1} \ensuremath{b_1^1} ]
                    ]
                }
            }, 
            \toplabelled{$ t_{C_1}^\star $}{%
                \ltree{%
                    \Tree[.\ensuremath{e} 
                        [.\ensuremath{e}
                            [.\ensuremath{e}
                                [.\ensuremath{e} \ensuremath{a_4^1} \ensuremath{a_4^1} ]
                                [.\ensuremath{e} \ensuremath{a_3^1} 
                                    [.\ensuremath{b_3^1} 
                                        \edge[draw=none];
                                        \node {};
                                    ] 
                                ]
                            ]
                            [.\ensuremath{e} \ensuremath{a_2^1} \ensuremath{b_2^1} ]
                        ] 
                        [.\ensuremath{e} \ensuremath{a_1^1} \ensuremath{b_1^1} ]
                    ]
                }
            }
        \right)
    \end{align*}
    
    The satisfying assignment $\alpha$ with $\alpha(p_1) = 1$, $\alpha(p_2) = 1$, $\alpha(p_3) = 0$, and $\alpha(p_4) = 0$ corresponds to the tree pattern transformation $\rho_\alpha$ which together with the transformation $\rho_\text{swap}$ from above explains the examples constructed from $\varphi$ in at most three steps:
    
    \begin{align*}
        \toplabelled{$ \rho_\alpha $}{%
        \trafo{%
            \ltree{%
                \Tree[.\ensuremath{d} 
                    [.\ensuremath{d}
                        [.\ensuremath{d}
                            [.\ensuremath{d} 
                                \ensuremath{a}
                                [.\ensuremath{d} \ensuremath{x_4} \ensuremath{y_4} ]
                            ]
                            [.\ensuremath{d} \ensuremath{x_3} \ensuremath{y_3} ]
                        ]
                        [.\ensuremath{d} \ensuremath{x_2} \ensuremath{y_2} ]
                    ] 
                    [.\ensuremath{d} \ensuremath{x_1} \ensuremath{y_1} ]
                ]
            }
        }{%
            \ltree{%
                \Tree[.\ensuremath{e} 
                    [.\ensuremath{e}
                        [.\ensuremath{e}
                            [.\ensuremath{e} \ensuremath{y_4} \ensuremath{x_4} ]
                            [.\ensuremath{e} \ensuremath{y_3} \ensuremath{x_3} ]
                        ]
                        [.\ensuremath{e} \ensuremath{x_2} \ensuremath{y_2} ]
                    ] 
                    [.\ensuremath{e} \ensuremath{x_1} \ensuremath{y_1} ]
                ]
            }
        }
    }
    \end{align*}
    Observe that $ x_i, y_i $ are swapped by $\rho_\alpha$ if and only if $\alpha(p_i)=0$. 
\qed
\end{example}

\subsection{Learning Tree Pattern Transformations: A Pragmatic Approach}
\label{section:sat-solving}
\newcommand{\nodes}{V}
\newcommand{\allLabels}{\Omega}
\newcommand{\phileaf}{\varphi_{\operatorname{leaf}}}
\newcommand{\descpos}[2]{#1 \in \N^* \text{\,s.t.\,} #2#1 \in \nodes}

One approach for dealing with the \NP-hardness of learning tree pattern transformations for small parameters is to employ the power of SAT solvers. We next sketch how the problem can be encoded into propositional formulas. We defer details and an application to data sets from CS education research to \cref{section:SAT-solving}. %

We point out that the approach discussed in the following is rather straightforward. Yet, it is also useful: while theory suggests that learning of tree pattern transformations is computationally hard, the SAT solving approach works well enough for our actual data. 

Suppose we are given %
pairs of trees $ \mathcal{D} \df \{ (t_1, t_1^\star), \dots, (t_n, t_n^\star) \} $ and numbers $s, r \in \N$ and are searching for a set of $ r $ transformations that explains each pair in $\calD$ in at most $ s $ steps. Our encoding closely follows a straightforward non-deterministic algorithm, which proceeds in two steps: first, it guesses (1) a set of $ r $ transformations and (2) a sequence of these transformations for each pair $ (t_i, t_i^\star) $ and how they are applied; second, it verifies that this information indeed represents a solution.

To encode this as an instance of \SAT, we use propositional variables encoding the information (1) and (2). We then construct a propositional formula 
with
the following properties: (a) it is satisfiable if and only if the instance of the tree pattern transformation learning problem it encodes is positive; and (b) from a model of the formula one can construct a set of transformations as well as sequences for each pair certifying that the instance is indeed positive.
For simplicity, we start by only considering single-step instances (i.e.\ instances with $s = 1$). Afterwards, we will outline the additional challenges posed by multiple steps, as well as the adaptions to our encoding that enable us to also support them.

\subparagraph*{Encoding single-step instances}

The basic idea for single-step instances is to encode the syntactical structure of transformations and constrain 
it
in accordance with the example pairs.
Our encoding uses three types of propositional variables: $ \mathrm{body}^j_{v,\lambda} $, $ \mathrm{head}^j_{v,\lambda} $, and $ \mathrm{map}^j_{v, i} $.
Intuitively, the variables $ \mathrm{body}_{v, \lambda}^j $ and $ \mathrm{head}_{v, \lambda}^j $ are used to encode the body and head of the $ j $-th transformation, where them being assigned the value $ 1 $ means that node $ v $ of the body, or head, respectively, of transformation $ j $ has label~$ \lambda $. 
The variables $ \mathrm{map}^j_{v, i} $ are used to indicate that transformation $ j $ transforms $ t_i $ into $ t_i^\star $ in such a way that $ v $ is the root of a valid match of the transformation's body into $ t_i $. 
\Cref{ex:prop-vars} demonstrates the correspondence between the propositional variables and transformations. We use $ \varepsilon $ to denote the root, $ 0 $ for its leftmost child, and so on.

\begin{example}\label{ex:prop-vars}
    A tree pattern transformation and the corresponding propositional variables:
    \begin{center}
        \adjustbox{valign=t}{%
        \toplabelled{$ \rho_1 $}{%
            \trafo{%
                \ltree{%
                    \Tree[.\ensuremath{x_1} [.\ensuremath{x_2} \ensuremath{Y_1} \ensuremath{Y_2} ] \ensuremath{Y_3} ]
                }
            }{%
                \ltree{%
                    \Tree[.\ensuremath{x_2} [.\ensuremath{x_1} \ensuremath{Y_2} \ensuremath{Y_1} ] [.\ensuremath{a} \ensuremath{Y_3} ] ]
                }
            }
        }}%
        \adjustbox{valign=t}
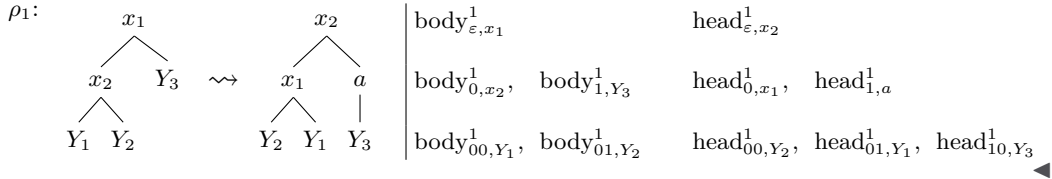
{%
        \begin{tabular}[b]{| l l @{\hspace{0.65cm}} l l l}
            $ \body^1_{\varepsilon, x_1} $ & & $ \head^1_{\varepsilon, x_2} $\\[1em]
            $ \body^1_{0, x_2} $, & $ \body^1_{1, Y_3} $ & $ \head^1_{0, x_1} $, & $ \head^1_{1, a} $\\[1em]
            $ \body^1_{00, Y_1} $, & $ \body^1_{01, Y_2} $
               & $ \head^1_{00, Y_2} $, & $ \head^1_{01, Y_1} $, & $ \head^1_{10, Y_3} $%
        \end{tabular}%
        }
        \qed
    \end{center}
\end{example}

The formula encoding the input instance is a conjunction of multiple formulas that roughly fall into two categories: (i) formulas ensuring syntactically valid transformations, and (ii), formulas ensuring that the chosen transformations do indeed explain all pairs of trees. We provide two sample formulas and refer to the appendix for more details.

\begin{align}
    \bigwedge_{1 \le j \le r}\bigwedge_{\lambda \in \calN \cup \calT} \left( \left(\bigvee_{v \in \nodes} \head^j_{v, \lambda}\right) \to \left(\bigvee_{v' \in \nodes} \body^j_{v', \lambda}\right) \right)
        \label{formula-ex:only-body-vars-in-head}\\
    \bigwedge_{1 \le i \le n} \bigwedge_{1 \le j \le r} \bigwedge_{\text{nodes } v} \left(\map^j_{v, i} \to 
        \bigwedge_{\substack{\text{non-descendants }\\ v' \text{ of } v}} \psi^=_{i, v'}%
    \right)
        \label{formula-ex:only-map-when-contexts-equal}
\end{align}

Formula \eqref{formula-ex:only-body-vars-in-head} is of category (i) and ensures that a variable $ \lambda $ (ranging over all node variables $ \calN $ and tree variables $ \calT $) may only be used in the head of a transformation, if it also occurs in its body. 
The formulas in category (ii) all ``depend'' on the previously mentioned $ \map_{v, i}^j $ variables.
For example, Formula \eqref{formula-ex:only-map-when-contexts-equal} ensures that a transformation may only be applied to a node $ v $ of pair $ i $, if the contexts of the $ i $-th source and target tree of this node are isomorphic. %
We achieve this here by requiring that the labels of $ t_i $ and $ t_i^\star $ are equal at all ``non-descendant positions'' $ v' $ of $ v $. The verification of equal labels is delegated to another simple subformula %
$ \psi_{i, v'}^= $. %
Additional formulas are used to ensure consistent mappings of labels, node variables and tree variables in both body and head.

\subparagraph*{Encoding multi-step instances}
\label{section:multi-step}

We now lift the encoding for single-step instances of the learning tree pattern transformations problem to multi-step instances. 
In contrast to single-step instances, we now need to deal with ``intermediate trees'', i.e.\ trees that are neither source nor target tree of a pair but occur when applying a sequence of transformations to a source tree.
As we do not know what intermediate trees look like in advance, they are encoded by propositional variables as well. Therefore, in addition to the variables we used before, we now also use variables $ \operatorname{int}_{i, k, v, \lambda} $ indicating that node $ v $ of the intermediate tree for pair $ (t_i, t_i^\star) $ after $ k $ steps has label $ \lambda $. 
The formulas from the previous section need to be adapted to these new variables. The variables $ \map_{v, i}^j $ are replaced by variables $ \map_{v, i}^{j,k} $, where $ k \le s $ refers to a step. The intended meaning for this variable is that the $ j $-th transformation is applied to node $ v $ of the intermediate tree of pair $ (t_i, t_i^\star) $ after step $ k-1 $, where intermediate tree $ 0 $ is just $ t_i $. 
The other parts of the formulas are updated accordingly.
For instance, in the single-step case, one formula ensures that a transformation with labels from $ \Sigma $ in its head only explains a pair $ (t_i, t_i^\star) $, if the labels in $ t_i^\star $ match those in the head. Here,
instead of requiring that the label of $ t_i^\star $ must be the same as the one in the head at the corresponding position, this requirement would be stated using the new variables for intermediate trees. The relevant subformula would look as follows:
\begin{align*}
    \map_{v, i}^{j,k} \to \left(\bigwedge_{\descpos{w}{v}} \bigwedge_{\lambda \in \Sigma} \left(\head_{w, \lambda}^j \to \operatorname{int}_{i, k, vw, \lambda}\right)\right)\text{.}
\end{align*}

This states that for transformation $ \rho_j $ to be applicable to pair $ i $ at node $ v $ in step $ k $, a label from $ \Sigma $ in the head of transformation $ \rho_j $ must be identical to the label at the corresponding position in the intermediate tree of pair $ i $ constructed in the previous $ k-1 $ steps.
All the other formulas need to be adapted similarly. Note that the adapted formulas can still be used for the case where $ s = 1 $, because there is a one-to-one correspondence between $ \map_{v, i}^j $ and $ \map_{v, i}^{j,1} $.

\subparagraph*{Dealing with noisy data}

Data can be noisy if, for example, the dataset contains pairs of trees that differ in such a way that no good explanations can be found for their difference. Therefore, it may not always be possible or desirable to find explanations (i.e.\ tree pattern transformations) that explain \textit{all} example pairs, but only most of them. To account for this, we have introduced the possibility of specifying a ratio of the number of pairs of trees that must be explainable for the instance to be positive. This does not affect the complexity of the learning problem, but has proved useful in our practical application of identifying mistakes students made when modelling with propositional logic.

\section{Extending Tree Pattern Transformations}
\label{section:interval-variables}
A tree pattern transformation language for explaining structural differences between trees has to be sufficiently expressive to express typical differences. Yet, it should not be too expressive for two reasons: (1) algorithmic problems such as whether a transformation explains differences betweens two pairs of trees become harder; and (2) learned transformations may overfit the data.   

In this section, we discuss one extension of our transformation language which is natural, but too powerful for our application. One shortcoming of our language is that patterns need to exactly adhere to the structure of the (sub)tree they are matching. A potential extension to circumvent this is to include, besides node and tree variables, also \emph{interval variables} which can match contiguous intervals of children of a node. Such variables loosen the strict adherence to the structure while still yielding good explanations of differences by requiring to ``capture'' everything that can be manipulated. 

\begin{example}\label{ex:motivating-interval-variables}
		Consider the transformation $ \rho\colon
            \scalebox{0.8}{\trafo{%
                \ltree{%
                    \Tree[.\ensuremath{\wedge} \ensuremath{Z_1} \ensuremath{\bot} \ensuremath{Z_2} ]
                }
            }{%
                \ltree{%
                    \Tree[.\ensuremath{\bot} ]
                }
            }
            }$
    with interval variables $Z_1$ and $Z_2$. This transformation encodes the equivalence transformation for propositional formulas which replaces conjunctions containing a ``false'' by a sole ``false'', e.g.\ it can be used to transform the syntax trees of $A \wedge \bot \wedge B$ and $A \wedge B \wedge \bot \wedge C$ into $\bot$.\qed
\end{example}

We formally introduce tree patterns with interval variables before briefly discussing their algorithmic properties and why they are too powerful for learning tree transformations.

\subparagraph*{Tree pattern with interval variables} 
A \emph{tree pattern with interval variables} $\sigma$ is a $\Sigma \uplus \calN \uplus \calT \uplus \calI$-labelled tree, where $\calT$- and $\calI$-labelled nodes must be leaves. The components $\Sigma$, $\calN$, and $\calT$ are as in the definition of tree patterns. The elements of $\calI$ are called \emph{interval variables}. A \emph{match} $\mu$ of a tree pattern $\sigma = (V_\sigma, E_\sigma, \ell_\sigma)$ with interval variables in a tree $t = (V_t, E_t, \ell_t)$ is a mapping $\mu\colon V_\sigma \rightarrow 2^{V_t}$ from nodes of $\sigma$ to sets of nodes of $t$ such that (1) the image sets are disjoint, and (2) $\Sigma \uplus \calN \uplus \calT$-labelled nodes of $\sigma$ are mapped to singletons. Also, additionally to the conditions on tree pattern matches, we have 
\begin{romanenumerate}
    \item[(iv)] interval variables $I$ are mapped to (possibly empty) contiguous intervals, i.e.\ $\mu(I) = \{u_\ell, \ldots, u_{\ell+k-1}\}$ for a contiguous sequence $u_\ell, \ldots, u_{\ell+k-1}$ of children of some node $v$ (called \emph{interval}); and
    \item[(v)] interval variables are mapped consistently, i.e.\ if $u, v \in V_\sigma$ are labelled with the same interval variable, then $|\mu(u)| = |\mu(v)|$ and the subtree sequences of $(t_{u_1}, \ldots, t_{u_k})$ and $(t_{v_1}, \ldots, t_{v_k})$ are isomorphic, where $u$ is mapped to the interval $u_1, \dots, u_k$ and $v$ is mapped to the interval $v_1, \ldots, v_k$ by $\mu$. 
\end{romanenumerate}

Tree pattern transformations with interval variables are defined analogously to tree pattern transformations. 

An example use case is to ``simulate'' the operations considered for tree edit distance:

\begin{example}\label{ex:ted-simulation}
    Tree pattern transformations with interval variables can easily express the three usual tree edit distances operations $ \mathrm{relabel} $, $ \mathrm{insert} $ and $ \mathrm{delete} $ %
    \cite{Bille05}:

    \begin{itemize}
        \item Relabelling an $ a $-labelled node to a $ b $-labelled node: 
            \trafo{%
                \ltree{%
                    \Tree[.\ensuremath{a} \ensuremath{Z_1} ]
                }
            }{%
                \ltree{%
                    \Tree[.\ensuremath{b} \ensuremath{Z_1} ]
                }
            }
        \item Deleting an inner $a$-labelled node:
            \trafo{%
                \ltree{%
                    \Tree[.\ensuremath{x} \ensuremath{Z_1} [.\ensuremath{a} \ensuremath{Z_3} ] \ensuremath{Z_2} ]
                }
            }{%
                \ltree{%
                    \Tree[.\ensuremath{x} \ensuremath{Z_1} \ensuremath{Z_3} \ensuremath{Z_2} ]
                }
            }
        \item Inserting an $ a $-labelled node:
            \trafo{%
                \ltree{%
                    \Tree[.\ensuremath{x_1} \ensuremath{Z_1} \ensuremath{Z_2} \ensuremath{Z_3} ]
                }
            }{%
                \ltree{%
                    \Tree[.\ensuremath{x_1} \ensuremath{Z_1} [.\ensuremath{a} \ensuremath{Z_2} ] \ensuremath{Z_3} ]
                }
            }
    \end{itemize}%
    Note that these edits could also be expressed without interval variables. However, a different transformation would be needed for every length of matched child sequence(s).
    \qed
\end{example}

\subparagraph*{Algorithmic properties}
\label{section:interval:algorithmic}

Unfortunately, already testing whether a tree pattern transformation with interval variables transforms a tree $t$ into a tree $t^\star$ is $\NP$-hard.
This can be seen by reducing from the following $\NP$-hard problem from \cite{Angluin80}. A \emph{string pattern} $p$ is a string from $\Sigma \cup \calX$ where $\calX$ is a set $\{x_1, x_2, \ldots\}$ of variables. The language $L(p)$ of a string pattern contains strings $w$ which are obtained from $p$ by replacing variables ``consistently'' by strings from $\Sigma^+$. Here, consistently means that all occurrences of a variable $x \in X$ are replaced by the same string. 

\algorithmicProblemDescription{StringPatternMembership}
        {
            String $ s \in \Sigma^* $, a string pattern $ p $.
        }
        {
            Is $ s \in L(p) $?
        }

For instance, $(s, p)$ with $s = 001110110$ and $\rho = 0 x_1 1 x_1 0$ is a positive instance of \algorithmicProblem{StringPatternMembership}, as $x_1$ can be replaced by $011$ to yield $s$. The problem 
\algorithmicProblem{StringPatternMembership} is \NP-complete \cite[Theorem 3.6]{Angluin80}.

\begin{proposition}\label{thm:proposition-simple-intervalvariableexplanation}
	Testing whether a tree pattern transformation with interval variables transforms a tree $t$ into a tree $t^\star$ is $\NP$-complete.
\end{proposition}

\begin{proofsketch}
    Membership in \NP is straightforward. To show hardness, 
    we reduce from \algorithmicProblem{StringPatternMembership}. From a string $s$ and a pattern~$p$, we construct trees $t$, $t^\star$ and a tree pattern transformation $\rho\colon \ttrafo{\sigma}{\sigma^\star}$ as follows:
		\begin{itemize}
		 \item $t$ has an $a$-labelled root with children labelled with the symbols of $s$;
		 \item $t^\star$ has an $a$-labelled root with children labelled with the $\Sigma$-labels of $p$;
		 \item $\sigma$ has an $a$-labelled root with children like $p'$, where $p'$ is obtained from $p$ by replacing variables $x_i$ by $y_iZ_i$. Here, $ y_i $ are node variables and $ Z_i $ are interval variables. We need both a node and an interval variable here, because the variables in \algorithmicProblem{StringPatternMembership} are required to be substituted by non-empty strings.
		 \item $\sigma^\star \df t^\star$
		\end{itemize}
                For an illustration of this construction, we refer to the following \cref{ex:simple-intervalexplanation-hardness}.%
\end{proofsketch}

\begin{example}\label{ex:simple-intervalexplanation-hardness}
	The reduction from the proof sketch above constructs, from an instance $(s, p)$ with $s = 001100110$ and $\rho = 0 x_1 1 x_1 0$, the following pair %
        of trees and transformation:
        \begin{align*}
            \left(%
                \ltree{%
                    \Tree[.\ensuremath{a} 0 0 1 1 1 0 1 1 0 ]
                },
                \ltree{%
                    \Tree[.\ensuremath{a} 0 1 0 ]
                }
            \right)
            \quad\text{and}\quad
            \toplabelled{$ \rho' $}{%
                \trafo{%
                    \ltree{%
                        \Tree[.\ensuremath{a} 0 \ensuremath{y_1} \ensuremath{Z_1} 1 \ensuremath{y_1} \ensuremath{Z_1} 0 ]
                    }
                }{%
                    \ltree{%
                        \Tree[.\ensuremath{a} 0 1 0 ]
                    }
                }
            }
        \end{align*}

\vspace{-1em}
\qed
\end{example}

The hardness in the above proof comes from the reuse of interval variables. It is tempting to restrict the reuse of interval variables to circumvent hardness -- also because reuse is often not necessary in practical examples (see \cref{ex:motivating-interval-variables,ex:ted-simulation} above). However, even if each interval variable may only be used once, the problem remains $\NP$-hard. 

\begin{restatable}{theorem}{intervalexplanationproblemhard}
    \label{thm:interval-explanation-problem-hard}
    Testing whether a tree pattern transformation $\rho\colon \ttrafo{\sigma}{\sigma^\star} $ with interval variables transforms a tree $t$ into a tree $t^\star$ is $\NP$-hard, even if each interval variable may occur at most once in $\sigma$ and at most once in $\sigma^\star$.
\end{restatable}

The proof of this theorem is rather technical and can be found in \cref{section:appendix-intervalvars}.%

\subparagraph*{Interval variables are (too) powerful}
\label{section:interval:expressive}

Adding interval variables makes tree pattern transformations very powerful. The following example indicates that for using tree pattern transformations for learning from examples, suitable restrictions need to be found.  The example shows that, given pairs $\{(t_i, t_i^\star)\}_i$, one can easily find two tree pattern transformations with interval variables that can be used to transform each $t_i$ into $t_i^\star$ in two steps. 
\begin{example}\label{ex:number}
    For a set $\{(t_i, t_i^\star)\}_i$ of pairs of trees, the following two transformations, in which $ Y_1 $ is a tree variable and $ Z_1 $ and $ Z_2 $ are interval variables, explain all pairs:

    \begin{center}
        \toplabelled{$ \rho_1$}{%
            \trafo{%
               \ltree{%
                   \Tree[.\ensuremath{Y_1} ]
               }
            }{%
                \ttree{%
                    \Tree[.\ensuremath{a} \ensuremath{t_1^\star} \edge[draw=none];\node[draw=none]{\ensuremath{\cdots}}; \ensuremath{t_n^\star} ]
                }
            }
        } 
        \quad and \quad
        \toplabelled{$ \rho_2$}{%
            \trafo{%
                \ltree{%
                    \Tree[.\ensuremath{a} \ensuremath{Z_1} \ensuremath{Y_1} \ensuremath{Z_2} ]
                }
            }{%
                \ltree{%
                    \Tree[.\ensuremath{Y_1} ]
                }
            }
        }
    \end{center}\qed
\end{example}

Another possible disadvantage of interval variables is that they can lead to a loss of types of differences we want to be able to explain. For example, the same transformation with interval variables can swap as well as not swap the children of a node:
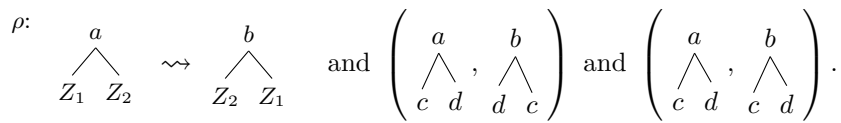
\begin{example}\label{ex:swap-and-not-swap}
    Consider 
    \begin{align*}
        \toplabelled{$ \rho $}{%
            \trafo{%
                \ltree{%
                    \Tree[.\ensuremath{a} \ensuremath{Z_1} \ensuremath{Z_2} ]
                }
            }{%
                \ltree{%
                    \Tree[.\ensuremath{b} \ensuremath{Z_2} \ensuremath{Z_1} ]
                }
            } 
        }
        \text{ and }
        \left(\ltree{%
            \Tree[.\ensuremath{a} \ensuremath{c} \ensuremath{d} ]
        }, 
        \ltree{%
            \Tree[.\ensuremath{b} \ensuremath{d} \ensuremath{c} ]
        }\right) \text{ and }
        \left(\ltree{%
            \Tree[.\ensuremath{a} \ensuremath{c} \ensuremath{d} ]
        }, 
        \ltree{%
            \Tree[.\ensuremath{b} \ensuremath{c} \ensuremath{d} ]
        }\right) \text{.}
    \end{align*}
    Both pairs of trees are explained by $ \rho $ in one step. \qed
\end{example}

While introducing interval variables can be useful, it also has disadvantages. 
The decision whether to include them in the language when learning differences depends on the differences one might expect to learn.
 
\section{Summary and Perspectives}
\label{section:summary}

 This paper takes a first step towards understanding how to algorithmically learn explanations for structural differences between trees. A language for specifying tree transformations based on tree patterns %
 was introduced. It was shown that learning such transformations from examples is computationally hard, even for restricted cases. Towards practical application, we proposed an encoding of the problem as a propositional satisfiability problem. We validated the usefulness of the specification language and the effectiveness of the encoding exemplarily on a dataset for educational tasks for logical modelling obtained with the educational support system \Iltis. 

In the future, we plan to address open theoretical questions as well as explore further applications. 
 
Several theoretical questions remain open regarding properties of (variants of) our tree transformation language. While we saw that already severe restrictions of the learning problem are $\NP$-hard, it is unclear whether the general problem is even in $\NP$. The main issue is that the size of intermediate trees can become exponential in the number of steps, since a transformation with $ m $ nodes can introduce $\bigO(m-1)$ copies of the tree to which it is applied, potentially resulting in trees of size $ \bigO(m^s) $ after $ s $ steps. We conjecture that it is never necessary to construct large intermediate trees and that, therefore, the answer to the following question is positive.
\begin{openproblem}
    Is \algorithmicProblem{LearningTreeTransformations} in \NP?
\end{openproblem}

All restrictions studied here remain $\NP$-hard. We leave open the complexity of the restriction asking  whether there exists a single transformation explaining all given pairs in one step (i.e.\ the case $s \df 1$ and $r \df 1$).

\begin{openproblem}
    Is \algorithmicProblem{LearningTreeTransformations} in \PTIME{} when fixing $ s \df 1 $ and $ r \df 1 $?%
\end{openproblem}

As a first step, we can prove that this case is solvable in polynomial time if all transformations  must be applied at the root node of a tree 
(see \cref{sec:towards-ptime}).

We made several design choices, including injective semantics and patterns that must match the shapes of trees very closely. Studying variants where different choices are made may be insightful as well. 
\begin{openproblem}
	How do slight changes in the definition of tree pattern transformations (e.g.\ injective vs. non-injective semantics) impact algorithmic properties and expressive power?
\end{openproblem}

Besides addressing the theoretical questions, we also plan to explore further applications, among other things, (1) the (empirical) application of this approach to identify typical mistakes in modelling for other formalisms (including further logics, query languages, etc.), (2) the extension of the approach to other educational tasks, such as the identification of mistakes in equivalence transformation tasks, and (3) similar approaches for educational tasks in other domains, such as the formal language domain. These applications are at the borderline of CS education research and theoretical computer science, in particular database theory, but we expect that they raise further interesting theoretical questions.

\bibliography{bibliography}

\begin{thebibliography}{10}

\bibitem{Angluin80}
Dana Angluin.
\newblock Finding patterns common to a set of strings.
\newblock {\em Journal of Computer and System Sciences}, 21(1):46--62, 1980.
\newblock URL:
  \url{https://www.sciencedirect.com/science/article/pii/0022000080900410},
  \href {https://doi.org/https://doi.org/10.1016/0022-0000(80)90041-0}
  {\path{doi:https://doi.org/10.1016/0022-0000(80)90041-0}}.

\bibitem{AvellanedaP18}
Florent Avellaneda and Alexandre Petrenko.
\newblock Inferring {DFA} without negative examples.
\newblock In Olgierd Unold, Witold Dyrka, and Wojciech Wieczorek, editors, {\em
  Proceedings of the 14th International Conference on Grammatical Inference,
  {ICGI} 2018, Wroc{\l}aw, Poland, September 5-7, 2018}, volume~93 of {\em
  Proceedings of Machine Learning Research}, pages 17--29. {PMLR}, 2018.
\newblock URL: \url{http://proceedings.mlr.press/v93/avellaneda19a.html}.

\bibitem{Bille05}
Philip Bille.
\newblock A survey on tree edit distance and related problems.
\newblock {\em Theoretical computer science}, 337(1-3):217--239, 2005.

\bibitem{BojanczykD20}
Mikolaj Bojanczyk and Amina Doumane.
\newblock First-order tree-to-tree functions.
\newblock In Holger Hermanns, Lijun Zhang, Naoki Kobayashi, and Dale Miller,
  editors, {\em {LICS} '20: 35th Annual {ACM/IEEE} Symposium on Logic in
  Computer Science, Saarbr{\"{u}}cken, Germany, July 8-11, 2020}, pages
  252--265. {ACM}, 2020.
\newblock \href {https://doi.org/10.1145/3373718.3394785}
  {\path{doi:10.1145/3373718.3394785}}.

\bibitem{CW16}
Sara Cohen and Yaacov~Y. Weiss.
\newblock The complexity of learning tree patterns from example graphs.
\newblock {\em {ACM} Trans. Database Syst.}, 41(2):14:1--14:44, 2016.
\newblock \href {https://doi.org/10.1145/2890492} {\path{doi:10.1145/2890492}}.

\bibitem{CzerwinskiMNP18}
Wojciech Czerwinski, Wim Martens, Matthias Niewerth, and Pawel Parys.
\newblock Minimization of tree patterns.
\newblock {\em J. {ACM}}, 65(4):26:1--26:46, 2018.
\newblock \href {https://doi.org/10.1145/3180281} {\path{doi:10.1145/3180281}}.

\bibitem{DayFMNS2018}
Joel~D. Day, Pamela Fleischmann, Florin Manea, Dirk Nowotka, and Markus~L.
  Schmid.
\newblock On matching generalised repetitive patterns.
\newblock In Mizuho Hoshi and Shinnosuke Seki, editors, {\em Developments in
  Language Theory}, pages 269--281, Cham, 2018. Springer International
  Publishing.

\bibitem{MouraB08}
Leonardo~Mendon{\c{c}}a de~Moura and Nikolaj~S. Bj{\o}rner.
\newblock {Z3:} an efficient {SMT} solver.
\newblock In C.~R. Ramakrishnan and Jakob Rehof, editors, {\em Tools and
  Algorithms for the Construction and Analysis of Systems, 14th International
  Conference, {TACAS} 2008, Held as Part of the Joint European Conferences on
  Theory and Practice of Software, {ETAPS} 2008, Budapest, Hungary, March
  29-April 6, 2008. Proceedings}, volume 4963 of {\em Lecture Notes in Computer
  Science}, pages 337--340. Springer, 2008.
\newblock \href {https://doi.org/10.1007/978-3-540-78800-3\_24}
  {\path{doi:10.1007/978-3-540-78800-3\_24}}.

\bibitem{GareyJ1979}
Michael~R Garey and David~S Johnson.
\newblock {\em Computers and intractability}, volume 174.
\newblock freeman San Francisco, 1979.

\bibitem{KratochvilK1988}
Jan Kratochvíl and Mirko Křivánek.
\newblock On the computational complexity of codes in graphs.
\newblock {\em Lecture Notes in Computer Science (including subseries Lecture
  Notes in Artificial Intelligence and Lecture Notes in Bioinformatics)}, 324
  LNCS:396 – 404, 1988.
\newblock \href {https://doi.org/10.1007/BFb0017162}
  {\path{doi:10.1007/BFb0017162}}.

\bibitem{LemayMN10}
Aur{\'{e}}lien Lemay, Sebastian Maneth, and Joachim Niehren.
\newblock A learning algorithm for top-down {XML} transformations.
\newblock In Jan Paredaens and Dirk~Van Gucht, editors, {\em Proceedings of the
  Twenty-Ninth {ACM} {SIGMOD-SIGACT-SIGART} Symposium on Principles of Database
  Systems, {PODS} 2010, June 6-11, 2010, Indianapolis, Indiana, {USA}}, pages
  285--296. {ACM}, 2010.
\newblock \href {https://doi.org/10.1145/1807085.1807122}
  {\path{doi:10.1145/1807085.1807122}}.

\bibitem{LemayNG06}
Aur{\'{e}}lien Lemay, Joachim Niehren, and R{\'{e}}mi Gilleron.
\newblock Learning n-ary node selecting tree transducers from completely
  annotated examples.
\newblock In Yasubumi Sakakibara, Satoshi Kobayashi, Kengo Sato, Tetsuro
  Nishino, and Etsuji Tomita, editors, {\em Grammatical Inference: Algorithms
  and Applications, 8th International Colloquium, {ICGI} 2006, Tokyo, Japan,
  September 20-22, 2006, Proceedings}, volume 4201 of {\em Lecture Notes in
  Computer Science}, pages 253--267. Springer, 2006.
\newblock \href {https://doi.org/10.1007/11872436\_21}
  {\path{doi:10.1007/11872436\_21}}.

\bibitem{NeiderG18}
Daniel Neider and Ivan Gavran.
\newblock Learning linear temporal properties.
\newblock In Nikolaj~S. Bj{\o}rner and Arie Gurfinkel, editors, {\em 2018
  Formal Methods in Computer Aided Design, {FMCAD} 2018, Austin, TX, USA,
  October 30 - November 2, 2018}, pages 1--10. {IEEE}, 2018.
\newblock \href {https://doi.org/10.23919/FMCAD.2018.8603016}
  {\path{doi:10.23919/FMCAD.2018.8603016}}.

\bibitem{RoyFN20}
Rajarshi Roy, Dana Fisman, and Daniel Neider.
\newblock Learning interpretable models in the property specification language.
\newblock In Christian Bessiere, editor, {\em Proceedings of the Twenty-Ninth
  International Joint Conference on Artificial Intelligence, {IJCAI} 2020},
  pages 2213--2219. ijcai.org, 2020.
\newblock URL: \url{https://doi.org/10.24963/ijcai.2020/306}, \href
  {https://doi.org/10.24963/IJCAI.2020/306}
  {\path{doi:10.24963/IJCAI.2020/306}}.

\bibitem{RoyGBNXT23}
Rajarshi Roy, Jean{-}Rapha{\"{e}}l Gaglione, Nasim Baharisangari, Daniel
  Neider, Zhe Xu, and Ufuk Topcu.
\newblock Learning interpretable temporal properties from positive examples
  only.
\newblock In Brian Williams, Yiling Chen, and Jennifer Neville, editors, {\em
  Thirty-Seventh {AAAI} Conference on Artificial Intelligence, {AAAI} 2023,
  Thirty-Fifth Conference on Innovative Applications of Artificial
  Intelligence, {IAAI} 2023, Thirteenth Symposium on Educational Advances in
  Artificial Intelligence, {EAAI} 2023, Washington, DC, USA, February 7-14,
  2023}, pages 6507--6515. {AAAI} Press, 2023.
\newblock URL: \url{https://doi.org/10.1609/aaai.v37i5.25800}, \href
  {https://doi.org/10.1609/AAAI.V37I5.25800}
  {\path{doi:10.1609/AAAI.V37I5.25800}}.

\bibitem{SchmellenkampLZ23}
Marko Schmellenkamp, Alexandra Latys, and Thomas Zeume.
\newblock Discovering and quantifying misconceptions in formal methods using
  intelligent tutoring systems.
\newblock In Maureen Doyle, Ben Stephenson, Brian Dorn, Leen{-}Kiat Soh, and
  Lina Battestilli, editors, {\em Proceedings of the 54th {ACM} Technical
  Symposium on Computer Science Education, Volume 1, {SIGCSE} 2023, Toronto,
  ON, Canada, March 15-18, 2023}, pages 465--471. {ACM}, 2023.
\newblock \href {https://doi.org/10.1145/3545945.3569806}
  {\path{doi:10.1145/3545945.3569806}}.

\bibitem{SchmellenkampVZ24}
Marko Schmellenkamp, Fabian Vehlken, and Thomas Zeume.
\newblock Teaching formal foundations of computer science with {Iltis}.
\newblock {\em Educational Column of the Bulletin of {EATCS}}, 2024.
\newblock URL:
  \url{http://bulletin.eatcs.org/index.php/beatcs/article/download/797/842}.

\bibitem{Schwentick07}
Thomas Schwentick.
\newblock Automata for {XML} - {A} survey.
\newblock {\em J. Comput. Syst. Sci.}, 73(3):289--315, 2007.
\newblock URL: \url{https://doi.org/10.1016/j.jcss.2006.10.003}, \href
  {https://doi.org/10.1016/J.JCSS.2006.10.003}
  {\path{doi:10.1016/J.JCSS.2006.10.003}}.

\bibitem{ZhangSS92}
Kaizhong Zhang, Richard Statman, and Dennis~E. Shasha.
\newblock On the editing distance between unordered labeled trees.
\newblock {\em Inf. Process. Lett.}, 42(3):133--139, 1992.
\newblock \href {https://doi.org/10.1016/0020-0190(92)90136-J}
  {\path{doi:10.1016/0020-0190(92)90136-J}}.

\end{thebibliography}

\newpage

\appendix
\section{Additional Material for \cref{section:algorithmic-limitations}: Hardness of Learning Tree Pattern Transformations}\label{section:appendix-hardness}
\subsection{Hardness for One Step With Arbitrary Alphabets}
\label{section:hardness:arbitrary-alphabets}

The proof of Proposition \ref{proposition:learning-hard-s-1} can be adapted to binary alphabets by encoding labels $ \ell_{u,v} $ into binary trees over $\Sigma = \{a, b\}$. Let $\ell$ be minimal such that $n \leq 2^\ell$. Then we encode labels $\ell_{u,v}$ by full binary trees of depth $\ell$, where the $u$-th and $v$-th node are labelled with $a$, and all other nodes are labelled with $b$. The intended transformations are as in the proof for arbitrary alphabet, but use tree variables instead of node variables. The proof itself is then analogous.

We illustrate this construction by adapting Example \ref{ex:s-one-t-in-hardness}.
\begin{example}\label{ex:getRidOfEdgeSpecificLabel}
    The pair 
    \[ \left(%
        \ltree{%
            \Tree[.\ensuremath{b} 
                    [.\ensuremath{b} \ensuremath{\ell_{1,4}} \ensuremath{b} ]
                    [.\ensuremath{b} \ensuremath{b} \ensuremath{\ell_{1,4}} ]
                ]
        }, 
        \ltree{%
            \Tree[.\ensuremath{\ell_{1,4}} ]
        }
    \right) \]
    from Example \ref{ex:s-one-t-in-hardness} is now encoded as 
    \[ \left(%
        \ltree{%
            \Tree[.\ensuremath{b} 
                    [.\ensuremath{b} 
                        [.\ensuremath{b}
                            [.\ensuremath{b} \ensuremath{a} \ensuremath{b} ]
                            [.\ensuremath{b} \ensuremath{b} \ensuremath{a} ]
                        ] 
                        \ensuremath{b} ]
                    [.\ensuremath{b} \ensuremath{b} [.\ensuremath{b}
                            [.\ensuremath{b} \ensuremath{a} \ensuremath{b} ]
                            [.\ensuremath{b} \ensuremath{b} \ensuremath{a} ]
                        ] ]
                ]
        }, 
        \ltree{%
            \Tree[.\ensuremath{b}
                            [.\ensuremath{b} \ensuremath{a} \ensuremath{b} ]
                            [.\ensuremath{b} \ensuremath{b} \ensuremath{a} ]
                        ]
        }
    \right) \]
    and the tree pattern transformations are now as follows
    \begin{center}
        \toplabelled{$ \rho_{2} $}{%
            \trafo{%
                \ltree{%
                    \Tree[.\ensuremath{b} 
                            [.\ensuremath{b} \ensuremath{Y_1} \ensuremath{Y_2} ]
                            [.\ensuremath{b} \ensuremath{Y_3} \ensuremath{Y_4} ]
                        ]
                }
            }{%
               $ Y_2 $
            }
        }
        $\quad$and$\quad$
        \toplabelled{$ \rho_4 $}{%
            \trafo{%
                \ltree{%
                    \Tree[.\ensuremath{b} 
                            [.\ensuremath{b} \ensuremath{Y_1} \ensuremath{Y_2} ]
                            [.\ensuremath{b} \ensuremath{Y_3} \ensuremath{Y_4} ]
                        ]
                }
            }{%
                \ensuremath{Y_4}
            }
        }.
    \end{center}
    
\end{example}

\subsection{Hardness for Two Transformations and Three Steps}
\label{section:hardness:two-transformations}

\subsubsection{Details for the Proof of Proposition \ref{proposition:learning-hard-s-3-m-2}}

\paragraph{From a Satisfying Assignment to a Solution Set of Transformations} The shape of $\rho_\alpha: \sigma_\alpha \rightsquigarrow \sigma^\star_\alpha$ is

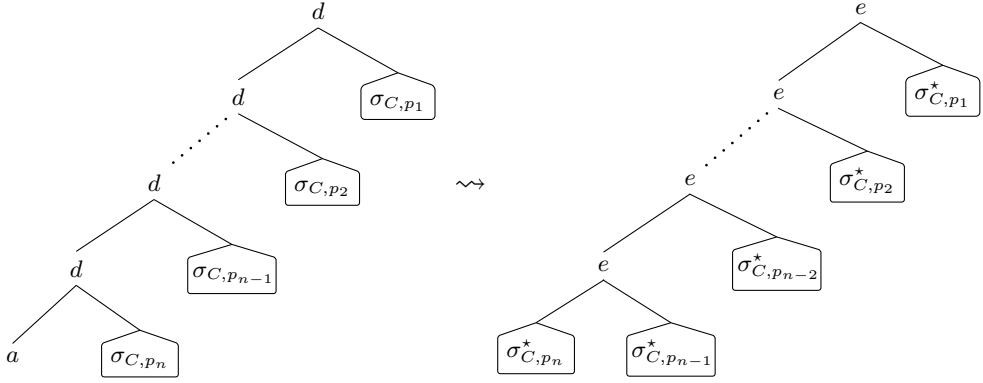
\begin{center}
    \trafo{%
        \lltree{%
            \tikzset{level 4+/.style={sibling distance=28pt}}
            \Tree[.\node(root) {\ensuremath{d}}; 
                [.\node (dottedStart) {\ensuremath{d}};
                    \edge[draw=none]{};
                    [.\node (dottedEnd) {\ensuremath{d}};
                        [.\node (000) {\ensuremath{d}};
                            \ensuremath{a}
                            \edge[draw=none]{};
                            \node (pn) {\ensuremath{\sigma_{C,p_n}}};
                        ]
                        \edge[draw=none]{};
                        \node (pn-1) {\,\ensuremath{\sigma_{C,p_{n-1}}}};
                    ]
                    \edge[draw=none]{};
                    \node (p2) {\ensuremath{\sigma_{C,p_2}}};
                ] 
                \edge[draw=none]{};
                \node (p1) {\ensuremath{\sigma_{C,p_1}}};
            ]
            \coordinate (dottop) at ($(dottedStart.south west) + (1pt,-2pt)$);
            \coordinate (dotbot) at ($(dottedEnd.north east) + (-1pt,-1pt)$);
            \coordinate (rootbr) at ($(root.south east) + (-5.7pt,0pt)$);
            \coordinate (dotrtop) at ($(p2.south west) + (-3pt,-2pt)$);
            \coordinate (dotrbot) at ($(pn-1.north east) + (-12.5pt,6pt)$);
            \coordinate (0br) at ($(dottedStart.south east) + (-5.7pt,0pt)$);
            \coordinate (00br) at ($(dottedEnd.south east) + (-5.7pt,0pt)$);
            \coordinate (000br) at ($(000.south east) + (-5.7pt,0pt)$);
            \coordinate (p1t) at ($(p1.north) + (0pt,5pt)$);
            \coordinate (p1tl) at ($(p1.north west) + (0pt,0pt)$);
            \coordinate (p1tr) at ($(p1.north east) + (0pt,0pt)$);
            \coordinate (p1bl) at ($(p1.south west) + (0pt,0pt)$);
            \coordinate (p1br) at ($(p1.south east) + (0pt,0pt)$);
            \coordinate (p2t) at ($(p2.north) + (0pt,5pt)$);
            \coordinate (p2tl) at ($(p2.north west) + (0pt,0pt)$);
            \coordinate (p2tr) at ($(p2.north east) + (0pt,0pt)$);
            \coordinate (p2bl) at ($(p2.south west) + (0pt,0pt)$);
            \coordinate (p2br) at ($(p2.south east) + (0pt,0pt)$);
            \coordinate (pn-1t) at ($(pn-1.north) + (0pt,5pt)$);
            \coordinate (pn-1tl) at ($(pn-1.north west) + (3pt,0pt)$);
            \coordinate (pn-1tr) at ($(pn-1.north east) + (-3pt,0pt)$);
            \coordinate (pn-1bl) at ($(pn-1.south west) + (3pt,0pt)$);
            \coordinate (pn-1br) at ($(pn-1.south east) + (-3pt,0pt)$);
            \coordinate (pnt) at ($(pn.north) + (0pt,5pt)$);
            \coordinate (pntl) at ($(pn.north west) + (0pt,0pt)$);
            \coordinate (pntr) at ($(pn.north east) + (0pt,0pt)$);
            \coordinate (pnbl) at ($(pn.south west) + (0pt,0pt)$);
            \coordinate (pnbr) at ($(pn.south east) + (0pt,0pt)$);
            \path[dotted pattern] (dottop) -- (dotbot);
            \path[draw, rounded corners=1.5pt] (p1t) -- (p1tl) -- (p1bl) -- (p1br) -- (p1tr) -- (p1t);
            \path[draw] (rootbr) -- (p1t);
            \path[draw, rounded corners=1pt] (p2t) -- (p2tl) -- (p2bl) -- (p2br) -- (p2tr) -- (p2t);
            \path[draw] (0br) -- (p2t);
            \path[draw, rounded corners=1pt] (pn-1t) -- (pn-1tl) -- (pn-1bl) -- (pn-1br) -- (pn-1tr) -- (pn-1t);
            \path[draw] (00br) -- (pn-1t);
            \path[draw, rounded corners=1pt] (pnt) -- (pntl) -- (pnbl) -- (pnbr) -- (pntr) -- (pnt);
            \path[draw] (000br) -- (pnt);
        }
        }{%
        \lltree{%
            \tikzset{level 4+/.style={sibling distance=17pt}}
            \Tree[.\node(root) {\ensuremath{e}}; 
                [.\node (dottedStart) {\ensuremath{e}};
                    \edge[draw=none]{};
                    [.\node (dottedEnd) {\ensuremath{e}};
                        [.\node (000) {\ensuremath{e}};
                            \edge[draw=none]{};
                            \node (pn) {\ensuremath{\sigma^\star_{C,p_n}}};
                            \edge[draw=none]{};
                            \node (pn-1) {\,\ensuremath{\sigma^\star_{C,p_{n-1}}}};
                        ]
                        \edge[draw=none]{};
                        \node (pn-2) {\,\ensuremath{\sigma^\star_{C,p_{n-2}}}};
                    ]
                    \edge[draw=none]{};
                    \node (p2) {\ensuremath{\sigma^\star_{C,p_2}}};
                ] 
                \edge[draw=none]{};
                \node (p1) {\ensuremath{\sigma^\star_{C,p_1}}};
            ]
            \coordinate (dottop) at ($(dottedStart.south west) + (1pt,-2pt)$);
            \coordinate (dotbot) at ($(dottedEnd.north east) + (-1pt,-1pt)$);
            \coordinate (rootbr) at ($(root.south east) + (-5.7pt,0pt)$);
            \coordinate (dotrtop) at ($(p2.south west) + (-3pt,-2pt)$);
            \coordinate (dotrbot) at ($(pn-2.north east) + (-12pt,4pt)$);
            \coordinate (0br) at ($(dottedStart.south east) + (-5.7pt,0pt)$);
            \coordinate (00br) at ($(dottedEnd.south east) + (-5.7pt,0pt)$);
            \coordinate (000br) at ($(000.south east) + (-5.7pt,0pt)$);
            \coordinate (000bl) at ($(000.south west) + (5.7pt,0pt)$);
            \coordinate (p1t) at ($(p1.north) + (0pt,3pt)$);
            \coordinate (p1tl) at ($(p1.north west) + (0pt,-2pt)$);
            \coordinate (p1tr) at ($(p1.north east) + (0pt,-2pt)$);
            \coordinate (p1bl) at ($(p1.south west) + (0pt,0pt)$);
            \coordinate (p1br) at ($(p1.south east) + (0pt,0pt)$);
            \coordinate (p2t) at ($(p2.north) + (0pt,3pt)$);
            \coordinate (p2tl) at ($(p2.north west) + (0pt,-2pt)$);
            \coordinate (p2tr) at ($(p2.north east) + (0pt,-2pt)$);
            \coordinate (p2bl) at ($(p2.south west) + (0pt,0pt)$);
            \coordinate (p2br) at ($(p2.south east) + (0pt,0pt)$);
            \coordinate (pn-2t) at ($(pn-2.north) + (0pt,3pt)$);
            \coordinate (pn-2tl) at ($(pn-2.north west) + (3pt,-2pt)$);
            \coordinate (pn-2tr) at ($(pn-2.north east) + (-3pt,-2pt)$);
            \coordinate (pn-2bl) at ($(pn-2.south west) + (3pt,0pt)$);
            \coordinate (pn-2br) at ($(pn-2.south east) + (-3pt,0pt)$);
            \coordinate (pn-1t) at ($(pn-1.north) + (0pt,3pt)$);
            \coordinate (pn-1tl) at ($(pn-1.north west) + (3pt,-2pt)$);
            \coordinate (pn-1tr) at ($(pn-1.north east) + (-3pt,-2pt)$);
            \coordinate (pn-1bl) at ($(pn-1.south west) + (3pt,0pt)$);
            \coordinate (pn-1br) at ($(pn-1.south east) + (-3pt,0pt)$);
            \coordinate (pnt) at ($(pn.north) + (0pt,3pt)$);
            \coordinate (pntl) at ($(pn.north west) + (0pt,-2pt)$);
            \coordinate (pntr) at ($(pn.north east) + (0pt,-2pt)$);
            \coordinate (pnbl) at ($(pn.south west) + (0pt,0pt)$);
            \coordinate (pnbr) at ($(pn.south east) + (0pt,0pt)$);
            \path[dotted pattern] (dottop) -- (dotbot);
            \path[draw, rounded corners=1.5pt] (p1t) -- (p1tl) -- (p1bl) -- (p1br) -- (p1tr) -- (p1t);
            \path[draw] (rootbr) -- (p1t);
            \path[draw, rounded corners=1pt] (p2t) -- (p2tl) -- (p2bl) -- (p2br) -- (p2tr) -- (p2t);
            \path[draw] (0br) -- (p2t);
            \path[draw, rounded corners=1pt] (pn-2t) -- (pn-2tl) -- (pn-2bl) -- (pn-2br) -- (pn-2tr) -- (pn-2t);
            \path[draw] (00br) -- (pn-2t);
            \path[draw, rounded corners=1pt] (pn-1t) -- (pn-1tl) -- (pn-1bl) -- (pn-1br) -- (pn-1tr) -- (pn-1t);
            \path[draw] (000br) -- (pn-1t);
            \path[draw, rounded corners=1pt] (pnt) -- (pntl) -- (pnbl) -- (pnbr) -- (pntr) -- (pnt);
            \path[draw] (000bl) -- (pnt);
        }
    }
\end{center}

That the set $\Gamma_S$ of transformations constructed in the main part indeed explains all example pairs can be seen as follows. To explain an example pair $(t_{C}, t_{C}^\star)$ in $\calT_\calC$, first apply $\rho_\alpha$ to $t_{C}$. The resulting tree $\rho_\alpha(t_{C})$ already ``almost'' looks like $t_{C}^\star$, except for the variable subtrees corresponding to variables occurring in $C$. Consider the three variable subtrees of $t_{C}$ corresponding to literals from $C$: since $\alpha$ is a satisfying assignment, at least one literal $L$ of $C$ becomes true under $\alpha$. We distinguish two cases: 
\begin{itemize}
 \item if $L$ is positive, and thus $\alpha(x_i)=1$, then $\rho_\alpha$ does not switch the leaves of the variable subtree;
 \item if $L$ is negative, and thus $\alpha(x_i)=0$, then $\rho_\alpha$ switches the leaves of the variable subtree;
\end{itemize}
 In both cases, the variable subtree of $\rho_\alpha(t_{C})$ corresponding to $L$ is as in $t_C^\star$. Thus, at most two other variable subtrees of $\rho_\alpha(t_{C})$ differ from the corresponding variable subtrees in $t_C^\star$. These two subtrees can be fixed by at most two applications of $\rho_\text{swap}$.  The examples in $\calT_\calS$ can be explained by one application of $\rho_\text{swap}$.
 
\paragraph{From Solution Set of Transformations to a Satisfying Assignment}
Here it remained to construct a satisfying assignment $\alpha$ from a solution set $\Gamma$ of tree pattern transformations for $f(\varphi)$.

 We start with two observations:
\begin{enumerate}
    \item There must be a transformation in $\Gamma$ whose head has at least four nodes, if $n$ is larger than a sufficiently large constant. This is required for explaining the examples in $\calT_\calC$,  because in examples $(t_{C}, t^\star_{C})$ in $\calT_\calC$ at least $n$ many $e$-labelled nodes need to be created, where $n$ is the number of variables of $\varphi$.
    \item\label{i:trafo-small-head} There must be (a) a transformation in $\Gamma$ whose head has at most three nodes, and (b) a transformation in $\Gamma$ whose body has at most three nodes. Both are required  for explaining the two examples in $\calT_\calS$, because both components of these examples have only three nodes.
\end{enumerate}

Note that a priori the conditions (2)a and (2)b could be satisfied by two different transformations. We first show: 
\begin{enumerate}
    \item[2'.] There must be a transformation in $\Gamma$ whose head and body each have at most three nodes.
\end{enumerate}

Towards a contradiction, suppose (2') is not true, i.e. none of the transformations has at most three nodes in both body and head. Thus, there is one transformation $ \rho_\downarrow $ with at most three nodes in the head and at least four nodes in the body, and one transformation $ \rho_\uparrow $ with at most three nodes in the body and at least four nodes in the head. Intuitively, applying $ \rho_\downarrow $ ``shrinks'' a tree and $ \rho_\uparrow $ ``blows up'' a tree. As the number of transformations is two, $ \rho_\downarrow $ and $ \rho_\uparrow $ are the only two available transformations. Observe that for each example in $ \calT_\calC $, we need to 
create many nodes with labels of the form $ a_i^j, b_i^j $. Since these labels are distinct for two different examples from $ \calT_\calC $, using them as explicit labels in a transformation would restrict the applicability of that transformation to exactly one example. Therefore, a transformation which needs to be applicable to more than one example, has to ``copy'' these labels from the first component of an example via node or tree variables. Because of the small head size of $ \rho_\downarrow $ and the small body size of $ \rho_\uparrow $, they both can copy at most three different labels. As a result, any sequence of length three of these transformations can only copy few distinct labels, which contradicts the requirement of having to copy a many distinct $ a_i^j, b_i^j $. Hence condition (2') holds.

Without loss of generality, suppose that $\Gamma = \{\rho_1, \rho_2\}$ where $\rho_1\colon \sigma_1 \rightsquigarrow \sigma_1^\star$ is the transformation required due to (1), and $\rho_2\colon \sigma_2 \rightsquigarrow \sigma_2^\star$ is the transformation required due to~(2'). 

Our first goal is to establish the shape of $\rho_2$
\begin{claim}
 The tree pattern transformation $\rho_2$ is of the form %
\scalebox{0.8}{
    \ttrafo{%
        \ltree{%
            \Tree[.\ensuremath{x} \ensuremath{y_1} \ensuremath{y_2} ]
        }
    }{%
        \ltree{%
            \Tree[.\ensuremath{x} \ensuremath{y_2} \ensuremath{y_1} ]
        }
    }
}
where $x$ is a node variable and $y_1, y_2$ are node or tree variables. 
\end{claim}

\begin{proofsketch}
    First we observe that solely $\rho_2$ must explain the examples in $\calT_\calS$, as $ \rho_1 $ cannot be applied to these examples, not even after applying $ \rho_2$. Since both examples in  $\calT_\calS$ use distinct labels, $ \rho_2 $ cannot use any labels from $ \Sigma $ and because in both cases, the root node is the lowest common ancestor of all nodes requiring a change, the body and head of $ \rho_2 $ must have the same structure as these examples. The label at the root of the target trees can only be copied from the root of the source trees, therefore the root of the body and head of $ \rho_2 $ must be labelled with the same node variable, say $ x $. Further, in both examples, the children of the root need to be swapped. This can be achieved by either node or tree variables. 
\end{proofsketch}

Our next goal is to analyse the shape of $ \rho_1 $. We start with an observation. Recall that in examples $(t_C, t_C^\star) \in \calT_\calC$ the source tree has a long $d$-labelled path and the target tree has a long $e$-labelled path. Applying $\rho_2$ to such a tree preserves that there is such a $d$- or $e$-labelled path. We first formalise this property.

A \emph{snake tree} $s = \text{snake}(s_1, \ldots, s_n)$ is a binary tree with a path $\pi = v_1, \ldots, v_n$ such that 
\begin{enumerate}
 \item the tree $s_i$ is one of the two children of $v_i$, for all $i \leq n$; and
 \item the second child of $v_n$ is an $a$-labelled leaf.
\end{enumerate}
A \emph{snake$^\star$ tree} $s^\star = \text{snake}^\star(s^\star_1, \ldots, s^\star_n)$ is a binary tree with a path $\pi = v_1, \ldots, v_{n-1}$ such that 
\begin{enumerate}
 \item the tree $s^\star_i$ is one of the two children of $v_i$, for all $i \leq n-1$; and
 \item the other child of $v_{n-1}$ is $s^\star_n$.
\end{enumerate}
The path $\pi$ is called \emph{long path}. A snake tree is called a $\eta$-snake tree if all nodes on $\pi$ are $\eta$-labelled; likewise for $\eta$-snake$^\star$ trees. Observe that in examples $(t_C, t_C^\star) \in \calT_\calC$, the tree $t_C = \text{snake}(t_1, \ldots, t_n)$ is a $d$-snake tree and the tree $t_C^\star = \text{snake}^\star(t^\star_1, \ldots, t^\star_n)$  is an $e$-snake$^\star$ tree such that the $(t_i, t^\star_i)$'s are pairs of variable subtrees of one of the following forms:
\begin{align*}
    \left(%
        \ltree{%
            \Tree[.\ensuremath{d} \ensuremath{a_{C, i}} \ensuremath{b_{C, i}} ]
        }, 
        \ltree{%
            \Tree[.\ensuremath{e} \ensuremath{a_{C, i}} \ensuremath{b_{C, i}} ]
        }
    \right),
    \left(%
        \ltree{%
            \Tree[.\ensuremath{d} \ensuremath{b_{C, i}} \ensuremath{a_{C, i}} ]
        }, 
        \ltree{%
            \Tree[.\ensuremath{e} \ensuremath{a_{C, i}} \ensuremath{b_{C, i}} ]
        }
    \right) \text{ or }
    \left(%
        \ltree{%
            \Tree[.\ensuremath{d} \ensuremath{a_{C, i}} \ensuremath{a_{C, i}} ]
        }, 
        \ltree{%
            \Tree[.\ensuremath{e} \ensuremath{a_{C, i}} \ensuremath{a_{C, i}} ]
        }
    \right)
\end{align*}

We call $d$-snake trees with such $t_1, \ldots t_n$ \emph{example $d$-snake trees}; likewise $e$-snake$^\star$ trees with such $t^\star_1, \ldots t^\star_n$ are called \emph{example $e$-snake$^\star$ trees}. 

It is easy to see that applying $\rho_2$ to a $d$-snake tree yields a $d$-snake tree; likewise for $e$-snake$^\star$ trees.

We now establish that $\rho_1$ transforms example $d$-snake trees into example $e$-snake$^\star$ trees.
\begin{claim}\label{claim:rho1}
    The tree pattern transformation $ \rho_1 $ is of the form 
    \[\text{snake}(s_1,\ldots, s_n) \rightsquigarrow  \text{snake}^\star(s^\star_1, \ldots, s_n^\star)\] 
    where all nodes on the long path of $\text{snake}(s_1,\ldots, s_n)$ are labelled with $d$ or a node variable, the tree $\text{snake}^\star(s^\star_1, \ldots, s^\star_n)$ is an $e$-snake$^\star$ tree, and all $(s_i, s^\star_i)$ are of one of the following forms
    \begin{enumerate}
    \item     $\left( \ltree{%
    \Tree[.\ensuremath{x_i} \ensuremath{y_{i,1}} \ensuremath{y_{i, 2}} ]
}, \ltree{%
    \Tree[.\ensuremath{e} \ensuremath{y_{i,1}} \ensuremath{y_{i,2}} ]
}
\right)$

     \item $\left( \ltree{%
    \Tree[.\ensuremath{x_i} \ensuremath{y_{i,1}} \ensuremath{y_{i, 2}} ]
}, \ltree{%
    \Tree[.\ensuremath{e} \ensuremath{y_{i,2}} \ensuremath{y_{i,1}} ]
}
\right)$
    \end{enumerate}

with $x$ a node variable or a label, and $y_{i,1}, y_{i,2}$ node or tree variables.
\end{claim}
We call $(s_i, s^\star_i)$ \emph{non-swapping} if they are of form (1) , and \emph{swapping} if they are of form (2).
\begin{proof}[Proof sketch (of Claim \ref{claim:rho1})] We only provide a rough sketch. 
    We know that all target trees of pairs in $ \calT_\calC $ are example $ e $-snake$ ^\star $ trees and all source trees are example $ d $-snake trees. Let $ (t_C, t_C^\star) $ be an arbitrary such example pair. Two of their differences are
    \begin{alphaenumerate}
        \item\label{i:claim-12-d-to-e} there are $ d $s on the long path of $ t_C $ and $ e $s on the long path of $ t_C^\star $, and
        \item\label{i:claim-12-bottom} at the bottom of an example $ d $-snake tree, we have a tree of the form
            \begin{align*}
                \ltree{%
                    \Tree[.\ensuremath{v_{n-1}} [.\ensuremath{v_n} \ensuremath{a} \ensuremath{t_n} ] \ensuremath{t_{n-1}} ]
                }
            \end{align*}
            and at the bottom of an example $ e $-snake$ ^\star $ tree, we have a tree of the form
            \begin{align*}
                \ltree{%
                    \Tree[.\ensuremath{v_{n-1}} \ensuremath{t^\star_n} \ensuremath{t^\star_{n-1}} ]
                }    
            \end{align*}
            Note that the children of $ v_n $ in the first and the children of $ v_{n-1} $ in both might be swapped, but we will just consider the case depicted here, as the others are analogous.
    \end{alphaenumerate}
    Recall that $ \rho_2 $ cannot eliminate either of these two differences. Since we only have one transformation, $ \rho_1 $, beside $ \rho_2 $ available, it must be responsible for the changes required here.

    It remains to argue that $ \rho_1 $'s body cannot be ``shorter'' than a $ d $-snake tree (it cannot be longer because we would not be able to find a match in $ t_C $). Due to (\ref{i:claim-12-bottom}), the ``bottom'' of $ \rho_1 $ has to be of the form
    \begin{center}
        \trafo{%
            \ltree{%
                \Tree[.\ensuremath{x_1} [.\ensuremath{x_2} \ensuremath{x_3} \ensuremath{Y_2} ]  \ensuremath{Y_1} ]
            }
        }{%
            \ltree{%
                \Tree[.\ensuremath{e} \ensuremath{Y_2} \ensuremath{Y_1} ]
            }
        }
    \end{center}
    where $ x_1 $ and $ x_2 $ are node variables or the label $ d $, $ x_3 $ is a node or tree variable, or the label $ a $ and $ Y_1 $ and $ Y_2 $ are tree variables.
    Suppose now that the body of $ \rho_1 $ does not have a long path, but rather a shorter one. Due to (\ref{i:claim-12-d-to-e}), $ \rho_1 $ needs to be applied to the root of $ t_C $ at least once. Then $ x_3 $ from above needs to be a tree variable, because otherwise, there would not be a match in $ t_C $. Since $ x_3 $ is not reused in the head, applying $ \rho_1 $ to the root would eliminate a subtree other than the single node labelled with $ a $. As neither $ \rho_2 $ nor $ \rho_1 $ would be able to restore this subtree but it is required to ``end up'' in $ t_C^\star $, $ \rho_1 $ cannot be of this ``shorter'' form. Therefore, $ \rho_1 $'s body needs to be a snake as claimed which then yields the snake$ ^\star $ head because of $ t_C^\star $.
\end{proof}

We now exploit the shape $\text{snake}(t_1,\ldots, t_n) \rightsquigarrow  \text{snake}^\star(t^\star_1, \ldots, t^\star_n)$ of $\rho_1$ established by Claim \ref{claim:rho1} to define a satisfying assignment $\alpha$ for $\varphi$. Define $\alpha$ by setting
\begin{itemize}
 \item $\alpha(p_i)=0$, if $(s_i, s^\star_i)$ is swapping.
 \item $\alpha(p_i)=1$, if $(s_i, s^\star_i)$ is non-swapping.
\end{itemize}

We now show that $\alpha$ is indeed a satisfying assignment of $\varphi$. 
Suppose towards a contraction that $ \alpha $ is not a satisfying assignment of $ \varphi $. Then there exists a clause $ C $ such that $ \alpha \not \models C $. 

As $ \rho_1 $ needs to be applied to every pair in $ \calT_\calC $, it has to be applied to $ (t_C, t^\star_C) $, the pair corresponding to $ C $. 
We now analyse the influence an application of $ \rho_1 $ has on the variable subtrees of $ t_C $ when applying it to $ t_C $.
Our goal will be to show that $ \rho_1 $ influences all three variable subtrees in a bad way and that every such problem has to be fixed by an application of $ \rho_2 $. Since there only two applications of $ \rho_2 $ available, we cannot fix all problems $ \rho_1 $ created, which means there is no sequence of $ \rho_1 $ and $ \rho_2 $ that explains $ (t_C, t_C^\star) $. This contradicts the fact that the learning tree pattern transformations instance we are considering is positive. 

Because $ \rho_1 $ and $ \rho_2 $ can be applied in an arbitrary order, when applying $ \rho_1 $ to $ t_C $, a variable subtree could either have been changed before, by applying $ \rho_2 $, or not changed, by applying $ \rho_2 $ twice or not at all.
Therefore, we now consider a variable subtree $ g $ of $ t_C $ corresponding to a literal $ L $ in $ C $, and distinguish two cases that can occur before applying $ \rho_1 $ to $ t_C $:

\par\medskip
\noindent
\textbf{Case \texttt{bad}:} $ g $ was not changed, i.e. the transformation $ \rho_2 $ was not applied to $ g $, or it was applied twice. We distinguish between $ L $ being positive or negative.
\begin{romanenumerate}
    \item $ L $ is positive, so $ L = x_j $ for some variable $ x_j $.\\
    Then $ g $ is of the form $\ltree{%
        \Tree[.\ensuremath{d} \ensuremath{a_{C, j}} \ensuremath{b_{C, j}} ]
    }$. Since $ \alpha \not \models C $ and $ x_j $ occurs positively in $ C $, we know that $ \alpha(x_j) = 0 $. By definition of $ \alpha $, $ (s_j, s_j^\star) $ is swapping, i.e. $ \rho_1 $ swaps the children of the variable subtree $ g $.
    \item $ L $ is negative, so $ L = \lnot x_j $.\\
        Here $ g $ is of the form $\ltree{%
            \Tree[.\ensuremath{d} \ensuremath{b_{C, j}} \ensuremath{a_{C, j}} ]
        }$ and $ \alpha(x_j) = 1 $. By definition of $ \alpha $, $ (s_j, s_j^\star) $ is now non-swapping, so the children remain in this order after applying $ \rho_1 $.
\end{romanenumerate}
In both cases, $ g $ has the form $\ltree{%
    \Tree[.\ensuremath{e} \ensuremath{b_{C, j}} \ensuremath{a_{C, j}} ]
}$ after applying $ \rho_1 $. This means that for each occurrence of \textbf{Case \texttt{bad}} we have before applying $ \rho_1 $, we need to apply $ \rho_2 $ after applying $ \rho_1 $ to fix the order of the children.

\par\medskip
\noindent
\textbf{Case \texttt{good}:} $ g $ was changed, i.e. the transformation $ \rho_2 $ was applied once to $ g $. We again distinguish two cases:
\begin{romanenumerate}
    \item $ L $ is positive, so $ L = x_j $ for some variable $ x_j $.\\
        At this point, $ g $ is of the form $\ltree{%
            \Tree[.\ensuremath{d} \ensuremath{b_{C, j}} \ensuremath{a_{C, j}} ]
        }$, even though $ x_j $ occurs positively in $ C $ because the children were swapped by an application of $ \rho_2 $. Recall that $ \alpha \not \models C $ and therefore $ \alpha(x_j) = 0 $. By definition of $ \alpha $, $ \rho_1 $ swaps the children of $ g $, i.e. $ (s_j, s_j^\star) $ is swapping.
    \item $ L $ is negative, i.e. $ L = \lnot x_j $.\\
        Then $ g $ is of the form $\ltree{%
            \Tree[.\ensuremath{d} \ensuremath{a_{C, j}} \ensuremath{b_{C, j}} ]
        }$ and since $ (s_j, s_j^\star) $ is now non-swapping, the two children remain in this order after applying $ \rho_1 $.
\end{romanenumerate}
In both cases, $ g $ has the form $\ltree{%
            \Tree[.\ensuremath{e} \ensuremath{a_{C, j}} \ensuremath{b_{C, j}} ]
        }$ after applying $ \rho_1 $.

We will now use \textbf{Case \texttt{bad}} and \textbf{Case \texttt{good}} from above to establish that if $ \alpha $ does not satisfy clause $ C $, then there is no sequence of $ \rho_1 $ and $ \rho_2 $ that can explain $ (t_C, t_C^\star) $.
Since we started with a positive instance of the learning tree pattern transformations problem, the tree $ t_C $ has to be transformed into the tree $ t_C^\star $ in three steps. This will give us the desired contradiction. 
We have the following possibilities to apply $ \rho_1 $ and $ \rho_2 $ to explain $ (t_C, t_C^\star) $:
\begin{enumerate}
    \item $ \rho_1 $ is applied first: Then we have three occurrences of \textbf{Case \texttt{bad}}, i.e. three variable subtrees in which the order of the children is different than the order of the children in the variable subtrees of $ t_C^\star $. Since at most two applications of $ \rho_2 $ remain, and one application of $ \rho_2 $ can change the order of children of (at most) one variable subtree, this pair cannot be explained by a sequences of transformations starting with $ \rho_1 $.
    \item $ \rho_1 $ is applied after one application of $ \rho_2 $: Then we have at least two occurrences of \textbf{Case \texttt{bad}} (exactly two if $ \rho_2 $ was applied to a variable subtree) but only one application of $ \rho_2 $ remains. Therefore, this pair cannot be explained by a sequence in which $ \rho_2 $ is applied first and $ \rho_1 $ second.
    \item $ \rho_1 $ is applied after two applications of $ \rho_2 $: Then we have at least one occurrence of \textbf{Case \texttt{bad}} but no applications of $ \rho_2 $ left. This excludes the sequence starting with two applications of $ \rho_2 $ and finishing with $ \rho_1 $ as a possible solution.
\end{enumerate}%
Since $ \rho_1 $ has to be applied once to each pair $ (t_{C_i}, t_{C_i}^\star) $, the above three cases cover all possible sequences of transformations to explain such a pair. Since none of these can explain $ (t_C, t_C^\star) $ even though the learning tree pattern transformations instance was positive, we get the desired contradiction. Therefore $ \alpha $ satisfies all clauses, so $ \alpha \models \varphi $. This concludes the proof.

\subsubsection{Continuation of Example \ref{ex:s-three-t-two}}
\begin{example}[Continuation of Example \ref{ex:s-three-t-two}]
    Consider the positive \algorithmicProblem{3SAT} instance
    \[ \varphi = (p_1 \lor \lnot p_2 \lor p_3) \land (p_2 \lor p_3 \lor p_4) \land (\lnot p_1 \lor \lnot p_3 \lor \lnot p_4).\] 
    The pairs $(t_{C_2}, t^\star_{C_2})$ and  $(t_{C_3}, t^\star_{C_3})$ for the clauses $C_2 = p_2 \lor p_3 \lor p_4$ and $C_3 = \lnot p_1 \lor \lnot p_3 \lor \lnot p_4$ are as follows:
    \begin{align*}
        \left(%
            \toplabelled{$ t_{C_2} $}{%
                \ltree{%
                \Tree[.\ensuremath{d} 
                    [.\ensuremath{d}
                        [.\ensuremath{d}
                            [.\ensuremath{d} 
                                \ensuremath{a}
                                [.\ensuremath{d} \ensuremath{a_4^2} \ensuremath{b_4^2} ]
                            ]
                            [.\ensuremath{d} \ensuremath{a_3^2} \ensuremath{b_3^2} ]
                        ]
                        [.\ensuremath{d} \ensuremath{a_2^2} \ensuremath{b_2^2} ]
                    ] 
                    [.\ensuremath{d} \ensuremath{a_1^2} \ensuremath{a_1^2} ]
                ]
            }
            }, 
            \toplabelled{$ t_{C_2}^\star $}{%
                \ltree{%
                \Tree[.\ensuremath{e} 
                    [.\ensuremath{e}
                        [.\ensuremath{e}
                            [.\ensuremath{e} \ensuremath{a_4^2} \ensuremath{b_4^2} ]
                            [.\ensuremath{e} \ensuremath{a_3^2} \ensuremath{b_3^2} ]
                        ]
                        [.\ensuremath{e} \ensuremath{a_2^2} \ensuremath{b_2^2} ]
                    ] 
                    [.\ensuremath{e} \ensuremath{a_1^2} \ensuremath{a_1^2} ]
                ]
                }
            }
        \right)\\
        \left(%
            \toplabelled{$ t_{C_3} $}{%
                \ltree{%
                \Tree[.\ensuremath{d} 
                    [.\ensuremath{d}
                        [.\ensuremath{d}
                            [.\ensuremath{d} 
                                \ensuremath{a}
                                [.\ensuremath{d} \ensuremath{a_4^3} \ensuremath{a_4^3} ]
                            ]
                            [.\ensuremath{d} \ensuremath{b_3^3} \ensuremath{a_3^3} ]
                        ]
                        [.\ensuremath{d} \ensuremath{a_2^3} \ensuremath{a_2^3} ]
                    ] 
                    [.\ensuremath{d} \ensuremath{b_1^3} \ensuremath{a_1^3} ]
                ]
            }
            }, 
            \toplabelled{$ t_{C_3}^\star $}{%
                \ltree{%
                \Tree[.\ensuremath{e} 
                    [.\ensuremath{e}
                        [.\ensuremath{e}
                            [.\ensuremath{e} \ensuremath{a_4^3} \ensuremath{a_4^3} ]
                            [.\ensuremath{e} \ensuremath{a_3^3} \ensuremath{b_3^3} ]
                        ]
                        [.\ensuremath{e} \ensuremath{a_2^3} \ensuremath{a_2^3} ]
                    ] 
                    [.\ensuremath{e} \ensuremath{a_1^3} \ensuremath{b_1^3} ]
                ]
            }
            }
        \right)
    \end{align*}
\end{example}

\section{Learning Tree Pattern Transformations in Practice: SAT Solving}
\label{section:SAT-solving}

As sketched in the introduction, our interest in learning tree pattern transformations comes from applications in CS education research. For overcoming the \NP-hardness results shown in \cref{section:learning-hard}, we follow a common strategy in the literature: we encode the learning problem for tree pattern transformations in propositional logic and employ a SAT solver to find solutions.

We first describe our encoding (see Section \ref{section:appendix-encoding}) and then report on our implementation and its prototypical application (see Section \ref{section:appendix-application}).

\subsection{Encoding {\algorithmicProblem{LearningTreeTransformations}} Instances}
\label{section:appendix-encoding}

In this section we describe our encoding. Suppose a \algorithmicProblem{LearningTreeTransformations} instance with examples $ \mathcal{D} \df \{ (t_1, t_1^\star), \dots, (t_n, t_n^\star) \} $ and numbers $s, r \in \N$ is given and we are looking for a solution set of $ r $ transformations that explains each pair in $\calD$ in at most $ s $ steps. Our encoding is inspired by the straightforward \NP algorithm, which proceeds in two steps: first, it guesses (1) a set of $ r $ transformations and (2) a sequence of these transformations for each pair $ (t_i, t_i^\star) $ and how they are applied; second, it verifies that this information indeed represents a solution.

To encode this approach as a \SAT instance, we use propositional variables encoding the information (1) and (2). We then construct a propositional formula $ \Phi_{s,r}^\mathcal{D} $ which verifies that an assignment to the variables is a solution. The formula has the following properties: (a) it is satisfiable if and only if the instance of the tree pattern transformation learning problem it encodes is positive; and (b) from a model of the formula one can construct a set of transformations as well as sequences for each pair certifying that the instance is indeed positive.

Next, we describe the encoding of single-step instances (i.e. instances with $s = 1$); afterwards, we sketch how to lift this to multiple-step instances. 

\subsubsection{Encoding Single-Step Instances}
\label{section:appendix-one-step}
We start with instances with $ s \df 1 $.
The basic idea is to encode the syntactical structure of transformations and constrain the structure in accordance with the example pairs to ensure that they are valid in the sense that a transformation chosen for a pair $ (t_i, t_i^\star) $ does indeed transform $ t_i $ into $ t_i^\star $. 

Since we only consider single-step instances here, we know that the sizes of the bodies and heads of transformations are bounded by the largest source and target trees, respectively. Therefore it is clear in advance which nodes a body or head of a transformation can have. We therefore fix the set $ \nodes = \{ u_1 \cdots u_k \mid u_i \in \{ 0, \dots, d-1 \}, 0 \le k \le h \} $ of these nodes, where $ d $ is the maximal degree and $ h $ is the maximal depth across all pairs of trees.

Let $ \Sigma$ be an alphabet, $\calN $ a set of node variables, and $ \calT $ a set of tree variables and denote $ \allLabels \df \Sigma \uplus \calN \uplus \calT \uplus \{-\}$, where $-$ is a special symbol which will be used to indicate that a node from $\nodes$ is not used in a pattern. For a pair $ (t_i, t_i^\star) $ of trees, denote their labelling functions by $ \ell_{t_i} $ and $ \ell_{t_i^\star} $. 

Our encoding uses three types of propositional variables:
\begin{itemize}
    \item $ \mathrm{body}^j_{v,\lambda} $ for $ \lambda \in \allLabels $,
    \item $ \mathrm{head}^j_{v,\lambda} $ for $ \lambda \in \allLabels $, and %
    \item $ \mathrm{map}^j_{v, i} $ for $ i \in \{ 1, \dots, n \} $,
\end{itemize}%
where $ j \in \{ 1, \dots, r \}$ and $ v \in \nodes $. 
Intuitively, the variables $ \mathrm{body}_{v, \lambda}^j $ and $ \mathrm{head}_{v, \lambda}^j $ are used to encode the body and head of the $ j $-th transformation, where them being assigned the value $ 1 $ means that node $ v $ of the body, or head, respectively, of transformation $ j $ has label~$ \lambda $. The special label $\lambda = -$ indicates that a node is not used. We will often use $ \phileaf(v)$ as a shorthand for $v$ being a leaf, i.e. $v$ having only children labelled with $-$.
The variables $ \mathrm{map}^j_{v, i} $ are used to indicate that transformation $ j $ transforms $ t_i $ into $ t_i^\star $ in such a way that $ v $ is the root of a valid match of the transformation's body into $ t_i $. 
Example \ref{ex:appendix-prop-vars} demonstrates the correspondence between the propositional variables and transformations.

\begin{example}\label{ex:appendix-prop-vars}
    Suppose all trees in the example pairs are binary and that their maximal depth $ h $ is two. Then $ \nodes = \{ \varepsilon, 0, 1, 00, 01, 10, 11 \} $, where $ \varepsilon $ refers to the empty string and therefore corresponds to the root node. Consider the transformation 
    \begin{center}
        \toplabelled{$ \rho_1 $}{%
            \trafo{%
                \ltree{%
                    \Tree[.\ensuremath{x_1} [.\ensuremath{x_2} \ensuremath{Y_1} \ensuremath{Y_2} ] \ensuremath{Y_3} ]
                }
            }{%
                \ltree{%
                    \Tree[.\ensuremath{x_2} [.\ensuremath{x_1} \ensuremath{Y_2} \ensuremath{Y_1} ] [.\ensuremath{a} \ensuremath{Y_3} ] ]
                }
            }
        }.
    \end{center}
    The corresponding propositional variables are the following:

    \begin{center}
        \begin{tabular}[b]{l l l l @{\hspace{0.65cm}} l l l l}
            $ \body^1_{\varepsilon, x_1} $ & & & & $ \head^1_{\varepsilon, x_2} $\\
            $ \body^1_{0, x_2} $, & $ \body^1_{1, Y_3} $ & & & $ \head^1_{0, x_1} $, & $ \head^1_{1, a} $\\
            $ \body^1_{00, Y_1} $, & $ \body^1_{01, Y_2} $, & $ \body^1_{10, -} $, & $ \body^1_{11, -} $ & $ \head^1_{00, Y_2} $, & $ \head^1_{01, Y_1} $, & $ \head^1_{10, Y_3} $, & $ \head^1_{11, -} $
        \end{tabular}%
        \qed
    \end{center}
\end{example}

To ensure that the variables $ \body^j_{v,\lambda} $ and $ \head^j_{v,\lambda} $ encode syntactically valid transformations, we use the formulas listed in Table \ref{table:syntactically-valid-transformations}. Formulas~\eqref{formula:exactly-one-body-label}, \eqref{formula:exactly-one-head-label}, and \eqref{formula:oslash-only-oslash-children} enforce that each node of each body and each head has exactly one label, and that if a node is not used, then neither are its children. 
Formulas~\eqref{formula:tree-var-leaf-in-body} and \eqref{formula:tree-var-leaf-in-head} ensure that nodes labelled with a tree variable only occur as leaves in all bodies and heads and due to Formula~\eqref{formula:only-body-vars-in-head}, all node and tree variables occurring in a head of a transformation must also be present in that transformation's body. 
Let $ \Phi_{s,r}^{\text{syntax}} $ be 
the conjunction of Formulas~\eqref{formula:exactly-one-body-label} to \eqref{formula:only-body-vars-in-head}.

\begin{table}
    \setlength{\jot}{5pt}
    {
    \footnotesize
    \begin{align}
        \bigwedge_{1 \le j \le r}\bigwedge_{v \in \nodes} \left(
            \bigvee_{\lambda \in \allLabels} \body^j_{v, \lambda} \land 
        \bigwedge_{\lambda_1,\lambda_2 \in \allLabels,\,\lambda_1 \neq \lambda_2} \left(\lnot\body^j_{v, \lambda_1} \lor \lnot\body^j_{v, \lambda_2}\right)\right)
        \label{formula:exactly-one-body-label}\\
        \bigwedge_{1 \le j \le r}\bigwedge_{v \in \nodes} \left(
            \bigvee_{\lambda \in \allLabels} \head^j_{v, \lambda} \land 
        \bigwedge_{\lambda_1,\lambda_2 \in \allLabels,\,\lambda_1 \neq \lambda_2} \left(\lnot\head^j_{v, \lambda_1} \lor \lnot\head^j_{v, \lambda_2}\right)\right)
        \label{formula:exactly-one-head-label}\\
        \bigwedge_{1 \le j \le r}\bigwedge_{v \in \nodes} 
            \left(
            \left(\body^{j}_{v, -} \to 
                \bigwedge_{w \in \N \text{\,s.t.\,} vw \in \nodes} \body^j_{vw, -} 
            \right)
            \land
            \left(\head^{j}_{v, -} \to 
                \bigwedge_{w \in \N \text{\,s.t.\,} vw \in \nodes} \head^j_{vw, -} 
            \right)
            \right)%
        \label{formula:oslash-only-oslash-children}\\
        \bigwedge_{1 \le j \le r}\bigwedge_{\lambda \in \calT}\bigwedge_{v \in \nodes} \left(\body^{j}_{v, \lambda} \to \phileaf(v)\right)%
        \label{formula:tree-var-leaf-in-body}\\
        \bigwedge_{1 \le j \le r}\bigwedge_{\lambda \in \calT}\bigwedge_{v \in \nodes} \left(\head^{j}_{v, \lambda} \to \phileaf(v)\right)
        \label{formula:tree-var-leaf-in-head}\\
        \bigwedge_{1 \le j \le r}\bigwedge_{\lambda \in \calN \cup \calT} \left( \left(\bigvee_{v \in \nodes} \head^j_{v, \lambda}\right) \to \left(\bigvee_{v' \in \nodes} \body^j_{v', \lambda}\right) \right)
        \label{formula:only-body-vars-in-head}
    \end{align}
    }
    \caption{Constraints ensuring the syntactic validity of $ r $ transformations.}
    \label{table:syntactically-valid-transformations}
    {
    \footnotesize
    \begin{align}
        \bigwedge_{1 \le i \le n} \bigvee_{1 \le j \le r} \bigvee_{v \in \nodes} \map^j_{v, i}
        \label{formula:mapping-for-each-pair}\\
        \bigwedge_{1 \le i \le n} \bigwedge_{1 \le j \le r} \bigwedge_{v \in \nodes} \left(\map^j_{v, i} \to [\context(t_i, v) \cong \context(t_i^\star, v)]\right)
        \label{formula:only-map-when-contexts-equal}\\
        \bigwedge_{1\le i \le n} \bigwedge_{1 \le j \le r} \bigwedge_{v \in \nodes} \left(\map^j_{v,i} \to 
            \varphi_{\mathrm{labels}}^{i, j, v} \land \varphi_{\mathrm{nodevars}}^{i,j,v} \land \varphi_{\mathrm{treevars}}^{i, j, v} \right)
        \label{formula:consistent-body-labels}
    \end{align}
    }
    where
    {
    \footnotesize
    \begin{align*}
        \varphi_{\mathrm{labels}}^{i, j, v} &\df \bigwedge_{w \in \N^* \text{\,s.t.\,} vw \in \nodes} \bigwedge_{\lambda \in \Sigma} \left(\body_{w, \lambda}^j \to [\ell_{t_i}(vw) = \lambda]\right)\\
        \varphi_{\mathrm{nodevars}}^{i, j, v} &\df
            \bigwedge_{\substack{w_1 \in \N^* \text{\,s.t.\,} vw_1 \in \nodes,\\ w_2 \in \N^* \text{\,s.t.\,} vw_2 \in \nodes\hphantom{,}}}
            \bigwedge_{\lambda \in \calN}
                \left(\left(\body^j_{w_1, \lambda} \land \body^j_{w_2, \lambda}\right) \to [\ell_{t_i}(vw_1) = \ell_{t_i}(vw_2)] \right)\\
        \varphi_{\mathrm{treevars}}^{i, j, v} &\df
            \bigwedge_{\substack{w_1 \in \N^* \text{\,s.t.\,} vw_1 \in \nodes,\\ w_2 \in \N^* \text{\,s.t.\,} vw_2 \in \nodes\hphantom{,}}}
            \bigwedge_{\lambda \in \calT}
                \left(\left(\body^j_{w_1, \lambda} \land \body^j_{w_2, \lambda}\right) \to [\subtree_{t_i}(vw_1) \cong \subtree_{t_i}(vw_2)] \right)
    \end{align*}
    }
    
    {
    \footnotesize
    \begin{align}
        \bigwedge_{1\le i \le n} \bigwedge_{1 \le j \le r} \bigwedge_{v \in \nodes} \left(\map^j_{v,i} \to \bigwedge_{\descpos{w}{v}} 
            \psi_{\mathrm{labels}}^{i, j, v, w} \land \psi_{\mathrm{nodevars}}^{i,j,v,w} \land \psi_{\mathrm{treevars}}^{i, j, v, w} \right)
        \label{formula:consistent-head-labels}
    \end{align}
    }
    where
    {
    \footnotesize
    \begin{align*}
        \psi_{\mathrm{labels}}^{i, j, v, w} &\df \bigwedge_{\lambda \in \Sigma} \left(\head_{w, \lambda}^j \to [\ell_{t_i^\star}(vw) = \lambda]\right)\\
        \psi_{\mathrm{nodevars}}^{i, j, v, w} &\df
            \bigwedge_{\lambda \in \calN}
                \left(\head_{w, \lambda}^j \to \bigvee_{\descpos{w'}{v}} \left(\body_{w', \lambda}^j \land [\ell_{t_i}(vw') = \ell_{t_i^\star}(vw)] \right)\right)\\
        \psi_{\mathrm{treevars}}^{i, j, v, w} &\df
            \bigwedge_{\lambda \in \calT}
                \left(\head_{w, \lambda}^j \to \bigvee_{\descpos{w'}{v}} \left(\body_{w', \lambda}^j \land [\subtree_{t_i}(vw') \cong \subtree_{t_i^\star}(vw)] \right)\right)
    \end{align*}
    }
    \caption{Constraints ensuring correct semantics based on matches induced by the $ \map^j_{v, i} $ variables. Here we denote the subtree of a tree $ t_i $ rooted at a node $ v $ as $ \subtree_{t_i}(v) $, rather than $ t_{i\,v} $, to avoid confusion with indices.}
    \label{table:semantic-constraints}
\end{table}

Given a satisfying assignment of $ \Phi_{s,r}^{\text{syntax}} $, it is straightforward to extract $ r $ syntactically valid transformations from the variables $ \mathrm{body}^j_{v,\lambda} $ and $ \mathrm{head}^j_{v,\lambda} $. To avoid too much clutter, we omit a formal definition and refer the reader to Example~\ref{ex:appendix-prop-vars} again.

Since the formula is yet independent of the examples, the transformations extracted from a model might not be a solution to our instance of the tree pattern transformation learning problem. 
We make the connection to the examples via the formulas in Table \ref{table:semantic-constraints}.
In some formulas we use the construct $ [\mathrm{condition}] $. When the formula is constructed, this is replaced by the constant value of true, if the condition is met, or false otherwise.
Formula~\eqref{formula:mapping-for-each-pair} ensures that for each example pair there exists a transformation as well as a (valid) match. The proper semantics are enforced by Formulas~\eqref{formula:only-map-when-contexts-equal} to \eqref{formula:consistent-head-labels}. Formula~\eqref{formula:only-map-when-contexts-equal} restricts the subtrees to which a transformation can be applied %
by requiring that a transformation may only be applied to a node $ v $ of a tree $ t_i $ if all differences between $ t_i $ and $ t_i^\star $ are restricted to the subtree induced by $ v $. %
Formulas~\eqref{formula:consistent-body-labels} and \eqref{formula:consistent-head-labels} ensure a consistent use of labels in bodies and heads of transformations with respect to matches. Specifically, labels from $ \Sigma $ in the body (head) require the identical label at the corresponding node, according to the match, in the source (target) tree due to $ \varphi_{\mathrm{label}}^{i, j, v} $ ($ \psi_{\mathrm{label}}^{i, j, v} $).
Note that we need to consider nodes other than $ v $, which is the root of a match of the body of transformation $ \rho_j $ in $ t_i $, as indicated by $ \map_{v, i}^j $, to ensure this consistency.
Node $ v $ in $ t_i $ corresponds to the root of the body because of the match. As a result, a node $ w $ of $ \rho_j $'s body corresponds to node $ vw $ of $ t_i $; and analogously for $ \rho_j $'s head and $ t_i^\star $. Therefore we need to ensure consistent labels for nodes $ w $ in the body (head) of $ \rho_j $ and $ vw $ in $ t_i $ ($ t_i^\star $).
Consistence of multiple occurrences of the same node (tree) variable in the body is ensured by $ \varphi_{\mathrm{nodevars}}^{i, j, v} $ ($ \varphi_{\mathrm{treevars}}^{i, j, v} $). Finally, the formulas $ \psi_{\mathrm{nodevars}}^{i, j, v} $ and $ \psi_{\mathrm{treevars}}^{i, j, v} $ enforce that values assigned to node and tree variables based on the match induced by $ \map^j_{v, i} $ are appropriate with regard to the target tree.%

Let us denote the conjunction of Formulas~\eqref{formula:mapping-for-each-pair} to \eqref{formula:consistent-head-labels} and $ \Phi_{s,r}^\text{syntax} $ as $ \Phi_{s,r}^\calD $. 
The following example illustrates an application of the approach to two pairs of trees.

\begin{example}\label{ex:appendix-sat}
    Given the two pairs of trees $ \calD \df \{ (t_1, t_1^\star), (t_2, t_2^\star) \} $, we want to find a single transformation $ \rho_1 $ such that $ t_1 \rightsquigarrow_{\rho_1} t_1^\star $ and $ t_2 \rightsquigarrow_{\rho_1} t_2^\star  $, where
    \begin{align*}
        \left(%
            \toplabelled{$ t_1 $}{%
                \ltree{%
                    \Tree[.\ensuremath{\land} 
                            [.\ensuremath{\to} \ensuremath{A} \ensuremath{B} ]
                            [.\ensuremath{C} ]
                        ]
                }
            }, 
            \toplabelled{$ t_1^\star $}{%
                \ltree{%
                    \Tree[.\ensuremath{\land} 
                            [.\ensuremath{\to} \ensuremath{B} \ensuremath{A} ]
                            [.\ensuremath{C} ]
                        ]
                }
            }
        \right) \quad \text{ and } \quad
        \left(%
            \toplabelled{$ t_2 $}{%
                \ltree{%
                    \Tree[.\ensuremath{\to} 
                                [.\ensuremath{\land} \ensuremath{A} \ensuremath{B} ]
                                [.\ensuremath{\lor} \ensuremath{C} \ensuremath{D} ]
                            ]
                }
            },
            \toplabelled{$ t_2^\star $}{%
                \ltree{%
                    \Tree[.\ensuremath{\to} 
                            [.\ensuremath{\lor} \ensuremath{C} \ensuremath{D} ]
                            [.\ensuremath{\land} \ensuremath{A} \ensuremath{B} ]
                        ]
                }
            }
        \right).
    \end{align*}
    
    After constructing the formula $ \Phi^\calD_{1,1} $ and feeding it to a SAT solver, we will receive a model which might contain the following, positively assigned variables: $ \body^1_{\varepsilon, \to}$, $ \body^1_{0, Y_1}$, $ \body^1_{1, Y_2}$, $ \head^1_{\varepsilon, \to}$, $ \head^1_{0, Y_2}$, $ \head^1_{1, Y_1}$, $ \map^1_{0, 1}$, and $ \map^1_{\varepsilon, 2} $. The $ \body $ and $ \head $ variables induce the transformation
    \begin{align*}
        \toplabelled{$ \rho_1 $}{%
            \trafo{%
                \ltree{%
                    \Tree[.\ensuremath{\to} \ensuremath{Y_1} \ensuremath{Y_2} ]
                }
            }{%
                \ltree{%
                    \Tree[.\ensuremath{\to} \ensuremath{Y_2} \ensuremath{Y_1} ]
                }
            }
        }
    \end{align*}
     and the two $ \map $ variables indicate the nodes to which $ \rho_1 $ can be applied successfully: the left child of the root of $ t_1 $, i.e. node $ 0 $, and the root of $ t_2 $, i.e. node $ \varepsilon $. Observe that applying $ \rho_1 $ to the subtrees induced by these nodes does indeed result in $ t_1^\star $ and $ t_2^\star $, respectively. \qed
\end{example}

\subsubsection{Encoding Multi-step Instances}
\label{section:appendix-multi-step}

We now lift the encoding for single-step instances of the learning tree pattern transformations problem to multi-step instances. In single-step instances, i.e. when $s=1$, the only trees that ever occur are the source and target trees of the input examples and the encoding therefore had access to these trees. In multi-step instances, i.e. when $s > 1 $, we are searching for transformations such that each pair can be solved by a sequence of transformations. The main difference is that now ``intermediate trees'' need to be constructed. As we do not know what intermediate trees look like in advance, they need to be encoded by propositional variables as well. Therefore, in addition to the variables that we used before, we now also use variables $ \operatorname{int}_{i, k, v, \lambda} $ indicating that node $ v $ of the intermediate tree for pair $ (t_i, t_i^\star) $ after $ k $ steps has label $ \lambda $. This is demonstrated in Example \ref{example:appendix-encoding-with-intermediate-trees}.

\begin{example}\label{example:appendix-encoding-with-intermediate-trees}
    Suppose we are given a positive instance of \algorithmicProblem{LearningTreeTransformations}  with $s = 2$, $r = 1$ and suppose one of the examples is the pair 

    \begin{align*}
        \left(%
            \toplabelled{$ t_i$}{%
                \ltree{%
                    \Tree[.\ensuremath{a} \ensuremath{b} [.\ensuremath{a} \ensuremath{c} \ensuremath{b} ] ]%
                }%
            }, 
            \toplabelled{$ t_i^\star $}{%
                \ltree{%
                    \Tree[.\ensuremath{a} [.\ensuremath{a} \ensuremath{b} \ensuremath{c} ] \ensuremath{b} ]%
                }%
            }%
        \right)\text{.}
    \end{align*}
    Further suppose that the transformation $\rho_1\colon$
        \ttrafo{%
            \ltree{%
                \Tree[.\ensuremath{x} \ensuremath{Y_1} \ensuremath{Y_2} ]
            }
        }{%
            \ltree{%
                \Tree[.\ensuremath{x} \ensuremath{Y_2} \ensuremath{Y_1} ]
            }} is a solution. 
    
\noindent
Then there are two ways to transform $ t_i $ into $ t_i^\star $ using $ \rho_1 $:

    \begin{center}
        \setlength{\tabcolsep}{3pt}
        \begin{tabular}{c c c c c}
            \adjustbox{valign=m}{%
                \tree{%
                    \Tree[.\ensuremath{a} \ensuremath{b} [.\ensuremath{a} \ensuremath{c} \ensuremath{b} ] ]%
                }
            } &
            \adjustbox{valign=m}{\Large$ \rightsquigarrow_{\!\rho_1} $} &
            \adjustbox{valign=m}{%
                \toplabelled{$ I_1 $}{%
                    \tree{%
                        \Tree[.\ensuremath{a} [.\ensuremath{a} \ensuremath{c} \ensuremath{b} ] \ensuremath{b} ]%
                    }
                }
            } &
            \adjustbox{valign=m}{\Large$ \rightsquigarrow_{\!\rho_1} $} &
            \adjustbox{valign=m}{%
                \tree{%
                    \Tree[.\ensuremath{a} [.\ensuremath{a} \ensuremath{b} \ensuremath{c} ] \ensuremath{b} ]%
                }
            }
        \end{tabular}%
    \end{center}
        and
    \begin{center}
        \setlength{\tabcolsep}{3pt}
        \begin{tabular}{c c c c c}
            \adjustbox{valign=m}{%
                \tree{%
                    \Tree[.\ensuremath{a} \ensuremath{b} [.\ensuremath{a} \ensuremath{c} \ensuremath{b} ] ]%
                }
            } &
            \adjustbox{valign=m}{\Large$ \rightsquigarrow_{\!\rho_1} $} &
            \adjustbox{valign=m}{%
                \toplabelled{$ I_2 $}{%
                    \tree{%
                    \Tree[.\ensuremath{a} \ensuremath{b} [.\ensuremath{a} \ensuremath{b} \ensuremath{c} ] ]%
                    }
                }
            } &
            \adjustbox{valign=m}{\Large$ \rightsquigarrow_{\!\rho_1} $} &
            \adjustbox{valign=m}{%
                \tree{%
                    \Tree[.\ensuremath{a} [.\ensuremath{a} \ensuremath{b} \ensuremath{c} ] \ensuremath{b} ]%
                }
            }
        \end{tabular}.
    \end{center}
    Here, $ I_1 $ and $ I_2 $ are the (only possible) intermediate trees for $ (t_i, t_i^\star) $ after one step. The corresponding propositional variables are 
    \begin{itemize}
        \item $ \operatorname{int}_{i, 1, \varepsilon, a}$, $ \operatorname{int}_{i, 1, 0, a}$, $ \operatorname{int}_{i, 1, 1, b}$, $ \operatorname{int}_{i, 1, 00, c} $, and $ \operatorname{int}_{i, 1, 01, b} $ for $ I_1 $; and
        \item $ \operatorname{int}_{i, 1, \varepsilon, a}$, $ \operatorname{int}_{i, 1, 0, b}$, $ \operatorname{int}_{i, 1, 1, a}$, $ \operatorname{int}_{i, 1, 10, b} $, and $ \operatorname{int}_{i, 1, 11, c} $ for $ I_2 $. \qed
    \end{itemize}%
\end{example}

The formulas from the previous section need to be adapted to these new variables. The variables $ \map_{v, i}^j $ are replaced by variables $ \map_{v, i}^{j,k} $, where $ k \le s $ refers to a step. The intended meaning for this variable is that the $ j $-th transformation is applied to node $ v $ of the intermediate tree of pair $ (t_i, t_i^\star) $ after step $ k-1 $, where intermediate tree $ 0 $ is just $ t_i $. In the other formulas, instead of formulating constraints based on the source and target trees of the example pairs, they are now formulated over the intermediate trees.
For instance, in the formula $ \psi_{\mathrm{labels}}^{i,j,v,w} $, instead of requiring that the label of $ t_i^\star $ must be the same as the one in the head at the corresponding position, this requirement would be stated using the new variables for intermediate trees. The relevant subformula would look as follows:
\begin{align*}
    \map_{v, i}^{j,k} \to \left(\bigwedge_{\descpos{w}{v}} \bigwedge_{\lambda \in \Sigma} \left(\head_{w, \lambda}^j \to \operatorname{int}_{i, k, vw, \lambda}\right)\right)\text{.}
\end{align*}

This states that for transformation $ \rho_j $ to be applicable to pair $ i $ at node $ v $ in step $ k $, a label from $ \Sigma $ in the head of transformation $ \rho_j $ must be identical to the label at the corresponding position in the intermediate tree of pair $ i $ constructed in the previous $ k-1 $ steps.
All the other formulas need to be adapted similarly. Note that the adapted formulas can still be used for the case where $ s = 1 $, because there is a one-to-one correspondence between $ \map_{v, i}^j $ and $ \map_{v, i}^{j,1} $.

\label{section:Experiments}
\subsection{Implementation and Application}
\label{section:appendix-application}

We implemented this encoding and applied it to datasets from CS education%
\footnote{The encoding has been implemented in Java and uses a Java wrapper for Z3 \cite{MouraB08}. All tests have been conducted on a Debian-based machine with an Intel i7-11850H CPU.}. Our implementation supports single-step and multi-step instances, an incremental mode, as well as some practical extensions such as providing a ratio of how many examples need to be explained.

Before applying our encoding to real data, we tested that it works as expected by applying it to the \algorithmicProblem{LearningTreeTransformations} instances from Examples \ref{ex:s-one-t-in-hardness} and \ref{ex:s-three-t-two}. %
Encodings of both example sets were solved and the returned models were as expected.

We then prototypically tested our framework on datasets from CS education. Many typical mistakes by students in propositional modelling were identified by Schmellenkamp et al. \cite{SchmellenkampLZ23} with a custom approach. Their dataset consists of pairs of formulas $(\varphi, \varphi^\star)$, where the second one is a valid solution while the first is an erroneous modelling attempt. They also provided a clustering of the dataset by typical mistakes that have been identified.

To test the applicability of our framework in this context, we performed two tests.
First, we verified that our tree pattern transformations are suitable to express typical mistakes in this context. 
For example, a typical mistake is to model ``only if'' statements as if they were ``if'' statements.
In syntax trees of formulas this corresponds to swapping the two children of an implication node. The dataset \cite{SchmellenkampLZ23} contains 83 pairs of formulas with this mistake. For instance, it contains pairs like $ (E \to (\lnot A \land \lnot C), (\lnot A \land \lnot C) \to E) $ and $ (B \to D, D \to B) $. %
For the \algorithmicProblem{LearningTreeTransformations} instance with these 83 pairs of (formula) trees and with $ s \df 1 $ and $ r \df 1 $, 
the model provided by the \SAT solver encodes the expected transformation \adjustbox{valign=t}{\ttrafo{%
        \ltree{%
            \Tree[.\ensuremath{\to} \ensuremath{Y_2} \ensuremath{Y_1} ]%
        }%
    }{%
        \ltree{%
            \Tree[.\ensuremath{\to} \ensuremath{Y_1} \ensuremath{Y_2} ]%
        }%
}}.

Second, we tested whether our framework can be used to automatically identify candidates for common modelling mistakes. To this end, ideally, datasets with pairs of formulas where similar mistakes are expected to happen are used in order to find few tree pattern transformations that explain many of the formula pairs. Such datasets naturally occur when clustering by linguistic operators that occur in the natural language statement that is part of the assignment for students and are provided in the dataset by Schmellenkamp et al. As an example, we consider the operator ``either-or'' in statements like \textit{I'm travelling to either Italy or Spain this year.} After a preprocessing step of unifying the propositional variables used in all the erroneous formula pairs for this linguistic operator, 19 different pairs remained. The encoding of the \algorithmicProblem{LearningTreeTransformations} instance containing these 19 pairs with $ s \df 1 $
was solved 
by using an incremental approach for $r = 1, 2, \ldots$.
The model provided by the \SAT solver encodes the following four transformations:
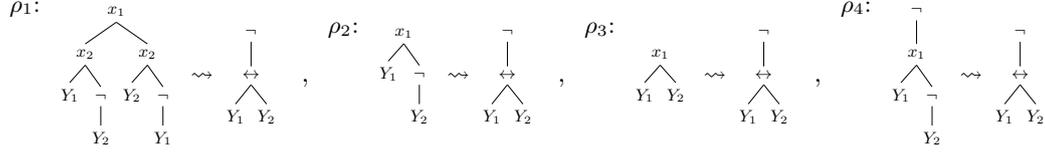
\begin{center}
    \toplabelled{$ \rho_1 $}{%
        \scalebox{0.7}{%
        \trafo{%
            \ltree{%
                \Tree[.\ensuremath{x_1} 
                [.\ensuremath{x_2} 
                \ensuremath{Y_1} [.\ensuremath{\lnot} \ensuremath{Y_2} ]
                ] 
                [.\ensuremath{x_2} 
                \ensuremath{Y_2} [.\ensuremath{\lnot} \ensuremath{Y_1} ] 
                ] 
                ] 
            }
        }{%
            \ltree{%
                \Tree[.\ensuremath{\lnot} [.\ensuremath{\leftrightarrow} \ensuremath{Y_1} \ensuremath{Y_2} ] ]
            }
        }}
    },
    \toplabelled{$ \rho_2 $}{%
        \scalebox{0.7}{%
        \trafo{%
            \ltree{%
                \Tree[.\ensuremath{x_1} \ensuremath{Y_1} [.\ensuremath{\lnot} \ensuremath{Y_2} ] ]
            }
        }{%
            \ltree{%
                \Tree[.\ensuremath{\lnot} [.\ensuremath{\leftrightarrow} \ensuremath{Y_1} \ensuremath{Y_2} ] ]
            }
        }}
    },
    \toplabelled{$ \rho_3 $}{%
        \scalebox{0.7}{%
        \trafo{%
            \ltree{%
                \Tree[.\ensuremath{x_1} \ensuremath{Y_1} \ensuremath{Y_2} ]
            }
        }{%
            \ltree{%
                \Tree[.\ensuremath{\lnot} [.\ensuremath{\leftrightarrow} \ensuremath{Y_1} \ensuremath{Y_2} ] ]
            }
        }}
    },
    \toplabelled{$ \rho_4 $}{%
        \scalebox{0.7}{%
        \trafo{%
            \ltree{%
                \Tree[.\ensuremath{\lnot} [.\ensuremath{x_1} \ensuremath{Y_1} [.\ensuremath{\lnot} \ensuremath{Y_2} ] ] ]
            }
        }{%
            \ltree{%
                \Tree[.\ensuremath{\lnot} [.\ensuremath{\leftrightarrow} \ensuremath{Y_1} \ensuremath{Y_2} ] ]
            }
        }}
    }
\end{center}

These transformations correspond to mistakes one would expect. For instance, transformations $ \rho_2 $ and $ \rho_3 $ explain pairs for which the erroneous formula uses a faulty operator like a conjunction instead of a bi-implication as in $(A \land \lnot B, \neg(A \leftrightarrow B)) $ or $ (A \land B,  \neg(A \leftrightarrow B))$.

\section{Additional Material for \cref{section:interval:algorithmic}: Hardness of Testing Interval Variables}\label{section:appendix-intervalvars}
\intervalexplanationproblemhard*

At the core is the hardness of the following algorithmic problem:%

\algorithmicProblemDescription{PartitionedRewriting}
        {
            Strings $u, v \in \Sigma^*$ and a permutation $\sigma: [k] \rightarrow [k]$
        }
        {
            Is there a partitioning $u = u_1 \cdots u_k$ such that $v = u_{\sigma(1)} \cdots u_{\sigma(k)}$?
        }

\newcommand{\Left}{\text{LEFT}}
\newcommand{\Right}{\text{RIGHT}}

We first provide an outline why \algorithmicProblem{PartitionedRewriting} is $\NP$-hard, when the factors $u_i$ may only be non-empty strings (denoted \emph{non-erasing case}). Afterwards we proceed with the technically harder general case (\emph{erasing case}).  

\paragraph*{Non-erasing case}
We denote the partitions $u_i$ in the following \emph{intervals}. They must be non-empty for this case.

\bigskip

\noindent
We show the $\NP$-hardness of \algorithmicProblem{PartitionedRewriting} via a polynomial many-one reduction from the \emph{perfect code problem} for 3-regular (undirected) graphs, an $\NP$-complete problem \cite{KratochvilK1988}. Our reduction is inspired by \cite[Thm 4]{DayFMNS2018}. A 3-regular graph is a graph $G = (V,E)$ where each node has degree 3. The \emph{neighborhood} $N_G(x)$ of a node $x \in V$ is defined as $N_G(x) = \{y \mid \{x,y\} \in E\} \cup \{x\}$, that is, all nodes adjacent to $x$ and $x$ itself.
A \emph{perfect code} is a set of nodes $C \subseteq V$ such that for each $x \in V$ it holds  $|N_G(x) \cap C| = 1$, that is, each neighborhood $N_G(x)$ contains exactly one node from $C$.
The \emph{perfect code problem} asks whether a given graph admits a perfect code.
Note that for a 3-regular graph $G= (V,E)$, for each node $x\in V$ it holds $|N_G(x)| = 4$, and if $G$ admits a perfect code $C$, then $n := |V|$ is a multiple of $4$ and $|C| = n/4$.

\medskip

\noindent
Be now $G = (V, E)$ a 3-regular graph, $|V|=n$.
We define ($u,v,\sigma)$ an instance of \algorithmicProblem{PartitionedRewriting} in the following.

\begin{align*}
u &= \#^{2n}b^nc(a^5*)^nd(\#a^8)^{2n-n/4}(a^4\#)^{n+n/4} \\
v &= cd*^n(\#a^8\#a^8\#a^4\#a^4\#b)^n
\end{align*}

We indicate the permutation $\sigma$ by permuting a sequence of intervals, thereby giving names that facilitate the presentation of the proof. For an integer $n$, denote by $[n]$ the set $\{1,\dots,n\}$ and be $V = \{t_1, \dots t_n\}$. 
To get a convenient access to the four nodes in the neighborhood $N_G(t_i)$ of a given $t_i \in V$, fix an arbitrary order of the members of $N_G(t_i)$ and be $p_r: [n] \rightarrow [n]$, for $r \in [4]$, the mapping where $p_r(i) = j$ indicates that the $r^{\text{th}}$ neighbor of $t_i$ is $t_j$. That is, for each $t_i \in V$, it holds $N_G(t_i) = \{t_{p_1(i)}, t_{p_2(i)}, t_{p_3(i)}, t_{p_4(i)}\}$.
Define the $n^2$ many intervals $u_{i,j}$ for $i,j \in [n]$. Although, we will only use those $u_{i,j}$ where there is an edge between $t_i$ and $t_j$, or vice versa. So, for each $u_{i,j}$ that we use, we will also use $u_{j,i}$. Accordingly, each $u_{i,j}$ we use corresponds to a "1"-entry in the (symmetric) incidence matrix of $G$.

Note that the incidence matrix of a 3-regular graph contains in each row exactly $4$ "1"-entries, and also in each column exactly $4$ "1"-entries.

\smallskip

\noindent
A property of a perfect code (for 3-regular graphs) which is key to the proof idea is the following.

\smallskip

\noindent
\textbf{Important Property.}\\
$G$ admits a perfect code $C$ ($|C| = n/4$), if and only if we can select in each column $j$ of the incidence matrix $A$ exactly one "1"-entry, (that is, a certain row $i$ such that $A[i,j] = 1$) in such a way that in exactly $n/4$ many rows we have selected exactly all $4$ "1"-entries, and in all other ($3/4n$ many) rows none at all.

\smallskip

\noindent
Further, we define the following shortcuts.

\begin{align*}
\alpha_i  &= u_{p_1(i),i}\; u_{p_2(i),i}\; u_{p_3(i),i}\; u_{p_4(i),i} \\
\alpha'_i &= u_{i,p_1(i)}\; u_{i,p_2(i)}\; u_{i,p_3(i)}\; u_{i,p_4(i)} \\
\end{align*}

\noindent
We can now rephrase the important property in terms of the $\alpha_i$'s and $\alpha'_i$'s: 

\medskip

\noindent
\textbf{Important Property rephrased.}\\
$G$ admits a perfect code $C$ ($|C| = n/4$), if and only if we can select in each $\alpha_i$ exactly one interval in such a way that in $n/4$ many $\alpha'_i$'s all $4$ intervals are selected, and in all other $3/4n$ many $\alpha'_i$'s no interval is selected.

\smallskip

\noindent
We now define the permutation $\sigma$ as mapping $X$ to $Y$, where
\begin{align*}
X = & \;\mu_1 \cdots \mu_n\nu_1\cdots \nu_n\zeta_1 \cdots \zeta_n\,\eta\,\alpha_1\delta_1 \cdots \alpha_n\delta_n\,\theta\,\beta_1\cdots \beta_n\gamma_1\cdots \gamma_n \\
Y = & \;\eta\theta\delta_1\cdots \delta_n\;\beta_1\mu_1\alpha'_1\nu_1\gamma_1\zeta_1\; \cdots\; \beta_n\mu_n\alpha'_n\nu_n\gamma_n\zeta_n
\end{align*}

\noindent
Note that we use $10 n + 2$ intervals.

\subparagraph*{Correctness}
"$\Rightarrow$": Be $C \subseteq V$ a perfect code. Note that $|C| = n/4$.
Choose all $\mu_i$, $\nu_i$, $\delta_i$, $\zeta_i$, $\eta$, $\theta$ of size 1. Note that this assures $\mu_i = \nu_i = \#$, $\zeta_i = b$, and $\eta = c$. Choose $u_{i,j}$ of size 2, if $t_i \in C$, and of size 1 otherwise. With the properties of a perfect code (in each neighborhood exactly one node belongs to $C$), and the previous choices, this implies that each $\alpha_i$ captures exactly $a^5$ (in $u$), and  each $\delta_i$ captures exactly $*$. We can now also deduce that $\theta$ captures $d$, and not more (because $\theta$ is followed by $\#$ in $u$, and by $*$ in $v$).  Further properties of a perfect code (through the neighborhood of each $t_i \in C$ we reach each node in $V$ exactly once) imply that $\alpha'_i = a^8$ if $t_i \in C$, and $\alpha'_i = a^4$ otherwise. Further choose $\beta_i$ of size 9 and $\gamma_i$ of size 10 if $t_i \in C$, and otherwise choose $\beta_i$ of size 18 and $\gamma_i$ of size 5. This allows to verify that, for all $i$, $\beta_i\mu_i\alpha'_i\nu_i\gamma_i\zeta_i$ expresses exactly $\#a^8\#a^8\#a^4\#a^4\#b$. Since $\eta\theta\delta_1 \cdots \delta_n$ expresses $cd*^n$ we have shown that

\begin{align*}
v  & = cd*^n(\#a^8\#a^8\#a^4\#a^4\#b)^n\\
   & = \eta\theta\delta_1\cdots \delta_n\;\beta_1\mu_1\alpha'_1\nu_1\gamma_1\zeta_1\; \cdots\; \beta_n\mu_n\alpha'_n\nu_n\gamma_n\zeta_n\\
   & = \sigma(\mu_1 \cdots \mu_n\nu_1\cdots \nu_n\zeta_1 \cdots \zeta_n\,\eta\,\alpha_1\delta_1 \cdots \alpha_n\delta_n\,\theta\,\beta_1\cdots \beta_n\gamma_1\cdots \gamma_n)
\end{align*}

\bigskip

\noindent
"$\Leftarrow$": Be $(u,v,\sigma)$ a yes instance. We will construct a perfect code for $G$. Since $\sigma(X) = Y = v$, and $\eta$, $\theta$ and all $\delta_i$ are of size at least 1 (non-erasing), and $v$ starts with $cd$, we deduce that $\eta$ must capture in $u$ the only $c$. Also, $\eta$ must be of size not more than 1, otherwise it would bring in $a$'s from $u$ between $c$ and $d$. Analogous reasoning allows to conclude that $\theta$ captures exactly $d$, and not more.

Since now we know that $\eta\theta = cd$, we deduce that the $\delta_i$ must be responsible for the $*$'s in $v$, no other intervals could be. Further, since in $u$ after each * never follows immediately another *, we conclude that each $\delta_i$ is exactly of size 1, and $\delta_i = *$. This allows further to conclude, since in $u$ each $\alpha_i$ is enclosed by $\eta$ or $\delta_i$'s (thus a $c$ or $*$'s), that each $\alpha_i$ is of size 5 and captures exactly $a^5$. In other words, $\eta\alpha_1\delta_1 \cdots \alpha_n\delta_n\theta$ captures exactly $c(a^5*)^nd$.

Consequently, $\beta_1\cdots \beta_n\gamma_1\cdots \gamma_n$ captures $(\#a^8)^{2n-n/4}(a^4\#)^{n+n/4}$, \\
and $\mu_1 \cdots \mu_n\nu_1\cdots \nu_n\zeta_1 \cdots \zeta_n$ captures $\#^{2n}b^n$. From the latter observation we conclude that $\mu_i = \nu_i = \#$ and $\zeta_i = b$, for each $i$.

\bigskip

\noindent
Since $Y = v$, we deduce:\\
For each $i$, $\beta_i\mu_i\alpha'_i\nu_i\gamma_i\zeta_i = \beta_i\#\alpha'_i\#\gamma_ib$ expresses $\#a^8\#a^8\#a^4\#a^4\#b$.\\
Since each $\alpha_i$ contains only a's, we deduce each $\alpha'_i$ contains only a's. Therefore, we deduce that each $\alpha'_i$ expresses either $a^4$ or $a^8$.\\
We deduce that each $\beta_i$ captures either $\#a^8$ (if $\alpha'_i = a^8$), or 
$\#a^8\#a^8$ (if $\alpha'_i = a^4$),  and that each $\gamma_i$ captures either $a^4\#a^4\#$ (if $\alpha'_i = a^8$), or $a^4\#$ (if $\alpha'_i = a^4$).
\\\\
This, together with $\beta_1\cdots \beta_n\gamma_1\cdots \gamma_n$ capturing $(\#a^8)^{2n-n/4}(a^4\#)^{n+n/4}$, lets us deduce that in exactly $n/4$ cases
we have $\beta_i = \#a^8$, $\alpha'_i = a^8$, and $\gamma_i = a^4\#a^4\#$, and in exactly $3/4n$ cases we have $\beta_i = \#a^8\#a^8$, $\alpha'_i = a^4$, and $\gamma_i = a^4\#$. This encodes a perfect code by choosing exactly those $t_i$ where $\alpha'_i$ expresses $a^8$ (confer \textbf{Important property rephrased}).

\paragraph*{Erasing case} In this case the intervals $u_i$ can be empty. We use $\epsilon$ to denote an empty interval.

\bigskip

The main difficulty arising now is that certain intervals that we intend to capture a \emph{certain part of $u$}, could in principle be empty, and (any) another interval could capture the \emph{certain part} instead. This would hinder any straight forward reasoning as in the non-erasing case.
In order to force certain intervals to be non-empty, and capture what we intend them to capture, we introduce some additional tricks and reasoning.

\bigskip

\noindent
Be $G = (V, E)$ a 3-regular graph, $|V|=n$.
We define $u$, $v$, and $\sigma$ in the following.

\begin{align*}
u &= (a*)^n(pq)^{4n}x(st)^{2n}y(\#a^4)^{3/4n}z\#^{n/4} \\
v &= *^nxyz\,p^{4n}(\#a^4q)^n\,s^{2n}q^{3n}t^{2n}
\end{align*}

\noindent
Define the shortcuts as above:
\begin{align*}
\alpha_i  &= u_{p_1(i),i}\; u_{p_2(i),i}\; u_{p_3(i),i}\; u_{p_4(i),i} \\
\alpha'_i &= u_{i,p_1(i)}\; u_{i,p_2(i)}\; u_{i,p_3(i)}\; u_{i,p_4(i)} \\
\end{align*}

\noindent
Permutation $\sigma$ maps $X$ to $Y$, where
\begin{align*}
X = & \;\alpha_1\delta_1 \cdots \alpha_n\delta_n\mu_1\nu_1 \cdots\mu_{4n}\nu_{4n}\,\eta\,\theta_1\pi_1 \cdots \theta_{2n}\pi_{2n}\,\omega\,\beta_1\cdots \beta_n\,\zeta\,\gamma_1\cdots \gamma_n \\
Y = & \;\delta_1\cdots \delta_n\;\eta\omega\zeta\,\mu_1\cdots\mu_{4n}\beta_1\gamma_1\alpha'_1\nu_1\; \cdots\; \beta_n\gamma_n\alpha'_n\nu_n\theta_1\cdots\theta_{2n}\nu_{n+1}\cdots\nu_{4n}\pi_1 \cdots \pi_{2n}
\end{align*}

\noindent
Observe: $\sigma$ has $19n+3$ intervals.

\subparagraph*{Correctness}

\noindent
"$\Rightarrow$": Be $C \subseteq V$ a perfect code. Note that $|C| = n/4$.
Choose all $\mu_i$, $\nu_i$, $\delta_i$, $\theta_i$, $\pi_i$, and $\eta$, $\omega$, $\zeta$ of size 1. Note that this assures (because of $Y$) that
$\delta_i = *$, $\eta = x$, $\omega = y$, $\zeta = z$, and $\mu_i = p$.
Choose $u_{i,j}$ of size 1, if $t_i \in C$, and of size 0 otherwise. With the properties of a perfect code (in each neighborhood exactly one node belongs to $C$), and the knowledge $\delta_i = *$, this implies that each $\alpha_i$ captures exactly $a$ (because of $X$). Now we can further deduce (because of $X$), that $\nu_i = q$, $\theta_i = s$, $\pi_i = t$.

Further properties of a perfect code (through the neighborhood of each $t_i \in C$ we reach each node in $V$ exactly once) imply that $\alpha'_i = a^4$ if $t_i \in C$, and $\alpha'_i = \epsilon$ otherwise.
Further, choose $\beta_i$ of size 0 and $\gamma_i$ of size 1 if $t_i \in C$, and otherwise choose $\beta_i$ of size 5 and $\gamma_i$ of size 0.
This allows to verify that, for all $i$, $\beta_i\gamma_i\alpha'_i$ expresses exactly $\#a^4$:
either
$\beta_i = \epsilon$, $\gamma_i = \#$, $\alpha'_i = a^4$ if $t_i \in C$ ($n/4$ many cases), or
$\beta_i = \#a^4$, $\gamma_i = \epsilon$, $\alpha'_i = \epsilon$ if $t_i \not \in C$ ($3/4n$ many cases).

In summary, the so chosen interval sizes allow to conclude that $\sigma(X) = Y = v$.

\medskip

\noindent
"$\Leftarrow$": This direction is more involving. We will eventually construct a perfect code, but first argue that, if $\sigma(u_1, \dots, u_{19n+3}) = v$, then at least $15n+3$ intervals must be non-empty. Hence, at most $4n$ intervals can be empty.

\medskip

\noindent
In the following we go systematically through all types of symbols in $u$ and argue by how many non-empty intervals they must be captured. Most of the times it will be single symbols who need to be captured by a single interval of size 1, due to the \emph{different surroundings argument}: for a symbol $b$, define $\Left_u(b)$ as the set of symbols that directly precede $b$ in $u$, and define $\Right_u(b)$ as the set of symbols that directly follow $b$ in $u$. Now it is immediate to conclude that, if $\Left_u(b)$ and $\Left_v(b)$ are disjoint, and so are $\Right_u(b)$ and $\Right_v(b)$, then each occurrence of $b$ must be captured by its own interval of size 1.

\medskip

\noindent
In the following enumeration, the first number indicates the number of non-empty intervals observed, followed by the argument behind it.

\begin{enumerate}
\item $1$: $\Left_u(z) = a$, $\Left_v(z) = y$, $\Right_u(z) = \#$, $\Right_v(z) = p$. Different surroundings argument applies: $z$ must be captured by an interval of size $1$.

\item $1$: $\Left_u(x) = q$, $\Left_v(x) = *$, $\Right_u(x) = s$, $\Right_v(x) = y$. Different surroundings argument applies: $x$ must be captured by an interval of size $1$.

\item $1$: $\Left_u(y) = t$, $\Left_v(y) = x$, $\Right_u(y) = \#$, $\Right_v(y) = z$. Different surroundings argument applies: $y$ must be captured by an interval of size $1$.

\item $n$: $\Left_u(*) = a$, $\Left_v(*) = \{*,\epsilon\}$, $\Right_u(*) = \{a,p\}$, $\Right_v(*) = \{*,x\}$. Different surroundings argument applies: each $*$ must be captured by an interval of size $1$.

\item $n$: Analogously to the different surroundings argument the first $n$ $a$'s in $v$ must be captured by $n$ intervals of size 1 each. Note that this reasoning can not be applied to the other $a$'s that occur further to the right in $u$.

\item $8n$: By the different surroundings argument the $p$'s and $q$'s in $u$ must be captured by $8n$ intervals of size $1$ each.

\item $4n$: By the different surroundings argument the $s$'s and $t$'s in $u$ must be captured by $4n$ intervals of size $1$ each.

\item $n/4$: Each of the $\#^{n/4}$ at the end of $u$ must be captured by an interval of size $1$, via the different surroundings argument. Note that this reasoning can not be applied to the other $\#$'s further to the left in $u$.

\item $3/4n$: Each block of $\#a^4$ in $u$ must be captured. It could be done by multiple intervals in principle, but it must be at least one interval per block (one interval can not capture multiple blocks, because in $u$ each block is followed by $\#$ or $z$ and in $v$ by a $q$ instead). Hence, we need at least $3/4n$ non-empty intervals (each of size at most $5$).
\end{enumerate}

\noindent
In summary, we have at least $1 + 1 + 1 + n + n + 8n + 4n + n/4 + 3/4n $ $= 15n + 3$ non-empty intervals. Since there are $19n + 3$ intervals in total, at most $4n$ intervals can be empty.

\bigskip

\noindent
We will now show that, if $\sigma(u_1, \dots, u_{19n+3}) = v$, then $\zeta = z$. Recall that we have already established that $z$ must be captured by an interval of size 1.
Assume for contradiction that $\zeta$ does not capture $z$. Then the $z$ in $u$ must be captured by another interval (of size 1). We now go systematically through all alternatives and show that each implies that more than $4n$ intervals must be empty, a contradiction.

\begin{enumerate}
\item $\beta_1$: if $\beta_1 = z$, then in $Y$ all intervals occurring before $\beta_1$ must express $*^nxy$, which, as argued above, must be done by exactly $n+2$ intervals of size 1. In $Y$ there are $5n + 3$ intervals occurring before $\beta_1$ ($n$ $\delta_i$'s, $\eta$, $\omega$, $\zeta$, $4n$ $\mu_i$'s). Thus, of these $5n + 3$ intervals, $n+2$ must be non-empty, but the rest,  $4n + 1$ in number, must empty. A Contradiction.

\item By the same argument no interval occurring in $Y$ to the right of $\beta_1$ can capture $z$ as it would only imply more empty intervals. This excludes any $\beta_i, \gamma_i, \nu_i, \theta_i, \pi_i$, and also any $\alpha_i$ or $\alpha'_i$ (which is all the $u_{i,j}$).

\item $\omega$: if $\omega = z$, then $\delta_1 \cdots \delta_n \eta$, that is, $n+1$ intervals, must express $*^nxy$. But according to the previous reasoning, each symbol in $*^nxy$, that is, $n + 2$ many, must be expressed by an individual interval of size 1. A contradiction.

\item $\eta$: if $\eta = z$, then in $X$ all intervals occurring after $\eta$ must capture $\#^{n/4}$, that is, $6n+2$ in number. Since $\#^{n/4}$ must be captured by $n/4$ intervals of size 1, of these $6n+2$ intervals $n/4$ are non-empty, but the rest, $6n+2-n/4 = 5n + 3/4n + 2$ many, must be empty. A contradiction.

\item By the same argument no interval occurring in $X$ to the left of $\eta$ can capture $z$ as it would only imply more empty intervals. This excludes any $\mu_i, \delta_i$ (and again all $\nu_i$, $\alpha_i$).

\end{enumerate}

\noindent
Since all alternatives lead to a contradiction, we have proved that $\zeta = z$, provided that $\sigma(u_1, \dots, u_{19n+1}) = v$.

\medskip

\noindent
From this we can now derive the following:
\begin{enumerate}
\item Since $\zeta = z$, we can (because of $Y$) derive that each $\delta_i = *$, and that $\eta = x$ and $\omega = y$.
\item From this, because of $X$, we can derive that each $\alpha_i$ captures exactly $a$. Thus, $\alpha_1\delta_1 \cdots \alpha_n\delta_n$ captures $(a*)^n$.
And we can derive that in each $\alpha_i$ there is one $u_{i,j} = a$ and the other three intervals are empty. We have thus necessarily $3n$ empty intervals.
Since in total there can be at most $4n$ empty intervals, we now know that there can only be $n$ additional empty intervals. We also know that each $\alpha'_i$ can only contain $a$'s, between $0$ and $4$.
\item Since $\alpha_1\delta_1 \cdots \alpha_n\delta_n = (a*)^n$, $\eta = x$, $\omega = y$, and $\zeta = z$, we can derive the following captures in $X$:
  \begin{enumerate}
	\item $\mu_1\nu_1 \cdots \mu_{4n}\nu_{4n} = (pq)^{4n}$ and hence $\mu_i = p$ and $\nu_i = q$, for $i \in [4n]$.
	\item $\theta_1\pi_1 \cdots \theta_{2n}\pi_{2n} = (st)^{2n}$ and hence $\theta_i = s$ and $\pi_i = t$, for $i \in [2n]$.
	\item $\beta_1 \cdots \beta_n = (\#a^4)^{3/4n}$
	\item $\gamma_1 \cdots\gamma_n = \#^{n/4}$
  \end{enumerate}

\item From this we can finally derive that for each $i \in [n]$, the sequence $\beta_i\gamma_i\alpha'_i$ must express exactly $\#a^4$.

\item Since $\gamma_1 \cdots\gamma_n = \#^{n/4}$, and each $\#$ must be captured by an interval of size $1$, we can derive that of the $\gamma_i$ exactly $n/4$ capture one $\#$ each and the other $3/4n$ many are empty.

\item As argued above, any non-empty interval capturing a block $\#a^4$ can not be of size more than 5, and must start with $\#$. Since $\beta_1 \cdots \beta_n = (\#a^4)^{3/4n}$, we now know that each $\beta_i$ is either empty, or it starts with $\#$ and is of size at most 5. Could it be of any size less than 5? No, it could not, since otherwise either the next non-empty $\beta_{i+x}$ would not start with  $\#$, or in case of the last $\beta_n$ some $a$'s would not be captured, since the following $\zeta = z$. In summary, we now know that exactly $3/4n$ of the $\beta_i$'s capture $\#a^4$ each, and the other $n/4$ are empty.

\item Now we can derive that there are only two possibilities for $\beta_i\gamma_i\alpha'_i$ to express $\#a^4$: Either 1) $\beta_i = \epsilon$, $\gamma_i = \#$ and $\alpha'_i = a^4$ ($n/4$ cases), or 2) $\beta_i = \#a^4$, $\gamma_i = \epsilon$ and $\alpha'_i = \epsilon$ ($3/4n$ cases).

\item This encodes a perfect code by choosing exactly those $t_i$ where $\alpha'_i$ expresses $a^4$ (confer \textbf{Important property rephrased}).
\end{enumerate}
 
\section{Towards Algorithms for Learning Tree Pattern Transformations}\label{sec:towards-ptime}
There are multiple ways to deal with the hardness of the learning problem. One is to encode the problem in propositional logic and employ a modern SAT solver, see \cref{section:sat-solving} and \cref{section:SAT-solving}. 
Another option is to restrict the input parameters even further.
A natural restriction of \algorithmicProblem{LearningTreeTransformations} is to try and find a single transformation explaining all pairs in a single step, i.e. fixing $ s \df 1 $ and $ r \df 1 $. 

\begin{restatable}{proposition}{oneoneatroot}\label{thm:proposition_one-one-root}
    When fixing $ s \df 1 $ and $ r \df 1 $ as well as requiring that transformations must be applied at the root, {\algorithmicProblem{LearningTreeTransformations}} is in \PTIME{}, %
    and if there exists a transformation explaining all input pairs, one can be computed in polynomial time.
\end{restatable}

Before proving this proposition, note that it can be generalised: instead of just applying to the root, for each pair, any predetermined fixed position can be used. To accommodate for this, one could extend \algorithmicProblem{LearningTreeTransformations}' input by an $ n $-tuple of positions denoting for each pair the position to which the transformation needs to be applied. 
This can then be solved by first checking, for each pair, whether the assigned position is valid, 
by verifying that differences between source and target trees may only occur in the subtrees rooted at the provided position.
After that, this 
directly reduces to the case of applications only at the root node by constructing a new instance consisting only of the subtrees rooted at the assigned positions.

In practice, the (pairs of) trees considered are often small. This is helpful due to the following:

\begin{proposition}
    When fixing $ s \df 1 $ and $ r $ to any constant, {\algorithmicProblem{LearningTreeTransformations}} is 
    fixed parameter tractable for parameters
    \begin{romanenumerate}
        \item $ k_1 \df $ size of the largest pair of trees;
        \item $ k_2 \df $ size of the transformation to be found.
    \end{romanenumerate}
\end{proposition}

Note that (i) implies (ii) since the size of the transformation is bounded by the size of the largest pair of trees.

\begin{proofsketch}
    We only sketch (i) and first consider the case where $ r \df 1 $. 
    The idea is to enumerate all possible transformations of size at most $ k_1 $ and test whether any of them explains all pairs.
    Let $ T(m) $ be the number of structurally different ordered trees of $ m $ nodes, let $ (t, t^\star) $ be a pair of size $ k_1 $ and let $ D $ be the active domain of $ t $.
    Observe that $ |D| \le k_1 $ and in a tree pattern of size $ k $ over $ \Sigma = D $, the number of possible labels of a node is at most $ | D | + k $.

    The number of patterns matching the tree $ t $ is bounded by
    \begin{align*}
        \sum_{i=1}^{k_1} (| D | + i)^i \cdot T(i) \le k_1 \cdot (2k_1)^{k_1} \cdot T(k_1) \df B
    \end{align*}
    and the number of transformations that could explain $ (t, t^\star) $ is bounded by $ B^2 $, which only depends on $ k_1 $. For each of these transformations one then checks if it explains all $ (t_i, t_i^\star) $, where each of these checks is possible in polynomial time due to \cref{thm:propositionTransformationExplains}.

    For $ r > 1 $, instead of only computing the possible transformations, we compute all $ r $-tuples of possible transformations, resulting in $ B^{2r} $ tuples, where $ r $ is constant and $ B $ only depends on $ k_1 $.
\end{proofsketch}

\noindent
We will now sketch the proof of \cref{thmt@@oneoneatroot}.

\begin{proofsketch}
    The proof will consist of five steps to compute a transformation $ \rho \colon \sigma \to \tau $.
    \begin{enumerate}
        \item We compute a tree pattern that will serve as the body $ \sigma $ of $ \rho $. It will be ``as large as possible'' while still matching all $ t_i $. As a result, if there exists any transformation explaining all pairs, then there also exists one with $ \sigma $ as its body.
        \item We compute a compact representation of all the requirements induced by the target trees of the input pairs which we call ``tree of requirements''.
        \item Using the body and the compact representation from above, we compute the head $ \tau $ of $ \rho $.
        \item The resulting transformation is a solution candidate, but not necessarily a solution. As a final step, it has to be applied to each $ t_i $ and the result needs to be compared to $ t_i^\star $.
        \item Finally we show that the instance is positive if and only if the transformation computed as described is a solution.
    \end{enumerate}%

    Before going through these steps, we first check whether there exists a label from $ \Sigma $ which only appears in target trees, but not all target trees. If that is the case, we already know that the instance is negative: the head of a transformation would need to use said label and, as a consequence, would not be able to explain a pair in whose target tree this label does not occur. Since at least one of these exists, there cannot be a transformation explaining all pairs.

    In the following, we use the more intuitive notion of ``positions'' in trees and tree patterns, rather than having nodes from $ \N^* $. It could easily be transferred back to nodes, e.g. an ancestor of a position $ p $ is just a node corresponding to a prefix of the node at position $ p $.

    \proofstep{Step 1.}

    \begin{algorithm}[h]
        \caption{\textsc{AlgBody}}\label{alg:body}
            \begin{algorithmic}[1]
                \REQUIRE Source trees $ t_1, \dots, t_n $
                \ENSURE Pattern $ \sigma $ matching all $ t_i $ at the root which cannot be enlarged
                \STATE $ \mathtt{positions} $ := all positions across all $ t_i $
                \FORALL {$ \mathtt{positions} $ $ p $ in $ t_i $}
                    \STATE $\mathtt{map}[ p ]$ := new array of size $ n $
                    \FOR {$ i = 1, \dots, n $}
                        \IF {$ t_i $ has position $ p $}
                            \STATE $\mathtt{map}[p][i]$ := label at position $ p $ of $ t_i $
                        \ELSE
                            \STATE $\mathtt{map}[p][i]$ := $ \oslash $
                        \ENDIF
                    \ENDFOR
                \ENDFOR
                \STATE $ \mathtt{body\_positions} $ := $ \{\varepsilon\} $ /\!/ $ \varepsilon $ is the root 
                \FORALL {positions $ p \in \mathtt{positions} $, top-down, omitting root}
                \IF {$ p $'s parent is in $ \mathtt{body\_positions} $ \AND $ \mathtt{map}[p] $ and $ \mathtt{map}[p^s] $ do not contain $ \oslash $, for all siblings $ p^s $ of $ p $}
                    \STATE $ \mathtt{body\_positions} $ := $ \mathtt{body\_positions} \cup \{p\} $
                \ENDIF
                \ENDFOR
                \FORALL {positions $ p \in \mathtt{body\_positions} $}
                    \STATE $ \ell_\sigma(p) $ := $ \begin{cases}
                        x_p & \text{if } p \text{ is an inner node in any } t_i \text{ or } p \text{ is a leaf in all } t_i\\ %
                        Y_p & \text{otherwise}
                    \end{cases}
                     $
                \ENDFOR
                \RETURN pattern $ \sigma $ induced by labelling function $ \ell_p $
            \end{algorithmic}
    \end{algorithm}%

    We use \cref{alg:body} to compute the body $ \sigma $ of $ \rho $. Note that no variable occurs more than once in $ \sigma $. Positions are used as indices in variable names to ensure this here, but of course increasing indices starting at $ 1 $, which we typically use in examples, could be used instead. Very roughly, this requires time $ \bigO(m^2 \cdot n) $, where $ n $ is the number of example pairs and $ m $ is the size of a largest pair. We split the correctness of \cref{alg:body} into multiple claims.

    \begin{claim}\label{claim:syntactically-valid}
        $ \sigma $ computed by \cref{alg:body} is a syntactically valid tree pattern.
    \end{claim}
    \begin{claimproof}[Proof sketch]
        It is clear that $ \mathtt{body\_positions} $ (lines 9 - 12), the set of nodes of $ \sigma $, induces a tree due to the \textbf{if} condition in line 11. 
        It remains to show that tree variables only occur as leaves.
        This is follows from the definition of the labelling function in line 14 ensuring that tree variables are only used for nodes that are not inner nodes, i.e. leaves.
    \end{claimproof}

    We would like to demonstrate here the reasoning behind the part of the \textbf{if} condition requiring that $ \mathtt{map}[p^s] $ must not contain $ \oslash $ for any sibling $ p^s $ of $ p $, which we will call \emph{sibling condition}, with the following example:

    \begin{example}\label{ex:siblib-condition-body}
        Consider the trees 
        \begin{center}
            \tree{%
                \Tree[.\ensuremath{a} [.\ensuremath{b} \ensuremath{c} \ensuremath{d} ] \ensuremath{e} ]
            }
            \quad and \quad
            \tree{%
                \Tree[.\ensuremath{a} [.\ensuremath{b} \ensuremath{c} \ensuremath{d} ] ]
            }.
        \end{center}
        Only a tree pattern consisting of a single tree variable can match both trees at the root. Therefore, only the root may be in $ \mathtt{body\_positions} $, which is ensured by the sibling condition: if $ p $ is the position labelled $ b $, then the position labelled $ e $ in the left tree is a sibling $ p^s $ for which $ \mathtt{map}[p^s] $ contains $ \oslash $ due to the second tree. \qed
    \end{example}
    
    \begin{claim}\label{claim:matches-all-ti}
        $ \sigma $ matches all $ t_i $ at the root.
    \end{claim}
    \begin{claimproof}[Proof sketch]
        Two parts need to be shown:
        \begin{alphaenumerate}
            \item The structure ``fits'', i.e. each $ t_i $ has all the positions that $ \sigma $ has.
            \item The degree of positions labelled with a node variable fits in all $ t_i $, specifically %
                \begin{romanenumerate}
                    \item leaf positions in $ \sigma $ are leaves in all $ t_i $; and
                    \item if $ p $ is a position labelled with $ x $ in $ \sigma $ having children $ p_1, \dots, p_k $, $ p $ has degree $ k $ in all $ t_i $
                \end{romanenumerate}%
        \end{alphaenumerate}

        Part (a) is immediate due to the definition of $ \mathtt{body\_positions} $: there cannot be a position $ p $ in $ \mathtt{body\_positions} $ which is not present in all $ t_i $ because $ \mathtt{map}[p] $ would contain $ \oslash $.
        Part (b)(i) is clear due to the condition for node variables in the definition of the labelling function.  Part (b)(ii) is ensured by the definition of $ \mathtt{body\_positions} $: the case that $ p $ has degree $ < k $ in one of the $ t_i $ is excluded due to the sibling condition. Therefore, suppose $ p $ has degree $ \ge k $ in all $ t_i $ and $ > k $ in at least one source tree, say $ t_j $. Let $ p^w $ be a child position of $ p $ such that $ p^w $ exists in $ t_j $ and $ p_j > p_k $. Then, again due to the sibling condition, $ p_1, \dots, p_k \notin \mathtt{body\_positions} $, a contradiction.
    \end{claimproof}

    \begin{claim}\label{claim:body-is-maximal}
        \cref{alg:body} constructs a ``largest pattern'' $ \sigma $,  i.e. no additional positions can be added while still matching all $ t_i $.
    \end{claim}
    \begin{claimproof}
        First observe that positions that are not present in all $ t_i $ cannot be added.
        Therefore, only positions existing in all $ t_i $ need to be considered. 
        We will show that there cannot exist a tree pattern having such a position that matches all $ t_i $ at the root.
        Let $ p $ be a position existing in all $ t_i $, but $ p \notin \mathtt{body\_positions} $. Let $ p^a $ be the closest ancestor of $ p $ such that $ p^a \in \mathtt{body\_positions} $ and let $ \hat{p} $ be the child of $ p^a $ which is either ancestor of $ p $, or $ p $ itself. Note that $ p^a $ is a leaf in $ \sigma $. Since $ \hat{p} \notin \mathtt{body\_positions} $ and $ p^a \in \mathtt{body\_positions} $, $ \hat{p} $ must have violated the second half of the \textbf{if} condition in line 11. Also, since $ p $ exists in all $ t_i $, so does $ \hat{p} $. Therefore, $ \hat{p} $ must have violated the sibling condition. This implies that $ p^a $ has different degrees in the $ t_i $, so the only way to match $ p $ in all $ t_i $ is a tree variable (due to Condition \ref{item:tree-pattern} in the definition of tree patterns). Since tree variables must be leaves, no descendant of $ p^a $, which includes $ p $, can be a position in any pattern matching all $ t_i $ at the root.
    \end{claimproof}
    
    At this point, \cref{claim:syntactically-valid,claim:matches-all-ti,claim:body-is-maximal} imply that $ \sigma $ is a largest pattern matching all $ t_i $.

    \begin{claim}\label{claim:exists-one-with-sigma}
        If there exists some transformation $ \tilde{\rho} $ with body $ \sigma_{\tilde{\rho}} $ explaining all pairs in one step, then there also exists a transformation with body $ \sigma $ (called $ \rho $ below).
    \end{claim}
    \begin{claimproof}[Proof idea]
        We only give an informal explanation here and demonstrate it with an example; it will be clear how this can be adapted to the general case. The idea is that a transformation using $ \sigma $ as its body is more ``fine-grained'' than one using a smaller pattern. Due to \cref{claim:body-is-maximal}, any pattern matching all $ t_i $ must be a ``subpattern'' of $ \sigma $ (or $ \sigma $ itself), and such a subpattern can only be constructed from $ \sigma $ by using a tree variable earlier than necessary. Consider the following example as a demonstration of this idea:

        \begin{example}\label{ex:sigma-can-be-used-as-body}
            Suppose $ \sigma $ contains a node labelled with a node variable $ x $ that has exactly the two children $ Y_l $ and $ Y_r $, in which place $ \sigma_{\tilde{\rho}} $ only uses a tree variable $ Y $. There are two cases: (i) $ Y $ does not appear in the head of $ \tilde{\rho} $, or (ii) $ Y $ does appear in the head of $ \tilde{\rho} $. To ``simulate'' (i) with a transformation having $ \sigma $ as body, $ x $, $ Y_l $ and $ Y_r $ are omitted in the corresponding place in $ \rho $'s head. To simulate (ii), instead of $ Y $ we use the combination of $ x $, $ Y_l $ and $ Y_r $ in the head of $ \rho $ at the appropriate position.

            Note that in both cases we use the fact that both $ \sigma_{\tilde{\rho}} $ and $ \sigma $ match all $ t_i $. \qed
        \end{example}

        To generalise this, all possible subpatterns of $ \sigma $ would have to be considered and for each it has to be shown that they can be simulated analogously to \cref{ex:sigma-can-be-used-as-body} above. %
    \end{claimproof}

    We now know that $ \sigma $ is a tree pattern that we can definitely use as the body of our transformation.

    \proofstep{Step 2.} The tree of requirements $ \mathtt{req} $ is computed like $ \mathtt{map} $ in \cref{alg:body}, but for $ t_i^\star $ rather than $ t_i $, i.e. it is a tree whose nodes consist of arrays of length $ n $ (the number of example pairs) where index $ i $ describes the label in $ t_i^\star $ at the position given by the node. It therefore contains every position that occurs in some target tree.

    \proofstep{Step 3.}
    \begin{algorithm}
        \caption{\textsc{AlgHead}}\label{alg:head}
            \begin{algorithmic}[1]
                \REQUIRE tree of requirements $ \mathtt{req} $, $ \sigma $, $ ((t_i, t_i^\star))_{1\le i \le n} $
                \ENSURE the only possible head (which might still not result in a valid transformation)
                \FORALL {positions $ p $ in $ \mathtt{req} $}
                    \STATE $ \mathtt{poss}[p] $ := new array of length $ n $
                    \FOR {$ i = 1, \dots, n $}
                        \STATE $ \mathtt{poss}[p][i] $ := \textsc{AlgPossibilities}($ \sigma $, $ \mathrm{label}_{t_i^\star}(p) $, $ p $, $ t_i $, $ t_i^\star $)
                    \ENDFOR
                    \STATE $ \mathtt{head}[p] $ := $ \bigcap_i \mathtt{poss}[p][i] $
                \ENDFOR
                \vspace{1em}
                \STATE $ \mathtt{head\_positions} $ := $\left\{ p \quad \left| \quad
                \begin{minipage}{8.5cm}
                    \begin{alphaenumerate}
                        \item $ \mathtt{head}[p] \neq \emptyset $;
                        \item $ \mathtt{head}[p^a] \neq \emptyset $ for all ancestors $ p^a $ of $ p $; and
                        \item there exists no ancestor $ p^a $ of $ p $ such that $ \mathtt{head}[p^a] $ contains a tree variable
                    \end{alphaenumerate}
                \end{minipage}%
                \right.\right\}$
                \FORALL {positions $ h \in \mathtt{head\_positions} $}
                    \STATE $ \ell_\tau(h) $ := $ 
                    \begin{cases}
                        \text{(some) }Y & \text{such that }Y \in \mathtt{head}[h], Y \in \mathcal{T}\\%
                        \text{(some) }x & \text{such that }x \in \mathtt{head}[h], x \in \mathcal{N} \text{ and no tree variable in } \mathtt{head}[h]\\
                        \ell & \mathtt{head}[h] = \{\ell\}
                    \end{cases}
                    $
                \ENDFOR
                \RETURN head $ \tau $ induced by labelling function $ \ell_\tau $
            \end{algorithmic}
    \end{algorithm}%

    We use \cref{alg:head} to compute a tree pattern $ \tau $ that will serve as the head of our transformation $ \rho $. For \textsc{AlgPossibilities} used on line 4, see \cref{alg:possibilities} below. 
    We would like to comment on some aspects of the definition of the labelling function $ \ell_\tau $ in \cref{alg:head}:
    \begin{itemize}
        \item the actual choice of variable in the first two cases is arbitrary, since they all work. To make this well defined, one might just use the variable whose name is the lexicographically smallest
        \item a tree variable is chosen ``as soon as'' and whenever possible, which is reflected by condition (c) in line 6 and the first two cases in the definition of $ \ell_\tau $
    \end{itemize}%
    \cref{alg:head} roughly takes time $ \bigO(m^5 \cdot n) $, where $ n $ is the number of example pairs and $ m $ is the size of a largest pair, considering that \cref{alg:possibilities} takes time $ \bigO(m^4) $.

    \begin{algorithm}
        \caption{\textsc{AlgPossibilities}} \label{alg:possibilities}
            \begin{algorithmic}[1]
                \REQUIRE $ \sigma $, target label $ \ell $ and its position $ p_\ell $, $ t_i $, $ t_i^\star $
                \ENSURE set of possibilities to ``construct'' $ \ell $ from $ t_i $ based on $ \sigma $, i.e. variables that would work (and $ \ell $ itself)
                \IF {$ p_\ell $ is not a position in $ t_i^\star $}
                    \RETURN $ \emptyset $
                \ENDIF
                \STATE $ \mathtt{possibilities} $ := $ \{\ell\} $
                \STATE $ \mathtt{positions} $ := set of positions labelled with $ \ell $ in $ t_i $
                \FORALL {positions $ p \in \mathtt{positions} $}
                    \IF {$ p $ exists in $ \sigma $}
                        \STATE $ \mathtt{label} $ := $ \mathrm{label}_\sigma(p) $
                        \IF {($ \mathtt{label} \in \mathcal{T} $ \AND $ \mathrm{subtree}_{t_i}[p]  \cong \mathrm{subtree}_{t_i^\star}[p_\ell] $) \OR $ \mathtt{label} \notin \mathcal{T} $}
                            \STATE $ \mathtt{possibilities} $ := $ \mathtt{possibilities} \cup \{\mathtt{label}\} $
                        \ENDIF
                    \ENDIF
                \ENDFOR
                \RETURN $ \mathtt{possibilities} $
            \end{algorithmic}
    \end{algorithm}%

    \begin{claim}\label{claim:syntactically-valid-head}
        $ \tau $ computed in \cref{alg:head}
        is either an empty tree, 
        or a valid tree pattern that only uses variables from $ \sigma $.
    \end{claim}
    \begin{claimproof}[Proof sketch]
        The nodes of $ \tau $ are determined by the set $ \mathtt{head\_positions} $. 
        This can only be empty, if there does not exist a uniform way to create the correct label for the root of all $ t_i^\star $.
        If that is the case, the given instance of \algorithmicProblem{LearningTreeTransformations} is negative and Steps 4 and 5 can be skipped.

        If $ \mathtt{head\_positions} $ is not empty, it clearly induces a tree, because it is constructed from the tree of requirements by ``shortening'' or omitting branches (see conditions (a) - (c) in line 6).%

        To see that $ \tau $ is a valid tree pattern using only variables from $ \sigma $, we inspect $ \ell_\tau $. Since it only assigns labels returned by \textsc{AlgPossibilities} (lines 3 - 5), which only returns labels occurring in $ \sigma $ (or constant labels that occur in target trees), only variables from $ \sigma $ are used. In addition, tree variables are always leaves because all descendant positions of a position to which $ \ell_\tau $ assigns a tree variable are eliminated by condition (c) in line 6.
    \end{claimproof}

    Now, $ \ttrafo{\sigma}{\tau} $ is a syntactically valid tree pattern transformation that can be applied to the root of all $ t_i $.

    In addition, all positions that are ``influenced'' by $ \tau $ when applying $ \rho $ to $ t_i $ are correct, i.e. the labels are as required by $ t_i^\star $, for all $ i $. More precisely:%
    \begin{claim}\label{claim:head-nonempty-implies-correct-label}
        Let $ p $ be a position in $ \tau $. %
        Then, for all pairs $ (t_i, t_i^\star) $, $ \operatorname{label}_{t_i^\star}[p] = \operatorname{label}_{t_i^\rho}[p] $, where $ t_i^\rho $ is the tree resulting from applying $ \rho $ to the root of $ t_i $.
        Further, if $ p $ is labelled with a tree variable in $ \tau $, the subtrees rooted at $ p $ in $ t_i^\star $ and $ t_i^\rho $ are isomorphic.
    \end{claim}
    \begin{claimproof}[Proof sketch]
        Since $ p $ is a position in $ \tau $, we have that $ \mathtt{head}[p] \neq \emptyset $. Let $ \lambda \in \mathtt{head}[p] $. This label $ \lambda $ can be (i) a constant label from $ \Sigma $, (ii) a node variable or (iii) a tree variable. We consider (iii) separately below. In case (i), due to the intersection in line 5 of \cref{alg:head}, all $ t_i^\star $ have the same label, specifically $ \lambda $, in position $ p $. If (ii), $ \lambda $ is a node variable $ x $, let $ p_x $ be the position in $ \sigma $ labelled with $ x $. Due to the definition of $ \mathtt{positions} $ in line 4 of \textsc{AlgPossibilities}, the label at position $ p_x $ in $ t_i $ is the same as the label at position $ p $ in $ t_i^\star $, for all $ i $. As a result, applying $ \rho $ to (the root of) $ t_i $ yields $ \operatorname{label}_{t_i}[p_x] = \operatorname{label}_{t_i^\star}[p] $ as label at position $ p $ of $ t_i^\rho $. 

        We now consider case (iii), i.e. position $ p $ is labelled with a tree variable $ Y $. Analogously to case (ii), let $ p_Y $ be the position in $ \sigma $ labelled with $ Y $. Due to line 8 in \textsc{AlgPossibilities}, the subtrees rooted at $ p_Y $ in $ t_i $ and $ p $ in $ t_i^\star $ are isomorphic, for all pairs $ (t_i, t_i^\star) $. Therefore, 
        the subtrees of $ t_i^\star $ and $ t_i^\rho $ rooted at position $ p $ are isomorphic, so 
        applying $ \rho $ to $ t_i $ will result in the required subtree at position $ p $, for all $ i $.
    \end{claimproof}

    \proofstep{Step 4.} 
    Observe that \cref{claim:head-nonempty-implies-correct-label} does not imply that $ \rho $ is a valid solution, because
    there are cases in which Steps 1 - 3 yield a transformation which is not a valid solution:
    \begin{example}\label{ex:computed-trafo-needs-to-be-applied}
        This simple example illustrates the need to apply the computed transformation to ensure it is in fact a solution. Consider the pairs of trees 

        \begin{center}
            $\left(
                \ltree{%
                    \Tree[.\ensuremath{a} \ensuremath{b} \ensuremath{c} ]
                }, 
                \ltree{%
                    \Tree[.\ensuremath{a} \ensuremath{b} ]
                }
            \right)$
            \quad and \quad
            $\left(
                \ltree{%
                    \Tree[.\ensuremath{a} \ensuremath{b} \ensuremath{c} ]
                }, 
                \ltree{%
                    \Tree[.\ensuremath{a} \ensuremath{c} ]
                }
            \right)$.
        \end{center}

        \noindent
        The body computed in Step 1 (ignoring variation in variable names) would be $ \sigma = \scalebox{0.7}{\ltree{%
                    \Tree[.\ensuremath{x_1} \ensuremath{x_2} \ensuremath{x_3} ]
            }} $, while the head computed in Step 3 would just be $ x_1 $. But the resulting transformation $ \ttrafo{\sigma}{x_1} $ would not explain either of the two pairs. In fact, there is no single transformation explaining both pairs, which will become clear in Step 5.\qed
    \end{example}

    Therefore, to ensure the computed transformation is a solution, it has to be applied to each $ t_i $ and the result compared to $ t_i^\star $ in order to check that it explains each pair, which is possible in $ \bigO(m^2 \cdot n) $ time.

    \proofstep{Step 5.}
    We will now show that if the transformation computed in Steps 1 - 3 turns out not to be a solution in Step 4, then there is no solution for the given instance of \algorithmicProblem{LearningTreeTransformations}, by showing its contraposition:
    \begin{claim}\label{claim:positive-implies-rho-works}
        If there exists a transformation $ \tilde{\rho} $ explaining all $ (t_i, t_i^\star) $, then so does $ \rho $ resulting from the above algorithms.
    \end{claim}
    \begin{claimproof}[Proof sketch]
        Let $ \tilde{\rho} $ be a tree pattern transformation that explains all $ (t_i, t_i^\star) $.
        Due to \cref{claim:exists-one-with-sigma}, we can assume, without loss of generality, that $ \tilde{\rho} $ has body $ \sigma $. We call its head $ \tilde\tau $.

        Intuitively, our goal is to show that whenever there is a way to create a correct label at a certain position, \textsc{AlgPossibilities} will ``discover'' it (see \cref{claim:head-nonempty-implies-correct-label}).

        Towards a contradiction assume that $ \rho $ does not explain all pairs, i.e. there exists a pair $ (t, t^\star) $ such that $ t \not\rightsquigarrow_\rho t^\star $. Let $ t' $ be the tree resulting from applying $ \rho $ to the root of $ t $. By assumption, $ t' \neq t^\star $. 
        Let $ p' $ be a position at which $ t' $ and $ t^\star $ differ.

        We distinguish between two cases:
        \begin{enumerate}
            \item $ p' $ exists in $ \tau $;
            \item $ p' $ does not exist in $ \tau $.
        \end{enumerate}%

        \proofstep{1. $ p' $ exists in $ \tau $.}
        Then by \cref{claim:head-nonempty-implies-correct-label}, $ t^\star $ and $ t' $ cannot differ at position $ p' $, a contradiction. %

        \proofstep{2. $ p' $ does not exist in $ \tau $.}
        Note that $ p' $ cannot be the root position in this case, because $ \tau $ being the empty tree implies that the instance is negative (see proof sketch of \cref{claim:syntactically-valid-head}), a contradiction to $ \tilde{\rho} $ being a solution.

        Let $ p^a $ be the closest ancestor of $ p' $ in $ \tau $ and $ \tilde{p}^a $ be the closest ancestor of $ p' $ in $ \tilde\tau $ (including $ p' $ itself if $ p' $ is in $ \tilde\tau $). 

        We further distinguish between two cases based on how $ p^a $ and $ \tilde{p}^a $ relate to each other:
        \begin{alphaenumerate}
            \item $ \tilde{p}^a $ is a strict descendant of $ p^a $ (i.e. a position on the path from $ p^a $ to $ p' $, excluding $ p^a $, but including $ p' $).
            \item $ \tilde{p}^a $ is an ancestor of or equal to $ p^a $ (and therefore not equal to $ p' $)
        \end{alphaenumerate}

        \proofstep{2(a) $ \tilde{p}^a $ is a strict descendant of $ p^a $.}
        Then the set $ \mathtt{head}[p_{\text{path}}] $ must be non-empty for all positions $ p_{\text{path}} $ on the path from $ p^a $ down to $ \tilde{p}^a $, since it will at least contain the label of $ \tilde\tau $ at said position.

        Because of this, $ p^a $ must be labelled with a tree variable $ Y $ in $ \tau $: 
        positions $ p_{\text{path}} $ for which $ \mathtt{head}[p_{\text{path}}] \neq \emptyset $ and $ \mathtt{head}[p_{\text{path}}^a] \neq \emptyset $ for all ancestors of $ p_{\text{path}} $ up to, but excluding, $ p^a $, can only have been eliminated by condition (c) in line 6 of \cref{alg:head}.

        Now \cref{claim:head-nonempty-implies-correct-label} yields that the subtrees of $ t^\star $ and $ t' $ at position $ p^a $ are isomorphic, and therefore $ t' $ and $ t^\star $ cannot differ at position $ p' $, a contradiction.

        \proofstep{2(b) $ \tilde{p}^a $ is an ancestor of or equal to $ p^a $.}
        By assumption we know that $ p' $ is not in $ \tilde\tau $. 

        First consider the case that $ p' $ exists in $ t^\star $. Then $ \tilde{p}^a $ must be labelled with a tree variable in $ \tilde\tau $, which will in turn be contained in $ \mathtt{head}[\tilde{p}^a] $. As a result, $ \tilde{p}^a $ must be labelled with a tree variable in $ \tau $.
        Note that since $ p^a $ exists in $ \tau $, no strict ancestor of $ p^a $ can be labelled with a tree variable in $ \tau $, so it can only be the case that $ p^a = \tilde{p}^a $. 

        Now consider the case that $ p' $ does not exist in $ t^\star $. Since $ t' $ and $ t^\star $ differ in $ p' $, $ p' $ must exist in $ t' $. Then $ p^a $ must be labelled with a tree variable in $ \tau $ (since $ p' \neq p^a $).
        In both cases, \cref{claim:head-nonempty-implies-correct-label} yields the desired contradiction, because an ancestor of $ p' $ is labelled with a tree variable in $ \tau $.

        \par\medskip
        All cases result in a contradiction, therefore $ \rho $ must explain all pairs if $ \tilde{\rho} $ does.
    \end{claimproof}

    Overall we have shown that $ ((t_i, t_i^\star))_{1\le i \le n} $ is a positive instance of \algorithmicProblem{LearningTreeTransformations} with $ s $ and $ r $ set to $ 1 $, if and only if $ \rho\colon \ttrafo{\sigma}{\tau} $ explains all $ (t_i, t_i^\star) $ (when only applying transformations at the root).
    The total time required for Steps 1 - 5 is $ \bigO(m^5 \cdot n) $ (due to \cref{alg:head}).
\end{proofsketch}
 \end{document}